\documentclass[11pt]{article}
\usepackage[margin=1in]{geometry}    

\usepackage{times}  
\usepackage{helvet}  
\usepackage{courier}  
\usepackage[hyphens]{url}  
\usepackage{graphicx} 
\usepackage{natbib}  
\usepackage{caption} 
\usepackage{algorithm}
\usepackage{algorithmicx}

\usepackage{newfloat}
\usepackage{listings}
\DeclareCaptionStyle{ruled}{labelfont=normalfont,labelsep=colon,strut=off} 
\floatstyle{ruled}
\newfloat{listing}{tb}{lst}{}
\floatname{listing}{Listing}
\usepackage{wrapfig}
\usepackage{caption}   
\usepackage[utf8]{inputenc} 
\usepackage[T1]{fontenc}    
\usepackage{hyperref}       
\usepackage{url}            
\usepackage{booktabs}       
\usepackage{amsfonts}       
\usepackage{graphicx}       
\usepackage{nicefrac}       
\usepackage{xcolor}         
\usepackage{amsmath} 
\usepackage{xspace}
\usepackage{amsthm}
\usepackage{minitoc}    
\usepackage{multirow}
\usepackage{booktabs}
\usepackage[normalem]{ulem}
\usepackage{caption}
\usepackage{amssymb}
\usepackage{pifont}
\usepackage{hyperref}
\usepackage{url}
\usepackage{color}
\usepackage{multirow}
\usepackage{graphicx}
\usepackage{subfigure}
\usepackage{textcomp}
\usepackage{algorithm}
\usepackage{algpseudocode} 
\usepackage{amsmath,amssymb} 
\usepackage{bbding}
\usepackage{graphicx}
\usepackage{tabularx}
\usepackage{amsmath,xparse,mleftright}
\usepackage[export]{adjustbox}
\usepackage{balance}
\usepackage{tabularray}
\usepackage{hyperref}
\usepackage{enumitem}
\usepackage{pifont}
\usepackage{xspace}
\usepackage{setspace}
\usepackage{tabularx,ragged2e}
\usepackage{bm}
\usepackage{algpseudocode}
\usepackage{tcolorbox}
\usepackage{fontawesome5}
\usepackage{bbm}
\usepackage{newtxtext}  
\usepackage{amsfonts}  
\usepackage{amssymb}
\tcbuselibrary{breakable,skins}

\newtheorem{assumption}{Assumption}[section] 

\newcommand{\warn}{\raisebox{\ups\height}{\includegraphics[width=1em]{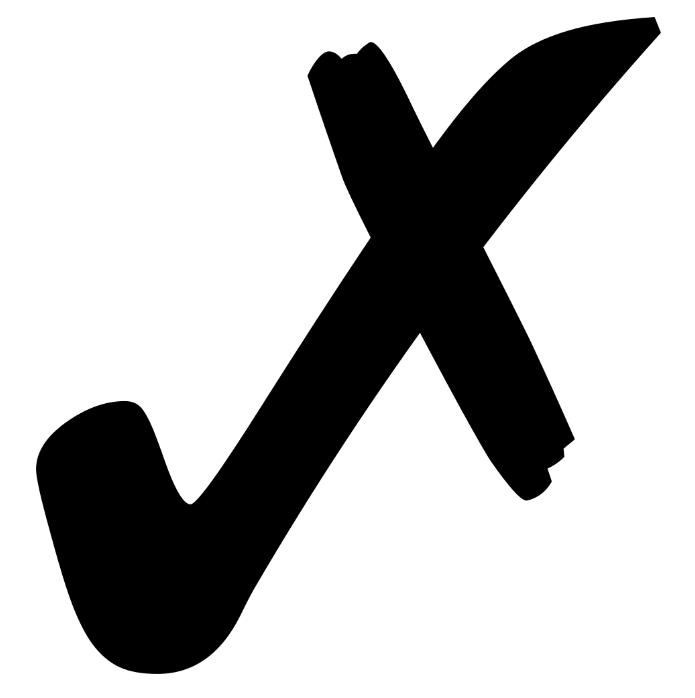}}}

\newtheorem{theorem}{Theorem}
\newtheorem{lemma}{Lemma}

\newcommand{\name}{\textsc{SkillGen}\xspace}

\def\ups{-0}

\title{\name: Learning Domain Skills for In-Context Sequential Decision Making}
\author{
  Ruomeng Ding\thanks{This work was done while the first author was affiliated with Georgia Institute of Technology.} \\
  University of North Carolina at Chapel Hill \\
  \texttt{ruomeng@unc.edu}
  \and
  Wei Cheng \\
  NEC Laboratories America \\
  \texttt{weicheng@nec-labs.com}
  \and
  \quad\quad\quad Minglai Shao\thanks{Corresponding author.} \\
 \quad\quad\quad Tianjin University \\
  \quad\quad\quad \texttt{shaoml@tju.edu.cn} 
  \and
  \quad\quad\quad Chen Zhao \\
  \quad\quad\quad Baylor University \\
  \quad\quad\quad \texttt{chen\_zhao@baylor.edu}
}

\date{}     
\begin{document}

\maketitle

\begin{abstract}
Large language models (LLMs) are increasingly applied to sequential decision-making through in-context learning (ICL), yet their effectiveness is highly sensitive to prompt quality. Effective prompts should meet three principles: focus on decision-critical information, provide step-level granularity, and minimize reliance on expert annotations through label efficiency. However, existing ICL methods often fail to satisfy all three criteria simultaneously. Motivated by these challenges, we introduce \name, a skill-based ICL framework for structured sequential reasoning. It constructs an action-centric, domain-level graph from sampled trajectories, identifies high-utility actions via temporal-difference credit assignment, and retrieves step-wise skills to generate fine-grained, context-aware prompts. We further present a theoretical analysis showing that focusing on high-utility segments supports task identifiability and informs more effective ICL prompt design. Experiments on ALFWorld, BabyAI, and ScienceWorld, using both open-source and proprietary LLMs, show that \name achieves consistent gains, improving progress rate by 5.9\%–16.5\% on average across models. The implementation of \name is available at https://github.com/ruomengd/SkillGen.
\end{abstract}

\section{Introduction\label{sec:intro}}

\begin{wraptable}{r}{7cm}
\centering
\vspace*{-1.3em}
\scriptsize {
\renewcommand{\arraystretch}{0.4}  
\renewcommand{\tabcolsep}{2pt} 
\caption{Comparison of ICL Methods.\label{tab:compare}}
\vspace*{-0.8em}
\begin{tabular}{lccc}
\toprule
\textbf{ICL Method} & \textbf{Focused} & \textbf{Granular} & \textbf{Label-efficient} \\
\midrule
Fixed Prompting         & \XSolidBrush & \XSolidBrush & \Checkmark \\
Task-level Retrieval  & \warn  & \warn & \XSolidBrush \\
Step-wise Retrieval  & \warn  & \Checkmark & \XSolidBrush \\
Insight Summary    & \Checkmark  & \XSolidBrush
& \warn \\
\midrule
\name (ours)      & \Checkmark & \Checkmark & \Checkmark \\
\bottomrule
\end{tabular}}
\vspace*{-1em}
\end{wraptable}

Large language models (LLMs) are increasingly applied to multi-step decision-making tasks across domains such as embodied control~\cite{li2024embodied, yang2025embodiedbench, ma2024agentboard}, text-based games~\cite{yao2023tree, liuagentbench, klissarov2024modeling}, and online shopping~\cite{yao2022webshop, zhou2023webarena}. These tasks require agents to operate in dynamic environments, interact with the world through sequences of actions, and pursue long-horizon goals. In contrast to supervised fine-tuning (SFT) methods~\cite{chen2023fireact, qintoolllm, zeng-etal-2024-agenttuning}, which depend on large-scale expert demonstrations, in-context learning (ICL) offers a more lightweight and efficient alternative by guiding inference with only a few examples~\cite{brown2020language, achiam2023gpt}. Consequently, ICL has emerged as a central reasoning paradigm in many LLM-based agent frameworks for decision-making~\cite{yao2023react, shinn2023reflexion, sun2023adaplanner, wu2025avatar}. To make ICL effective for multi-step tasks, prompt design should adhere to three key principles: (1) \textit{Focused}: emphasize decision-critical information while minimizing redundant context; (2) \textit{Granular}: offer fine-grained, step-level guidance that aligns to the evolving task state; (3) \textit{Label-efficient}: replaces costly expert trajectories with subgoal completion and success signals, which offer a more scalable and structurally aligned form of supervision.

Recent advances in ICL have enhanced performance by shifting from fixed, hand-crafted prompts to more context-sensitive prompt designs~\cite{liu2022makes, zhou2022large, zhang2022active, liskavets2025prompt}. As summarized in Table~\ref{tab:compare}, various methods target different aspects of prompt design. Fixed prompting relies on limited examples that ignore task-specific context. Although label-free, such prompts fail to provide actionable, decision-level guidance. Task-level retrieval methods, such as Synapse~\cite{synapse2023}, retrieve full expert demonstrations based on task metadata and use them as few-shot exemplars. While this improves contextual relevance, the retrieved prompts often include redundant steps and lack step-level resolution, limiting both focus and granularity. Step-level retrieval strategies, such as Trad~\cite{zhou2024trad}, enhance granularity by retrieving trajectory fragments at each decision step. However, the retrieved actions are often disconnected and lack structural coherence, which undermines decision focus. These approaches often depend on expert-annotated trajectories, thereby reducing their label efficiency. Insight summarization approaches, including Leap~\cite{leap2024} and ExpeL~\cite{expel2024}, generate high-level insights by comparing correct and wrong solutions. Yet these summarized insights are often too abstract to support intermediate decisions, providing limited fine-grained guidance. While each of these approaches addresses different aspects of prompt design, none of them simultaneously meets all three core principles.

\begin{figure*}[!t]
\centering
\includegraphics[width=0.95\textwidth]{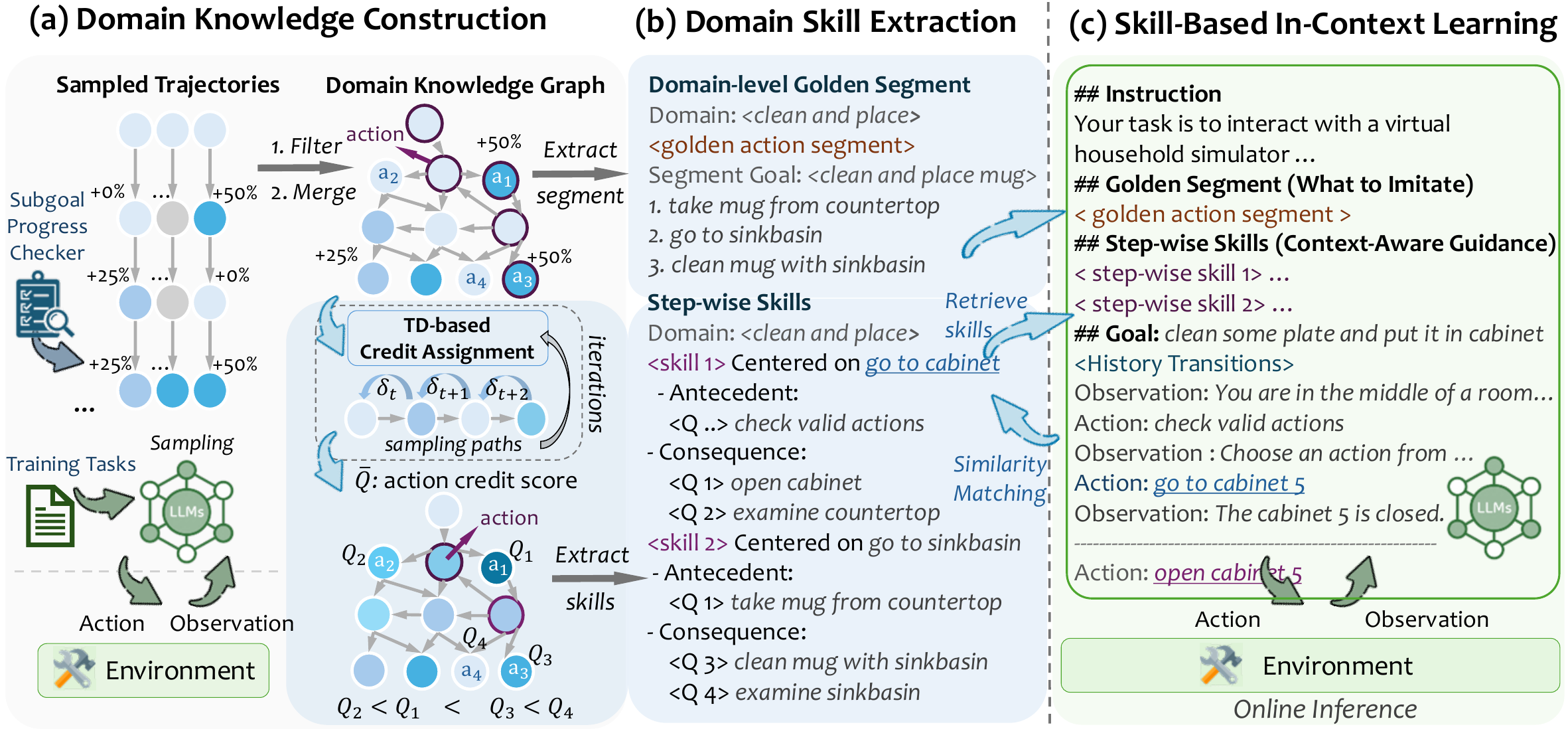}
\caption{Framework of \name.\label{fig:framework}}
\end{figure*}

To address these limitations, we propose \name, an ICL framework that extracts and applies \textbf{domain-oriented} and \textbf{action-centric skills}. As illustrated in Figure~\ref{fig:framework}, \name operates in three stages, the first two stages are performed offline, while the third leverages the extracted skills to generate actions step-by-step during inference:
(1) \textit{Domain Knowledge Construction} – We construct an action-centric domain knowledge graph from sampled trajectories, effectively capturing the structural dynamics of the task.
(2) \textit{Domain Skill Extraction} – Temporal-Difference (TD) based credit assignment is employed to identify actions that consistently contribute to task progress;
(3) \textit{Skill-Based In-Context Learning} – During inference, \name combines a golden segment with step-wise skills retrieved based on the current transition history to guide action generation. 
To support focused reasoning, \name constructs prompts around decision-critical skill segments while filtering out irrelevant context. For fine-grained guidance, it encodes temporal and structural dependencies from the domain graph to retrieve skills that align with the current task history. To promote label efficiency, \name leverages subgoal completion progress as weak supervision and applies TD-based action credits to extract skills, eliminating the need for full expert trajectories.

\name improves average progress rate by 5.9\%–16.5\% across ALFWorld, BabyAI, and ScienceWorld, showing clear gains in sequential decision-making. To conclude, our primary contributions are as follows:
\begin{itemize} [leftmargin=*]
\item We address the challenge of designing in-context learning (ICL) prompts that jointly support decision focus, step-level granularity, and label efficiency in sequential decision-making tasks.
\item We propose \name, a framework that learns domain-oriented skills to support focused, fine-grained ICL. Theoretical analysis shows that high‑utility segments enhance task identifiability and ICL prompts.
\item Our empirical results show that \name consistently improves progress and success rates across various tasks. Ablations show that both the golden segment and step-wise skill retrieval contribute to performance gains.
\end{itemize}

\section{Background}

\paragraph{In-Context Learning.} 
Xie et al.~\cite{xieexplanation} model in-context learning (ICL) as an instance of implicit Bayesian inference. In this view, a language model infers a latent task parameter \( \phi \in \Phi \) from the observed context \( \mathcal{C} \) (e.g., a sequence of demonstrations or interaction history), forming a posterior \( p(\phi \mid \mathcal{C}) \)~\cite{ min2022rethinking, falck2024context}. Given a query input \( x \in \mathcal{X} \), the model predicts by computing the posterior predictive distribution:
\begin{equation}
p(y \mid x, \mathcal{C}) = \int_\phi p(y \mid x, \phi) \, p(\phi \mid \mathcal{C}) \, d\phi,
\end{equation}
where \( y \in \mathcal{Y} \) is the predicted output and \( p(y \mid x, \phi) \) is the task-specific likelihood. Wies et al.~\cite{wies2023learnability} formalize this intuition within a PAC framework by modeling pretraining as sampling from a latent task mixture \( D = \sum_{\phi \in \Phi} \pi(\phi) P_\phi \), where \( \pi(\phi) \) is the prior of latent task \( \phi \), and \( P_\phi \) denotes the corresponding data distribution. In this view, ICL serves to recover the underlying task \( \phi^\star \) from the prompt, enabling accurate prediction without updating model parameters.


\paragraph{LLMs for Sequential Decision-making.}
We consider adopting LLMs as autonomous agents for sequential decision-making. In such environments, agents cannot directly observe the underlying state. We model this setting as a Partially Observable Markov Decision Process (POMDP)~\cite{he2024words, sun2024interactive}, defined by $\text{POMDP} = (\mathcal{S}, \mathcal{A}, \Omega, P, R, O)$, where $\mathcal{S}$ denotes the latent state space, $\mathcal{A}$ the discrete action space, $\Omega$ the observation space, $P$ the state transition function, $R$ a sparse, progress-based reward function, and $O$ the observation model. At each time step $t$, the agent receives a partial observation $o_t \sim O(o_t \mid s_t)$ and selects an action $a_t$ based on the interaction transition: $h_t = \{(o_0, a_0), (o_1, a_1), \dots, o_t\}$. Given the current history $h_t$, the LLM generates the next action through a prompting mechanism:
\begin{equation}
\pi(a_t \mid h_t) = \text{LLM}(a_t \mid \text{Prompt}(h_t)).
\end{equation}
This formulation enables LLMs to serve as decision-making agents in partially observable settings, leveraging contextual information without access to full environment states or explicit policy optimization. The agent aims to maximize cumulative task progress over long-horizon episodes.

\section{Method}

To enable focused and fine-grained in-context learning, we introduce two forms of decision-critical knowledge derived from structured action knowledge~\cite{jain2022know,10.1007/s10514-023-10133-5,zhu2024knowagent}: 
(1) \emph{Golden segment} – a concise, high-quality action sequence extracted from training trajectories within the task domain, selected for its maximal contribution toward goal completion; (2) \emph{Step-wise skills} – reusable local patterns centered on a key action, summarizing its typical antecedents and consequences within the domain. For example, for the action \textit{“go to cabinet”}, common antecedents include \textit{“check valid actions”} and \textit{“examine countertop”}, while a likely consequence is \textit{“open cabinet”}.


\subsection{Domain Knowledge Construction}

To induce domain-level knowledge, we first sample diverse trajectories from LLMs using high-temperature stochastic decoding. Each training instance is denoted by \( d = (m, g) \in \mathcal{D}_{\text{train}} \), where \( m \) is the task domain information and \( g \) is the task goal. For each instance, \( N \) trajectories are sampled, forming:
\begin{equation}
\begin{aligned}
\mathbb{T}_{\text{train}} &= \{(m, g, \mathcal{T}) \mid (m, g) \in \mathcal{D}_{\text{train}} \}, \\
\mathcal{T} &= \{(o_t, a_t, p_t)\}_{t=0}^T,
\end{aligned}
\end{equation}
where each step includes the observation \( o_t \), action \( a_t \), and progress signal \( p_t \). To reduce noise and task-specific variance, trajectories are filtered by discarding invalid actions and those with zero final progress (\( p_T = 0 \)). Actions are abstracted by removing object-specific identifiers (e.g., \textit{``open cabinet 5''} \(\rightarrow\) \textit{``open cabinet''}) to reveal transferable patterns across instances, yielding sequences of action–progress pairs. The filtered trajectories for each domain \( m \) form:
\begin{equation}
\mathbb{T}_m = \{ \mathcal{T}_1, \mathcal{T}_2, \dots, \mathcal{T}_K \}, \quad \text{with } \mathcal{T}_k = \{(a_t, p_t)\}_{t=0}^{T_k}.
\end{equation}
From these, a directed domain knowledge graph \( \mathbb{G}_m = (\mathbb{V}, \mathbb{E}) \) is constructed, where each node \( a \in \mathbb{V} \) denotes an action, and each edge \( (a_i, a_j) \in \mathbb{E} \) indicates a transition between consecutive actions. All paths share a common start \( a_{\text{start}} \) and end node \( a_{\text{end}} \) to ensure structural consistency. Edges are annotated with sets of observed progress deltas:
\begin{equation}
\begin{aligned}
\mathcal{P}_{\Delta}(a_i, a_j) = \{ p_{t+1} - p_t \mid (&a_t, p_t), (a_{t+1}, p_{t+1}) \in \mathcal{T},\\ 
\; &a_t = a_i,\; a_{t+1} = a_j\}
 \end{aligned}
\end{equation}

To maintain graph quality, self-loops are removed and low-impact nodes pruned based on $\mathcal{P}_{\Delta}(a_i, a_j)$ once the graph exceeds a reasonable scale. The resulting graph captures reusable action patterns and domain-level decision dynamics, offering a structured foundation for skill extraction. We extract a \emph{golden segment}—a short action sequence with the highest subgoal progress rate within the domain—to serve as a concise, decision-critical exemplar for focused prompting.

\subsection{Domain Skill Extraction}

\begin{wrapfigure}{r}{8cm}
\centering
\vspace*{-1.5em}
\includegraphics[width=8cm]{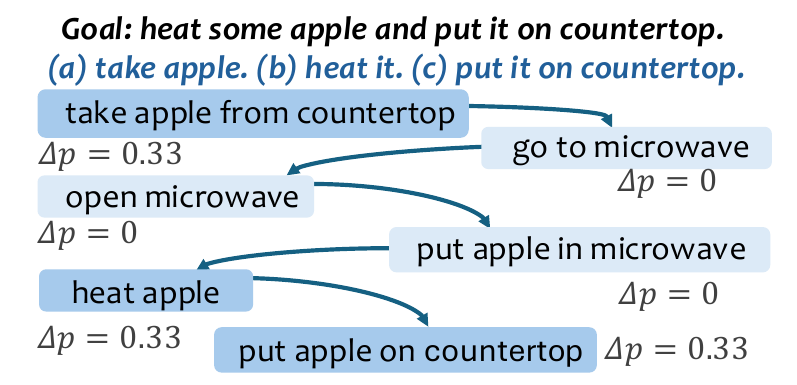} 
\caption{Sparse $\mathcal{P}_{\Delta}(a_i, a_j)$ reward only subgoal completions, omitting intermediate actions.\label{fig:illustration}}
\vspace*{-1em}
\end{wrapfigure}

The progress delta $\mathcal{P}_{\Delta}$, annotated on graph edges, provides sparse, subgoal-level feedback by marking major milestones such as completing a cleaning step or reaching a destination. However, it often fails to capture the contribution of intermediate actions that enable these outcomes. As illustrated in Figure~\ref{fig:illustration}, only a few steps in the task \textit{“heat some apple and put it in countertop”} receive positive deltas (e.g., \textit{heating and placing the apple}), while prerequisite actions—like navigating to and opening the microwave—remain uncredited despite being causally necessary. This illustrates the need for credit assignment to assign action value more accurately.

To address it, we treat $\mathcal{P}_{\Delta}$ as delayed rewards and propagate them backward along sampled action paths to estimate fine-grained action utility. While various metrics~\cite{agarwal2020optimistic, mesnard2021counterfactual} provide alternative means of estimating action importance, we adopt temporal difference learning with eligibility traces (TD($\lambda$))~\cite{sutton1988learning} to learn an action-value \( Q(a) \), capturing the expected long-term credit of each action.


\paragraph{TD-based credit assignment.}  
In the action-centric graph \( \mathbb{G}_m = (\mathbb{V}, \mathbb{E}) \), each iteration samples a set of action paths from a designated start node \( a_{\text{start}} \) to an end node \( a_{\text{end}} \) for downstream credit estimation. Formally, for iteration \( i = 1, \dots, N \), we sample:
\begin{equation}
\tau^{(i)} = (a^{(i)}_1, a^{(i)}_2, \dots, a^{(i)}_{T_i}) \sim \mathcal{S}_{\text{path}}(a_{\text{start}}, a_{\text{end}}, \mathbb{G}_m),
\end{equation}
where each \( \tau^{(i)} \) is a valid path in \( \mathbb{G}_m \) satisfying $(a^{(i)}_t, a^{(i)}_{t+1}) \in \mathbb{E}$, $a^{(i)}_1 = a_{\text{start}}$, $a^{(i)}_{T_i} = a_{\text{end}}$. Here, \( \mathcal{S}_{\text{path}} \) denotes a uniform distribution over all valid paths from \( a_{\text{start}} \) to \( a_{\text{end}} \) in \( \mathbb{G}_m \).
For each path \( \tau = [a_0, a_1, \dots, a_T] \), action credits are estimated using temporal-difference learning with eligibility traces. At each step \( t \), the reward \( r_t \) is defined based on empirical progress deltas. If such records exist, i.e., \( \mathcal{P}_{\Delta}(a_t, a_{t+1}) \ne \emptyset \), a value is uniformly sampled and perturbed with Gaussian noise:
\begin{equation}
r_t \sim \mathrm{Uniform}(\mathcal{P}_{\Delta}(a_t, a_{t+1})) + \mathcal{N}(0, \sigma^2),
\end{equation}
where \( \mathrm{Uniform}(\cdot) \) denotes a uniform distribution over finite set $\mathcal{P}_{\Delta}(a_t, a_{t+1})$. Otherwise, the reward defaults to pure noise, \( r_t=\epsilon \), treating the transition as a step without observable progress. For each action \(a_t\) along path \(\tau\), the temporal-difference (TD) error is computed as:
\begin{equation}
\delta_t = r_t + \gamma Q(a_{t+1}) - Q(a_t).
\end{equation}
To propagate credit backward, the eligibility trace for the current action \( a_t \) is incremented, and all actions with nonzero trace values are updated (see detailed algorithm in Appendix~\ref{appendix:td_estimation}):
\begin{equation}
\begin{aligned}
E(a_t) &\leftarrow E(a_t) + 1, \quad
Q(a) \leftarrow Q(a) + \alpha \delta_t E(a),\\
E(a) &\leftarrow \gamma \lambda E(a), \quad \forall a \in \{ a' \mid E(a') > 0 \}.
\end{aligned}
\end{equation}
Here, \( \alpha \) is the learning rate, \( \gamma \) the discount factor, and \( \lambda \) the trace decay rate. The eligibility trace \( E(a) \) accumulates credit for recently visited actions, enabling delayed reward signals to influence earlier decisions across the sampled paths. After learning, \( Q(a) \) values are normalized into a dense credit distribution:
\begin{equation}
\bar{Q}(a) = \frac{\max(Q(a), 0)}{\sum_{a'} \max(Q(a'), 0)},
\end{equation}
emphasizing both goal-reaching actions and the intermediate steps that enable them. Additional details are provided in Appendix~\ref{appendix:graph_construction}. Structural statistics of domain knowledge graphs are provided in Appendix~\ref{appendix:domin_graph}.

\paragraph{Skill Extraction.}
For each action node \( a \) in the domain graph, we extract a local skill centered on \( a \) using its immediate neighbors and credit scores. We define the \emph{antecedent set} as the set of predecessor actions (incoming edges), and the \emph{consequence set}  as the set of successor actions (outgoing edges). Each action is assigned a normalized credit score from TD-based propagation, and both sets are ranked accordingly. As shown in Figure~\ref{fig:framework} (b), the resulting skill is formulated as:
\begin{equation}
\text{Skill}(a) = \left( a, \textit{Antecedents}(a), \textit{Consequences}(a) \right),
\end{equation}
where \( a \) is the central action, and the ranked context captures its typical usage patterns. This structure supports context-aware retrieval of reusable action-centric skills.

\subsection{Skill-Based In-Context Learning}
Building on structured skills extracted via TD-based credit assignment, we develop a retrieval-augmented prompting strategy that conditions a frozen LLM on both reusable domain knowledge and the agent’s interaction history (Figure~\ref{fig:framework}, right). The prompt integrates two levels of contextual information: (i) a domain-level \emph{golden segment}—a concise, high-impact action sequence selected offline to serve as a focused exemplar; and (ii) \emph{step-wise skills}—local, action-centric transitions retrieved based on the most recent action \( a_{t-1} \).

Concretely, at each time step \( t \), given a history transition \( h_t = \{(o_0, a_0), \dots, (o_{t-1}, a_{t-1}), o_t\} \), a pretrained semantic retriever processes the recent action \( a_{t-1} \) and retrieves the most relevant action \( \hat{a} \) from the domain graph. The final prompt is formed by concatenating the golden segment from domain \( m \), the retrieved skills \( \text{Skill}(\hat{a}) \), and the current trajectory history, each serialized into natural language. This composite prompt enables context-aware, credit-guided reasoning without requiring any parameter updates to the LLM. The next action is then sampled autoregressively:
\begin{equation}
\begin{aligned}
\Phi &= \text{GoldenSegment}(m) \oplus \text{Skill}(\hat{a}),\\
\pi(a_t &\mid h_t, \Phi) = \text{LLM}(a_t \mid \text{Prompt}(h_t, \Phi)).
\end{aligned}
\end{equation}

\subsection{Theoretical Analysis}

In sequential reasoning tasks, prompts often include both decision-relevant and irrelevant segments, whereas uninformative tokens may obscure task-specific signals. Following the latent task mixture framework of Wies et al.~\cite{wies2023learnability}, we study how selecting a focused subset of high-utility content improves task identification. Assume prompts are drawn from a mixture distribution \( \mathcal{D} = \sum_{\phi \in \Phi} \pi(\phi) P_\phi \), where \(\phi\) indexes latent tasks, \(\pi(\phi)\) is the prior, and \(P_\phi\) the task-specific sequence distribution. Each input \(x_t = (g, h_t)\) includes a goal \(g\) and history \(h_t = \{(o_0, a_0), \dots, o_t\}\), and \(y_t = a_t\) is the action. We decompose the prompt as \(p = p_{\text{focused}} \cup p_{\text{irrelevant}}\), defined relative to the ground-truth task \(\phi^\star\): \(p_{\text{focused}} := p_{\text{focused}} \mid \phi^\star\), capturing informative segments, and \(p_{\text{irrelevant}} := p_{\text{irrelevant}} \mid \phi^\star\), capturing segments with little or misleading task evidence. Under Assumptions A.1–A.3 (see Appendix~\ref{appendix:theory}), we show that the focused portion of the prompt alone suffices for task identifiability. See Appendix~\ref{appendix:theory} for full derivation.

\begin{theorem}[Task Identifiability] \label{thm:task-identifiability}
Let \(p \sim P_{\phi^\star}^{\otimes k}\) be a prompt sampled from the true task \(\phi^\star \in \Phi\), and suppose it admits a decomposition \(p = p_{\text{focused}} \cup p_{\text{irrelevant}}\), where the partition is defined relative to \(\phi^\star\). Suppose that $\min_{\phi \ne \phi^\star} \mathrm{KL}(P_{\phi^\star}(p) \| P_\phi(p)) > 8 \log \left( \frac{1}{c_1 \cdot c_2} \right)$. Then, there exists a sample complexity threshold \(\tilde{m}_\mathcal{D} : (0, 1)^2 \rightarrow \mathbb{N}\) such that for any \(\epsilon, \delta > 0\), if the number of in-context examples \(k \ge \tilde{m}_\mathcal{D}(\epsilon, \delta)\), the following holds with probability at least \(1 - \delta\) over the sampling of \(p\):
\begin{equation}
\forall \phi \ne \phi^\star, \quad
\frac{P_\phi(p_{\text{focused}})}{P_{\phi^\star}(p_{\text{focused}})} 
\le 
\frac{P_\phi(p)}{P_{\phi^\star}(p)} 
< \epsilon.
\end{equation}
\end{theorem}
This result suggests that removing irrelevant segments sharpens the contrast between tasks, leading to a more task-aligned prompt with reduced ambiguity.

\section{Experimental Setup}

\paragraph{Datasets.}
We conduct experiments on three sequential decision-making datasets: ALFWorld~\cite{shridhar2021alfworld}, BabyAI~\cite{chevalier-boisvert2018babyai}, and ScienceWorld~\cite{wang-etal-2022-scienceworld}. These benchmarks span household tasks, grid-based navigation, and scientific reasoning, requiring agents to perform multi-turn interactions to achieve final goals. They cover diverse domains and increase in complexity—from ALFWorld to ScienceWorld. We employ four-fold cross-validation, partitioning each dataset into four equal splits, using three folds for training and one for testing, and report results averaged over all folds. Based on the subgoal annotations provided by AgentBoard~\cite{ma2024agentboard}, we compute the subgoal achieved rate to quantify the model's step-wise progress. Notably, we use subgoal labels to construct domain graphs but do not rely on full expert trajectories during training or inference.

\paragraph{Evaluation Metrics.}
All methods are evaluated using four metrics: Grounding Rate (GR), Progress Rate (PR), Success Rate (SR), and Area Under the Progress Curve (AUPC). GR measures whether the agent’s action is valid given the current environment state, reflecting its grounding and understanding capabilities. PR quantifies task advancement based on the proportion of subgoals achieved. SR indicates whether the agent successfully completes the entire task. AUPC captures the cumulative task progress over time, rewarding agents that complete tasks more efficiently.

\paragraph{Baselines.}
We consider the following baselines: \textit{0-shot} asks the agent to perform the task without any in-context examples. \textit{1-shot} provides a single demonstration trajectory as an in-context example. \textit{Leap}~\cite{leap2024} enables the agent to self-revise by identifying and learning from mistakes in provided examples. \textit{Synapse}~\cite{synapse2023} retrieves and prompts entire expert trajectories from memory based on task meta-data. \textit{Trad}~\cite{trad2024} guides inference by retrieving observation-action pairs from past interaction history. All baseline methods are built on the \textit{Act} prompting framework~\cite{yao2023react}, chosen for its simplicity and broad compatibility with instruction-following LLMs. To reduce foundation model bias, we evaluate all baselines using three models: Qwen2.5-7B-Instruct~\cite{yang2024qwen2}, Qwen-Turbo~\cite{yang2024qwen2}, and GPT-4o-mini~\cite{hurst2024gpt}. We use each language model for both sampling and inference. For more experimental details, please refer to Appendix~\ref{appendix:setup}.

\begin{table*}[!t]
\centering
\tiny
\caption{Grounding Rate $[\%]$ ($\uparrow$), Progress Rate $[\%]$ ($\uparrow$), Success Rate $[\%]$ ($\uparrow$), and AUPC $[0, 1]$ ($\uparrow$) on ALFWorld. The best method for each  LLM is in \textbf{bold}; the second-best method is \underline{underlined}.~\label{tab:main_alf}}
\vspace{-1.2em}
\resizebox{0.95\linewidth}{!}{
\renewcommand{\arraystretch}{0.9}  
\begin{tabular}{c|cccc|cccc|cccc} 
\toprule
\multirow{2}{*}{\textbf{Method}} & \multicolumn{4}{c|}{\textbf{Qwen2.5-7B-Instruct}}      & \multicolumn{4}{c|}{\textbf{Qwen-Turbo}}      & \multicolumn{4}{c}{\textbf{GPT-4o-mini}}       \\
                                 & \textbf{GR }   & \textbf{PR }   & \textbf{SR } &\textbf{AUPC}  & \textbf{GR }   & \textbf{PR }   & \textbf{SR }  &\textbf{AUPC} & \textbf{GR }   & \textbf{PR }   & \textbf{SR }  &\textbf{AUPC}  \\ 
\midrule
0-shot                     & 10.5                & 6.0                  & 0.8                  & 0.027          & 56.3                & 32.2                 & 9.7                  & 0.212          & 73.7                 & 26.8                 & 1.5                  & 0.184                 \\
1-shot                     & 28.1                & 16.0                 & 2.2                  & 0.095          & 63.9                & 55.3                 & 36.5                 & 0.380          & 77.3                 & 43.3                 & 10.5                 & 0.292                 \\
Leap                             & 27.7                & 21.2                 & 5.2                  & 0.125          & 66.3                & 55.6                 & 37.3                 & 0.386          & 78.6                 & 50.8                 & 11.2                 & 0.348         \\
Synapse (1-shot)                 & 61.5                & 41.6                 & 17.1                 & 0.278          & 74.9                & 54.7                 & 35.8                 & 0.379          & 76.3                 & 48.8                 & 14.8                 & 0.340                 \\
Synapse (3-shot)                 & \underline{71.4}        & \underline{44.8}         & 19.4                 & \underline{0.302}  & \underline{78.4}        & \underline{60.6}         & \underline{47.0}         & \underline{0.421}  & 77.4                 & \underline{52.9}        & \underline{17.8}         & \underline{0.360}        \\
Trad                             & 65.4                & 44.2                 & \underline{22.4}         & 0.296          & 65.5                & 54.8                 & 35.8                 & 0.372          & \underline{79.1}         & 49.1                 & 16.4                 & 0.341                 \\
\name (ours)                         & \textbf{84.9}       & \textbf{68.0}        & \textbf{55.2}        & \textbf{0.464} & \textbf{85.9}       & \textbf{67.6}        & \textbf{53.8}        & \textbf{0.460} & \textbf{83.6}        & \textbf{55.1}         & \textbf{29.8}        & \textbf{0.369}                 \\
\bottomrule
\end{tabular}}
\end{table*}

\begin{table*}[!t]
\centering
\tiny
\caption{Grounding Rate $[\%]$ ($\uparrow$), Progress Rate $[\%]$ ($\uparrow$), Success Rate $[\%]$ ($\uparrow$), and AUPC $[0, 1]$ ($\uparrow$) on BabyAI. The best method for each  LLM is in \textbf{bold}; the second-best method is \underline{underlined}.~\label{tab:main_ba}}
\vspace{-1.2em}
\resizebox{0.95\textwidth}{!}{
\renewcommand{\arraystretch}{0.9}  
\begin{tabular}{c|cccc|cccc|cccc} 
\toprule
\multirow{2}{*}{\textbf{Method}} & \multicolumn{4}{c|}{\textbf{Qwen2.5-7B-Instruct}}                       & \multicolumn{4}{c|}{\textbf{Qwen-Turbo}}                       & \multicolumn{4}{c}{\textbf{GPT-4o-mini}}                        \\
                                 & \textbf{GR}   & \textbf{PR}   & \textbf{SR}   & \textbf{AUPC}  & \textbf{GR}   & \textbf{PR}   & \textbf{SR}   & \textbf{AUPC}  & \textbf{GR}   & \textbf{PR}   & \textbf{SR}   & \textbf{AUPC}   \\ 
\midrule
0-shot                     & 31.8          & 21.8          & 7.1           & 0.037          & 50.2          & 32.7          & 19.6          & 0.092          & 55.3          & 34.2          & 22.3          & 0.129           \\
1-shot                     & 59.2          & 36.5          & 18.8          & 0.112          & 61.4          & 37.2          & 16.3          & 0.076          & 76.6          & 42.6          & 28.6          & 0.154           \\
Leap                             & 66.6          & \underline{46.3}  & 27.7  & 0.151          & \underline{68.4}          & 52.9          & 38.4          & 0.206          & 73.1          & 43.8          & 29.4          & 0.170           \\
Synapse (1-shot)                 & 67.2          & 39.4          & 21.4          & 0.153          & 65.0  & \underline{55.8}  & \underline{45.5}  & \underline{0.242}  & 86.5          & 44.9          & 33.9          & 0.169           \\
Synapse (3-shot)                 & \textbf{78.6} & 44.6          & \underline{28.6}          & \textbf{0.163}  & 62.1          & 50.8      & 38.4  & 0.191 & \textbf{92.8} & \underline{49.5}  & \underline{38.4}  & \underline{0.188}   \\
Trad                             & \underline{68.2}  & 36.9          & 19.7          & 0.115          & 64.9          & 46.9          & 34.8          & 0.157          & 87.3          & 40.9          & 30.4          & 0.126   \\
\name (ours)                         & 66.7          & \textbf{50.0} & \textbf{31.2} & \underline{0.158} & \textbf{73.9} & \textbf{59.4}          & \textbf{45.5}          & \textbf{0.254}          & \underline{89.5}  & \textbf{57.6} & \textbf{41.1} & \textbf{0.248}  \\
\bottomrule
\end{tabular}}
\end{table*}

\begin{table*}[!t]
\centering
\tiny
\caption{Grounding Rate $[\%]$ ($\uparrow$), Progress Rate $[\%]$ ($\uparrow$), Success Rate $[\%]$ ($\uparrow$), and AUPC $[0, 1]$ ($\uparrow$) on ScienceWorld. The best method for each  LLM is in \textbf{bold}; the second-best method is \underline{underlined}.~\label{tab:main_sc}}
\vspace{-1.2em}
\resizebox{0.95\textwidth}{!}{
\renewcommand{\arraystretch}{0.9}  
\begin{tabular}{c|cccc|cccc|cccc} 
\toprule
\multirow{2}{*}{\textbf{Method}} & \multicolumn{4}{c|}{\textbf{Qwen2.5-7B-Instruct}}      & \multicolumn{4}{c|}{\textbf{Qwen-Turbo}}      & \multicolumn{4}{c}{\textbf{GPT-4o-mini}}       \\
                                 & \textbf{GR }   & \textbf{PR }   & \textbf{SR } &\textbf{AUPC}  & \textbf{GR }   & \textbf{PR }   & \textbf{SR }  &\textbf{AUPC} & \textbf{GR }   & \textbf{PR }   & \textbf{SR }  &\textbf{AUPC}  \\ 
\midrule
0-shot                     & \underline{10.8}  & \underline{27.1}  & 9.0           & 0.136          & \textbf{28.8} & 19.3          & 4.4           & 0.113          & \underline{34.3}  & 44.3          & 7.7           & 0.206                 \\
1-shot                     & 9.6           & 19.1          & 5.5           & 0.108          & 11.8          & 19.1          & 7.7           & 0.111          & \textbf{34.3} & 46.8          & 18.8          & 0.294                 \\
Leap                             & 8.5           & 25.8          & \underline{11.1}  & \underline{0.155}  & 4.4           & 21.8          & 6.6           & 0.124          & 11.4          & 51.7          & 21.1          & 0.330                 \\
Synapse (1-shot)                 & 8.6           & 15.4          & 4.4           & 0.090          & 5.4           & 16.0          & 3.3           & 0.106          & 14.0          & 52.8          & 25.6          & 0.334                 \\
Synapse (3-shot)                 & 6.0           & 15.3          & 5.5           & 0.086          & 7.2           & 24.0          & \underline{10.1}  & 0.155          & 15.5          & 60.0          & \underline{32.4}  & \underline{0.390}         \\
Trad                             & 7.3           & 21.1          & 4.4           & 0.135          & 7.1           & \underline{29.3}  & 8.8           & \underline{0.180}  & 16.9          & \underline{61.4}  & 29.0          & 0.375                 \\
\name (ours)                         & \textbf{16.1} & \textbf{46.7} & \textbf{23.4} & \textbf{0.298} & \underline{13.3}  & \textbf{37.7} & \textbf{11.1} & \textbf{0.242} & 25.3         & \textbf{67.3} & \textbf{40.2} & \textbf{0.442}        \\
\bottomrule
\end{tabular}}
\vspace{-1.5em}
\end{table*}

\section{Evaluation Results}
\name is evaluated on three decision‑making tasks with performance, ablation, and sensitivity studies. Appendix~\ref{appendix:additional_results} further reports token‑cost comparisons, TD‑sampling strategies, and an Act/ReAct inference study. Prompt examples and case studies can be found in Appendices~\ref{appendix:examples} and \ref{appendix:case_study}.
Our key findings are as follows:
\vspace{-0.3em}
\begin{itemize}[leftmargin=*]
\item \name consistently achieves higher PR and SR than baseline methods across all tasks. AUPC measures \name’s efficiency in steady progress beyond final success.
\item Step‑wise skills and golden segments boost performance, TD‑based credit assignment refines skills for better guidance, and subgoal supervision aids decisions via intermediate structure.
\item \name demonstrates strong performance across a range of sampling scales and retrieval granularities, with balanced configurations performing consistently well.
\end{itemize}

\subsection{Main Results}

\paragraph{ALFWorld.} 
Table~\ref{tab:main_alf} presents the comparison of prompting strategies on ALFWorld. While \name achieves the highest GR across all models, the most notable gains appear in PR and SR. On Qwen2.5-7B-Instruct, \name achieves a PR of 68.0 and an SR of 55.2, substantially outperforming the strongest baseline, Synapse (3-shot), which achieves 44.8 (PR) and 19.4 (SR). On Qwen-Turbo, \name reaches 67.6 (PR) and 53.8 (SR), surpassing the second-best Synapse (3-shot) with 60.6 and 47.0, respectively. Similar trends are observed on GPT-4o-mini, where \name boosts SR from 17.8 to 29.8, again outperforming all alternatives. \name outperforms Synapse and Trad by providing more reusable skills, yielding denser step‑wise progress, higher subgoal completion, and the best AUPC for efficient task execution.

\paragraph{BabyAI.} 
Table~\ref{tab:main_ba} shows that \name achieves the highest PR scores across all models—reaching 59.4 on Qwen-Turbo and 57.6 on GPT-4o-mini.
For SR, \name shows significant improvements: a +7.0\% gain on GPT-4o-mini (41.1 vs. 38.4). On Qwen2.5-7B-Instruct, \name achieves 50.0 (PR) and 31.2 (SR), outperforming Synapse (3-shot) at 44.6 and 28.6, respectively. Compared to \textit{Leap} and \textit{Trad}, \name achieves higher PR and SR—for example, +10.7 SR over \textit{Trad} on GPT-4o-mini—highlighting the advantage of structured skill prompting. These results demonstrate that \name outperforms both insight-summary and trajectory-retrieval baselines in spatially constrained environments.

\paragraph{ScienceWorld.} 
In Table~\ref{tab:main_sc}, \name achieves a PR of 67.3 and an SR of 40.2 on GPT-4o-mini—a +7.8\% improvement in SR over the strongest baseline, Synapse (3-shot), and +11.2\% over \textit{Trad}—while also attaining the highest AUPC of 0.442. On Qwen-Turbo, the best baseline is \textit{Trad} with 29.3 (PR) and 8.8 (SR), while \name improves these to 37.7 and 11.1. Notably, naive baselines such as \textit{0-shot} show inflated GR scores by repeatedly issuing generic queries like \textit{"check valid action"}, which accumulate steps but fail to make substantive progress. \name consistently leads in AUPC across models, highlighting its efficiency in making steady progress. These results demonstrate that \name’s structured prompting enables more coherent, goal-directed reasoning and higher task success, even in knowledge-intensive, instruction-heavy scenarios like ScienceWorld.

\subsection{Ablation Study}

\begin{figure}[!t]
\centering
\subfigure[ALFWorld]{
\includegraphics[width=0.32\linewidth]{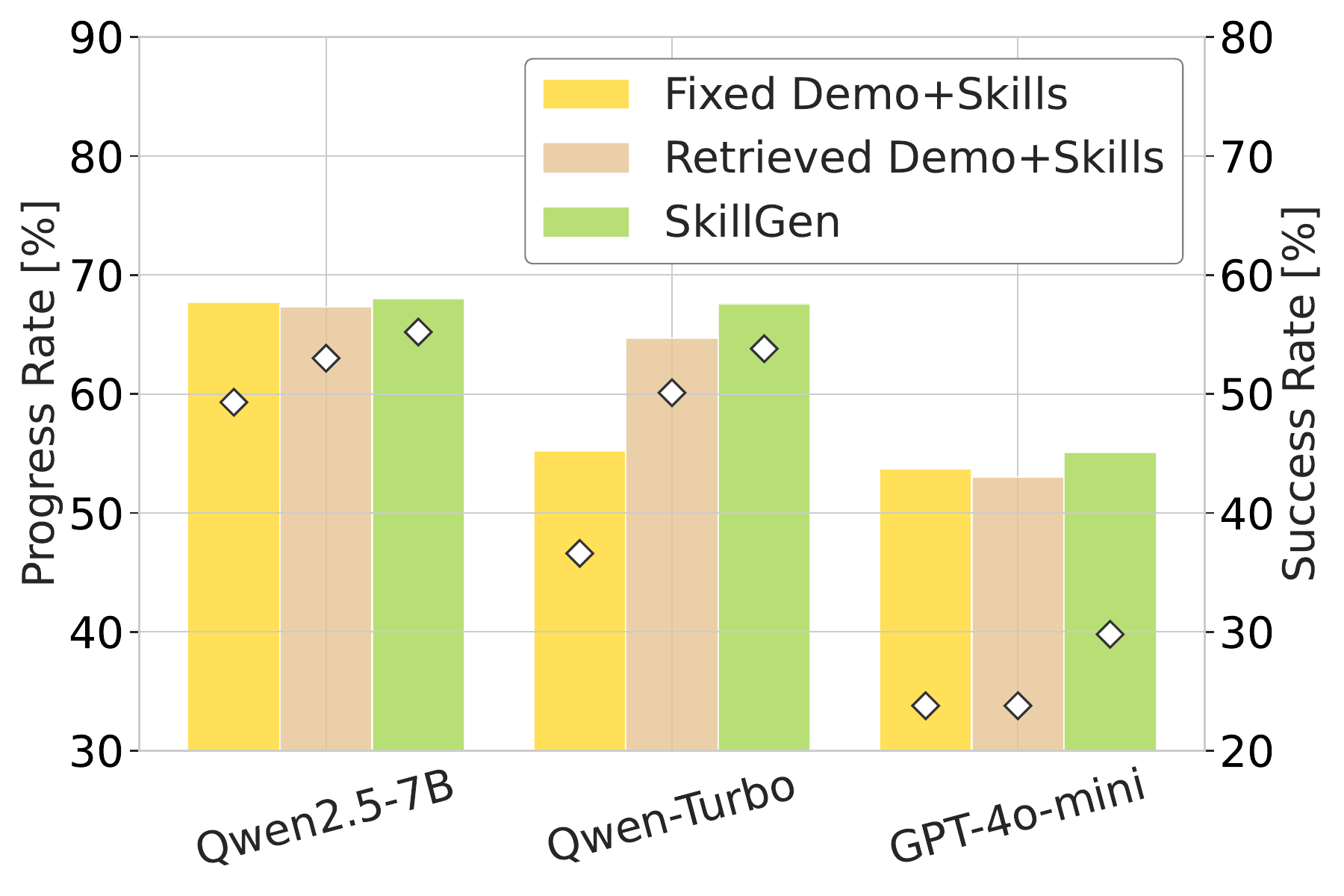}}
\subfigure[BabyAI]{
\includegraphics[width=0.32\linewidth]{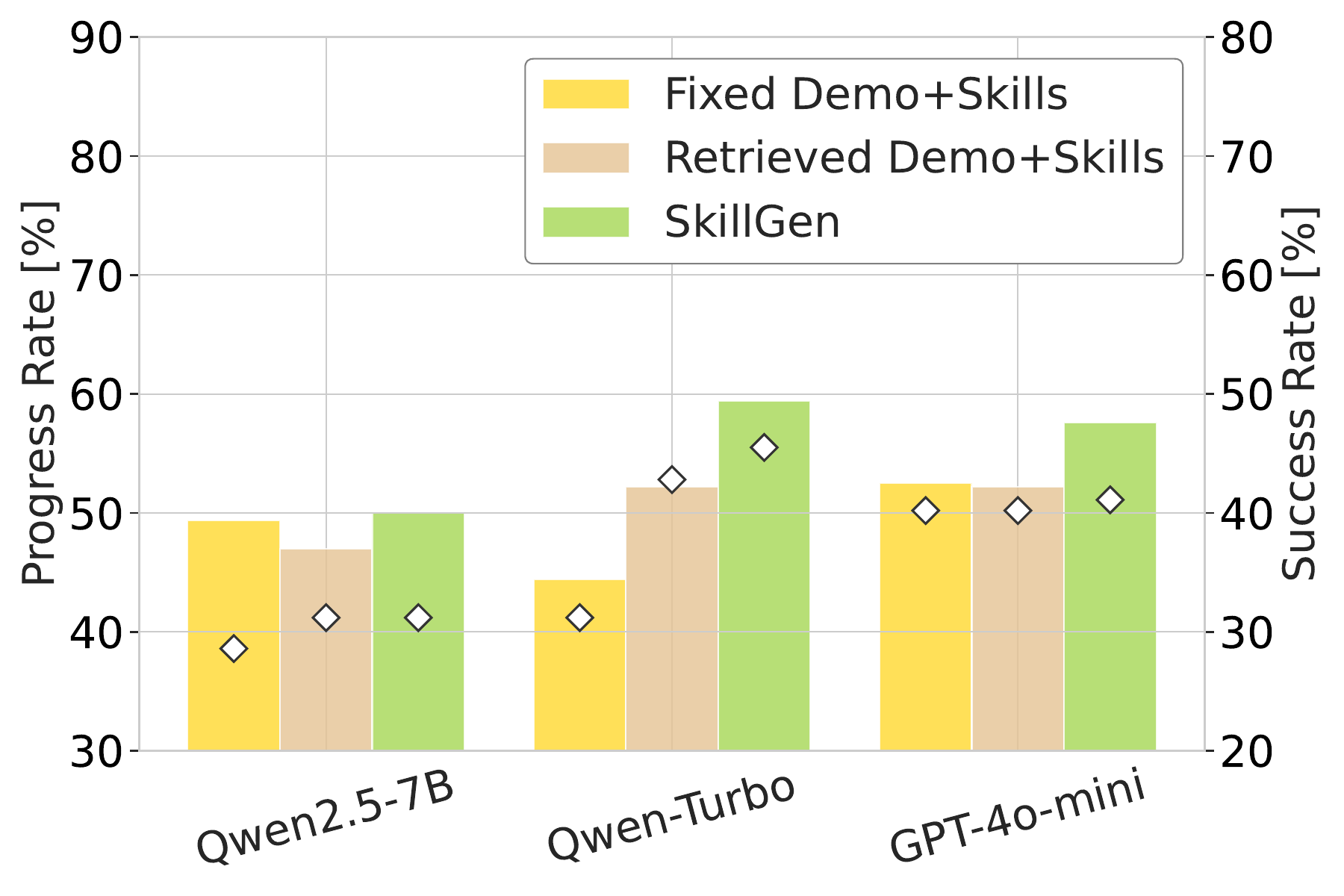}}
\subfigure[ScienceWorld]{
\includegraphics[width=0.32\linewidth]{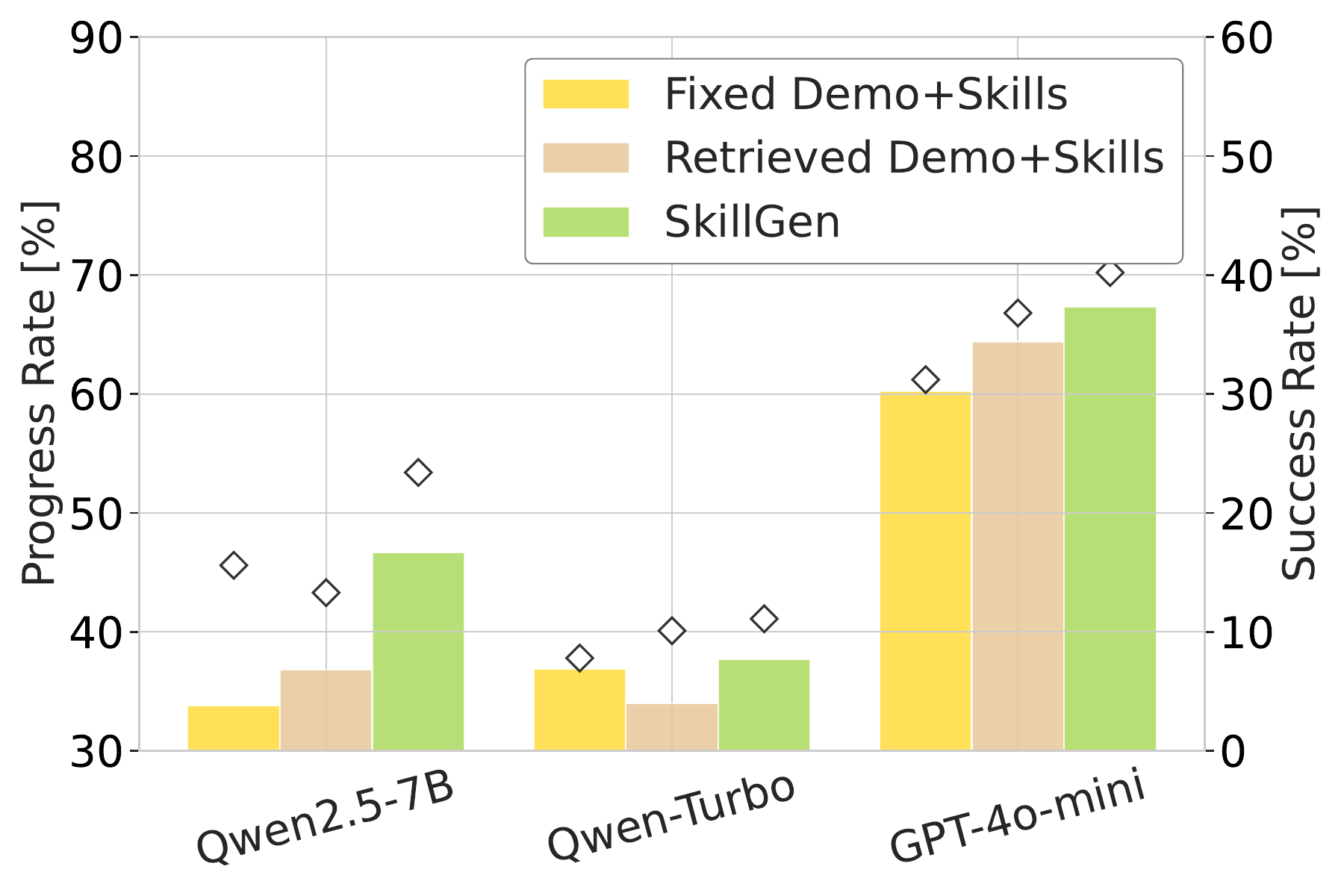}}
\vspace{-0.8em}
\caption{Ablation study of \name showing the effect of golden segments. Bars represent PR, $\Diamond$ markers indicate SR.\label{fig:ablation_segment}}
\vspace{-1.2em}
\end{figure}

\begin{figure}[!t]
\centering
\subfigure[ALFWorld]{
\includegraphics[width=0.32\linewidth]{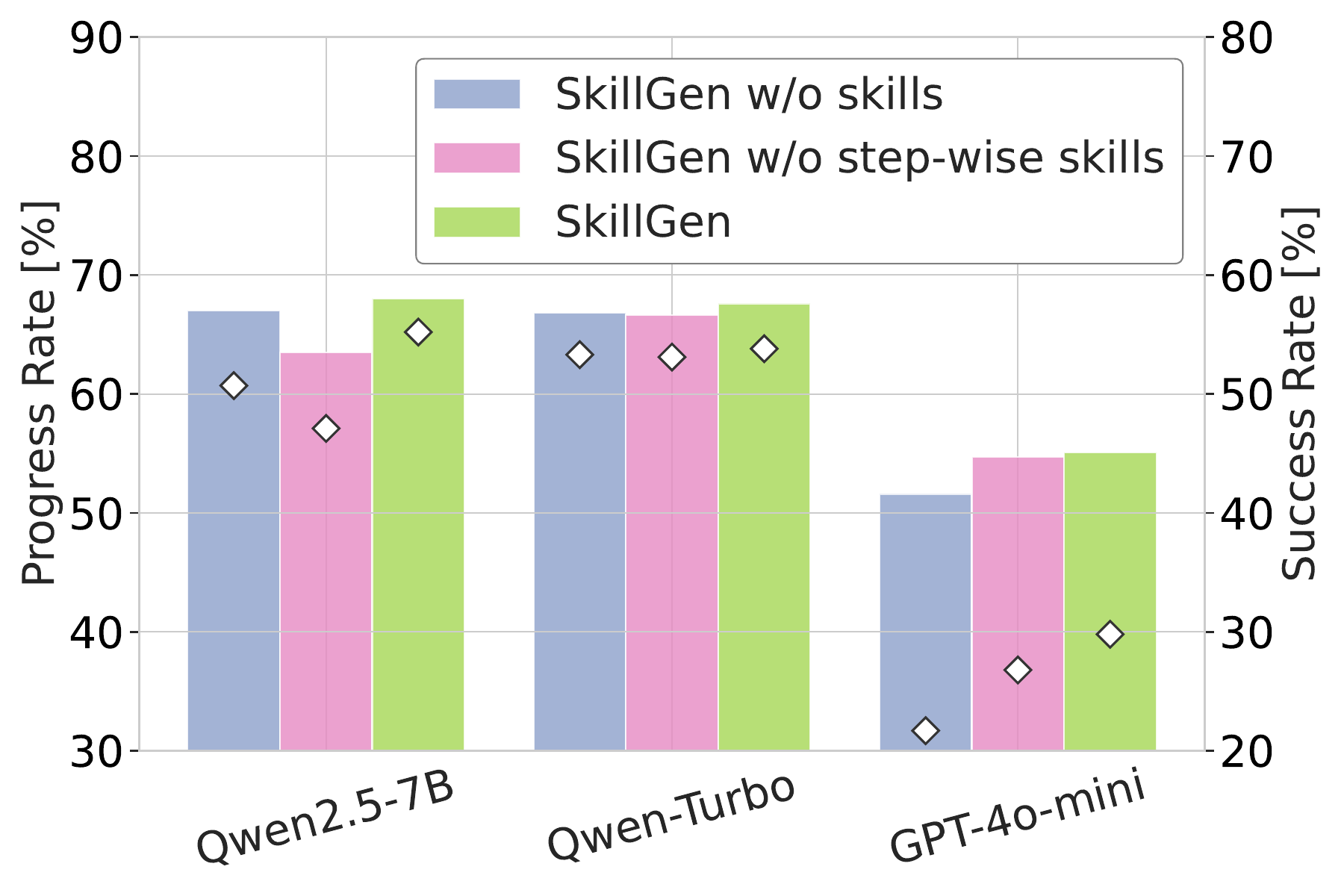}}
\subfigure[BabyAI]{
\includegraphics[width=0.32\linewidth]{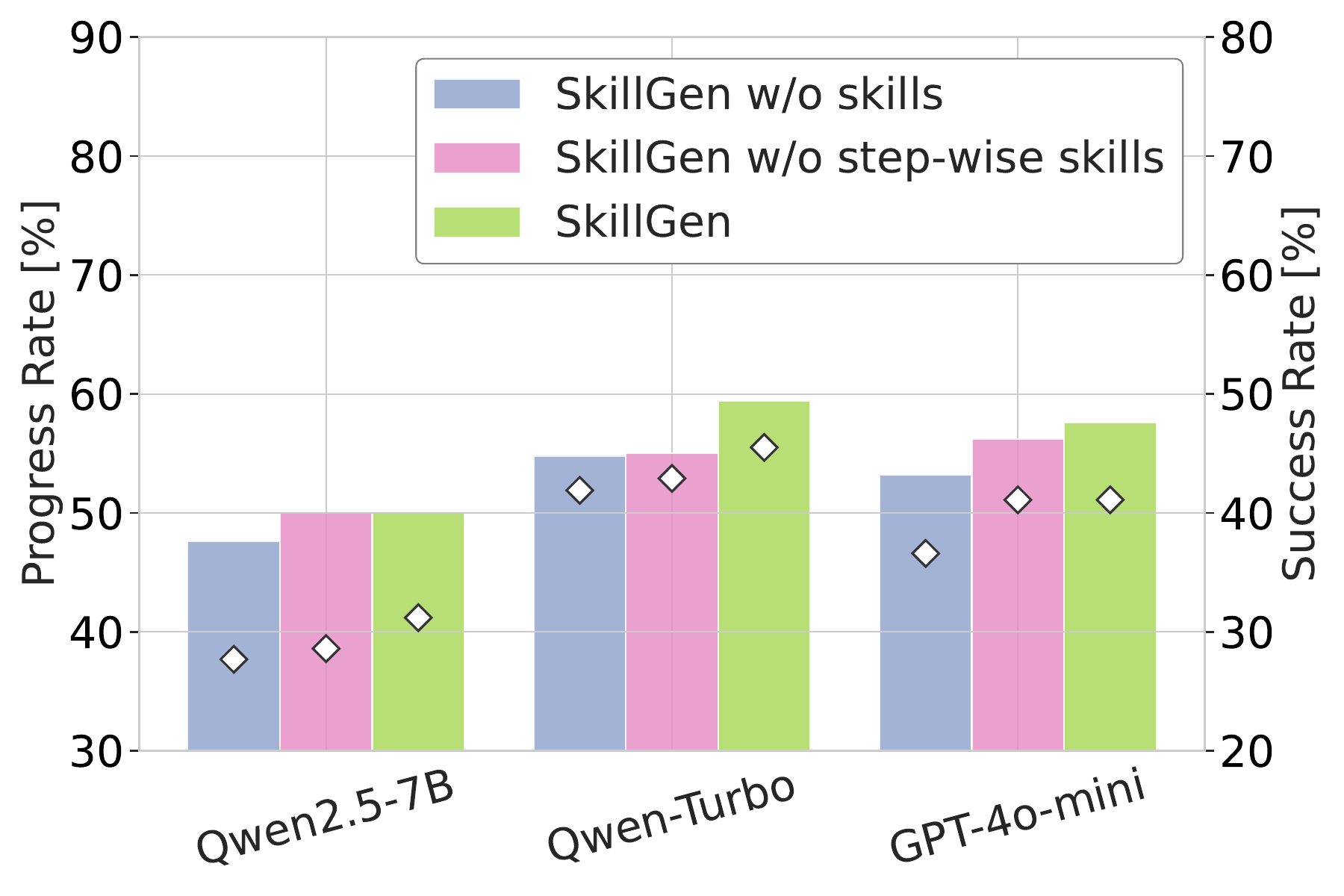}}
\subfigure[ScienceWorld]{
\includegraphics[width=0.32\linewidth]{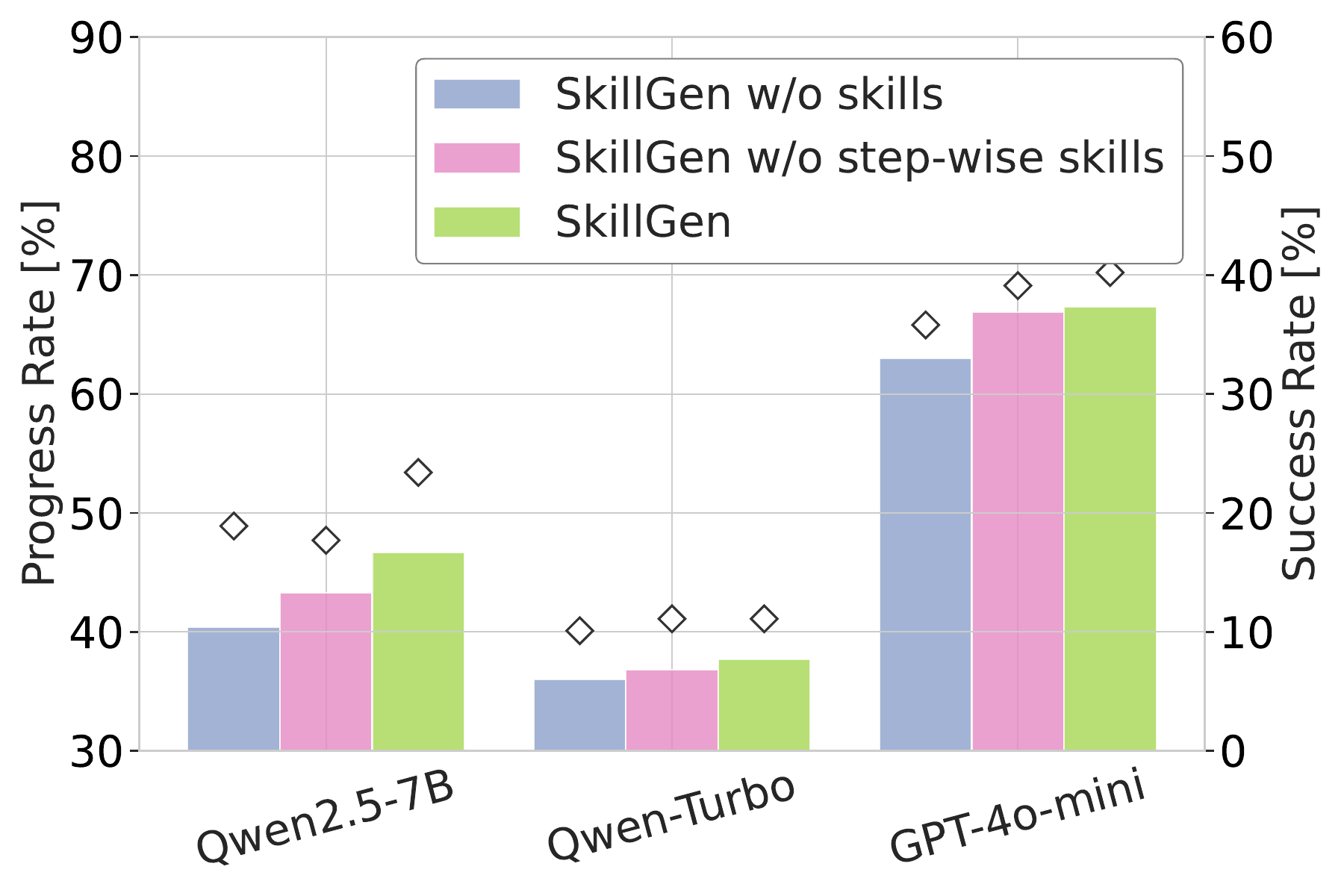}}
\vspace{-0.8em}
\caption{Ablation study of \name showing the effect of step-wise skills. Bars represent PR, $\Diamond$ markers indicate SR.\label{fig:ablation_granular}}
\vspace{-1.2em}
\end{figure}

\begin{figure}[!t]
\centering
\subfigure[ALFWorld]{
\includegraphics[width=0.32\linewidth]{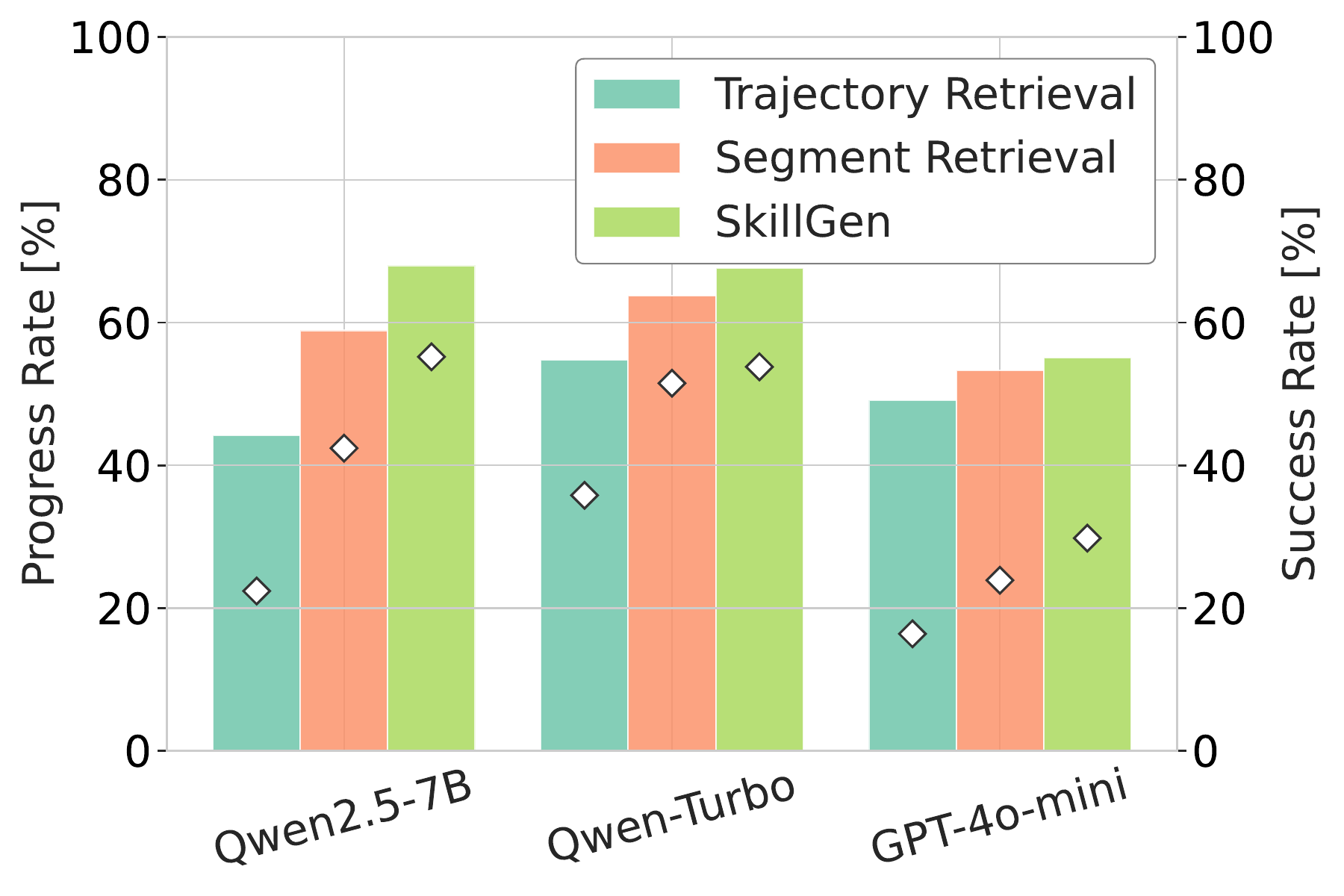}}
\subfigure[BabyAI]{
\includegraphics[width=0.32\linewidth]{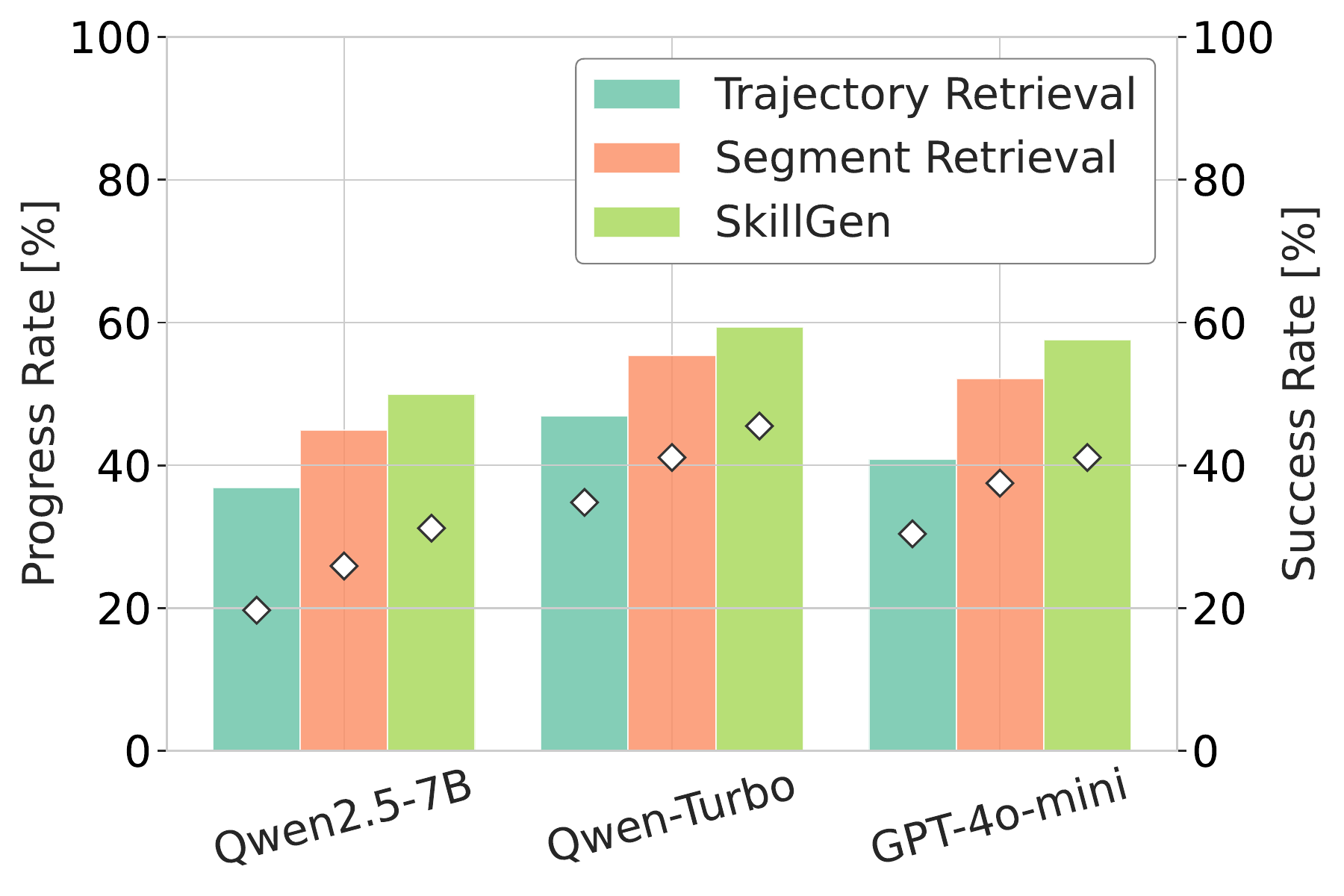}}
\subfigure[ScienceWorld]{
\includegraphics[width=0.32\linewidth]{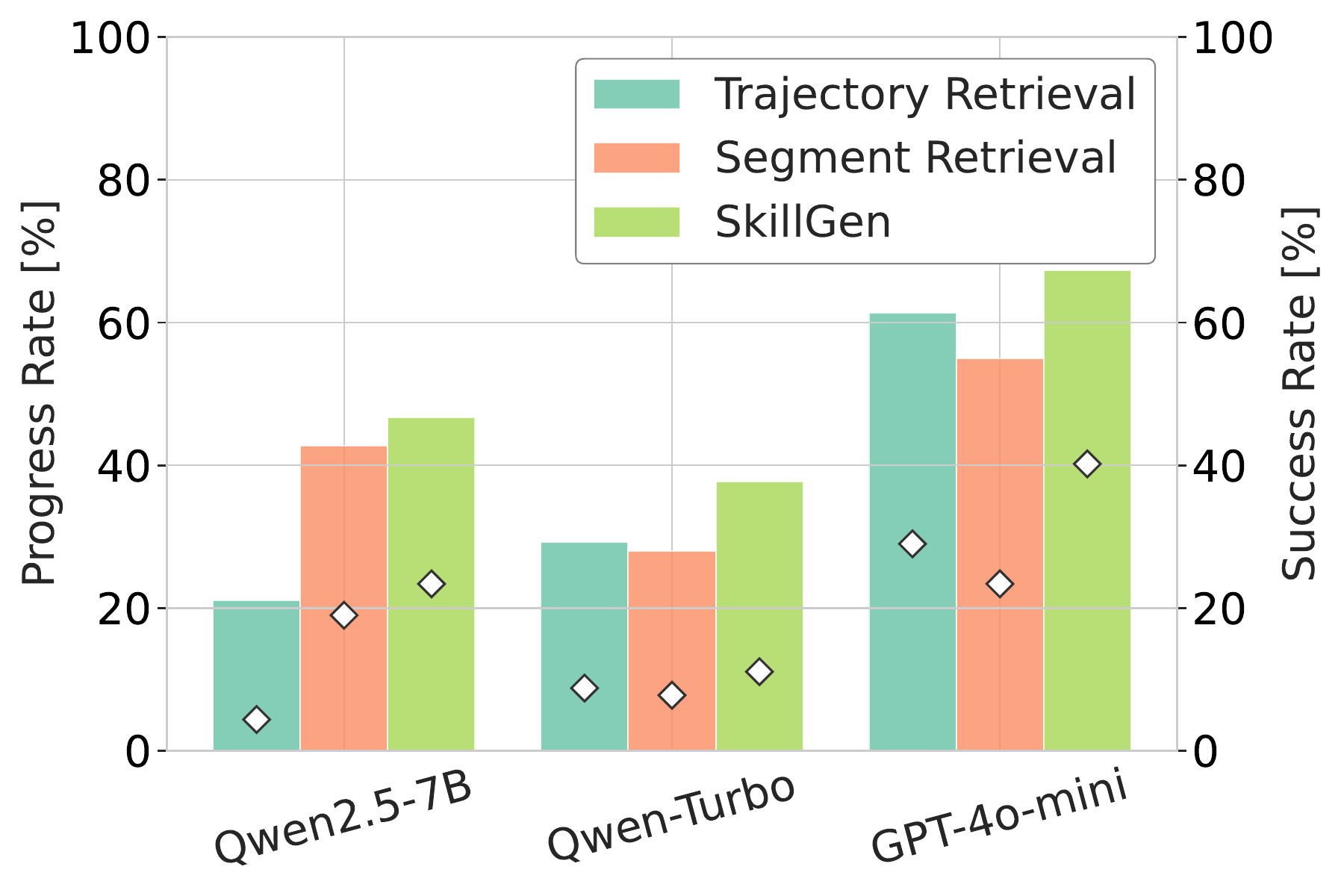}}
\vspace{-0.8em}
\caption{Ablation study of \name showing the effect of TD-based credit estimation. Bars represent PR, $\Diamond$ markers indicate SR.\label{fig:ablation_td}}
\end{figure}


\paragraph{Effect of Golden Segment.}
To validate the effectiveness of \textit{focused} prompting, we compare three strategies: (i) \textit{Fixed Demo+Skills}, which uses a static, full demonstration augmented with skills; (ii) \textit{Retrieved Demo+Skills}, which retrieves a relevant full trajectory from the training set. In Figure~\ref{fig:ablation_segment}, \textit{\name} consistently outperforms the other two strategies across all tasks and model backbones. The largest gain is observed on ALFWorld with Qwen-Turbo, where \name improves SR by +17.2\% over Fixed and +3.7\% over Retrieved. These results demonstrate that focused, high-impact segments lead to more effective decision-making by eliminating irrelevant context.

\vspace{-0.5em}
\paragraph{Effect of Step-wise Skills.} 
To assess the effect of \textit{granular} prompting, we conduct ablations comparing (i) \textit{\name w/o skills} (golden segment only), (ii) \textit{\name w/o step-wise retrieval} (naive skill injection). In Figure~\ref{fig:ablation_granular}, \name consistently achieves the best performance across all datasets and models. It improves SR over \textit{w/o skills} by +8.1\% on ALFWorld and +4.5\% on ScienceWorld, highlighting the value of structured skill integration. Step-wise retrieval further boosts SR by +3.0\% on ALFWorld and +5.7\% on ScienceWorld, showing the benefit of context-aligned guidance. While injecting all domain skills without retrieval yields moderate gains—e.g., +3.6\% PR on ScienceWorld. These results underscore that skill utility depends on contextual relevance.

\paragraph{Effect of TD-based Credit Assignment.}
To assess the effect of TD-based credit assignment, we compare three step-wise retrieval methods: (i) \textit{Trajectory Retrieval} (i.e., \textit{Trad}), which retrieves step-wise observation–action pairs; (ii) \textit{Segment Retrieval}, which retrieves action-only segments without applying credit assignment, using the original sparse progress signal. In Figure~\ref{fig:ablation_td}, \name consistently achieves the highest performance, followed by \textit{Segment Retrieval} with moderate gains. In ALFWorld, \name achieves a SR of 55.2 (Qwen2.5-7B-Instruct), outperforming \textit{Segment Retrieval} by +12.8 SR. On the more compositional ScienceWorld, \name yields substantial gains, achieving a SR of 40.2 on GPT-4o-mini—a 16.8\% increase over \textit{Segment Retrieval}. These results demonstrate that TD-based credit assignment is crucial for extracting decision-critical skills.

\begin{wraptable}{r}{8.5cm}
\centering
\vspace*{-1.5em}
\small {
\renewcommand{\tabcolsep}{2pt} 
\caption{Results of \name w/o subgoal annotations.\label{tab:ablation_subgoal}}
\vspace*{-0.8em}
\resizebox{8.5cm}{!}{
\begin{tabular}{c|cc|cc|cc} 
\toprule
\multirow{2}{*}{\textbf{Dataset}} & \multicolumn{2}{c|}{\textbf{Qwen2.5-7B}} & \multicolumn{2}{c|}{\textbf{Qwen-Turbo}} & \multicolumn{2}{c}{\textbf{GPT-4o-mini}}  \\
                                  & \textbf{PR} & \textbf{SR}                & \textbf{PR} & \textbf{SR}                & \textbf{PR} & \textbf{SR}                 \\ 
\midrule
ALFWorld                          & 63.2        & 50.1                       & 64.6        & 52.2                       & 32.9        & 14.2                        \\
BabyAI                            & 43.2        & 32.1                       & 50.5        & 36.6                       & 56.8        & 41.1                        \\
ScienceWorld                      & 29.1        & 18.9                       & 27.7        & 10.0                       & 56.2        & 32.4                        \\
\bottomrule
\end{tabular}}
}
\vspace*{-1.3em}
\end{wraptable}

\paragraph{Effect of Subgoal Annotations.}
Table~\ref{tab:ablation_subgoal} presents the results of \name without subgoal annotations, using only fully successful trajectories for skill extraction (see Appendix~\ref{appendix:additional_results} Table~\ref{tab:appendix_subgoal} for the complete results). Subgoal supervision consistently improves performance, particularly on complex tasks such as ScienceWorld, where the SR increases from 32.4\% to 40.2\% on GPT-4o-mini. Nonetheless, even without subgoal labels, \name outperforms strong baselines. For instance, on ALFWorld with Qwen2.5-7B-Instruct, \name raises the success rate from 19.4\% (\textit{Synapse 3-shot}) to 50.1\%, while on ScienceWorld, it improves PR from 29.1\% to 46.7\%. These results demonstrate that \name’s interaction-driven skill extraction enables effective generalization across domains, regardless of the presence of explicit subgoal supervision.



\begin{figure*}[!t]
\centering
\subfigure[ALFWorld]{
\includegraphics[width=0.235\linewidth]{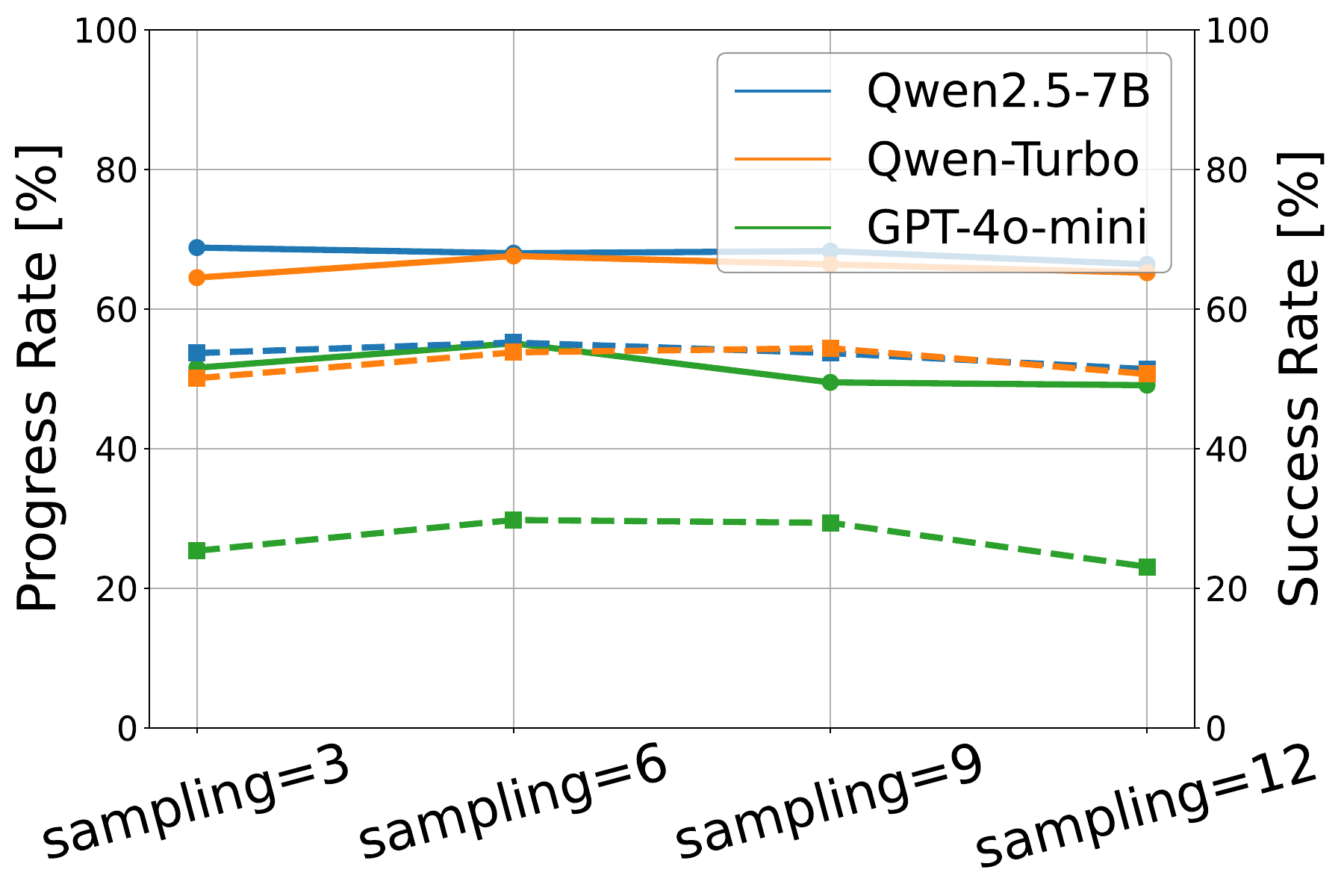}}
\subfigure[ScienceWorld]{
\includegraphics[width=0.235\linewidth]{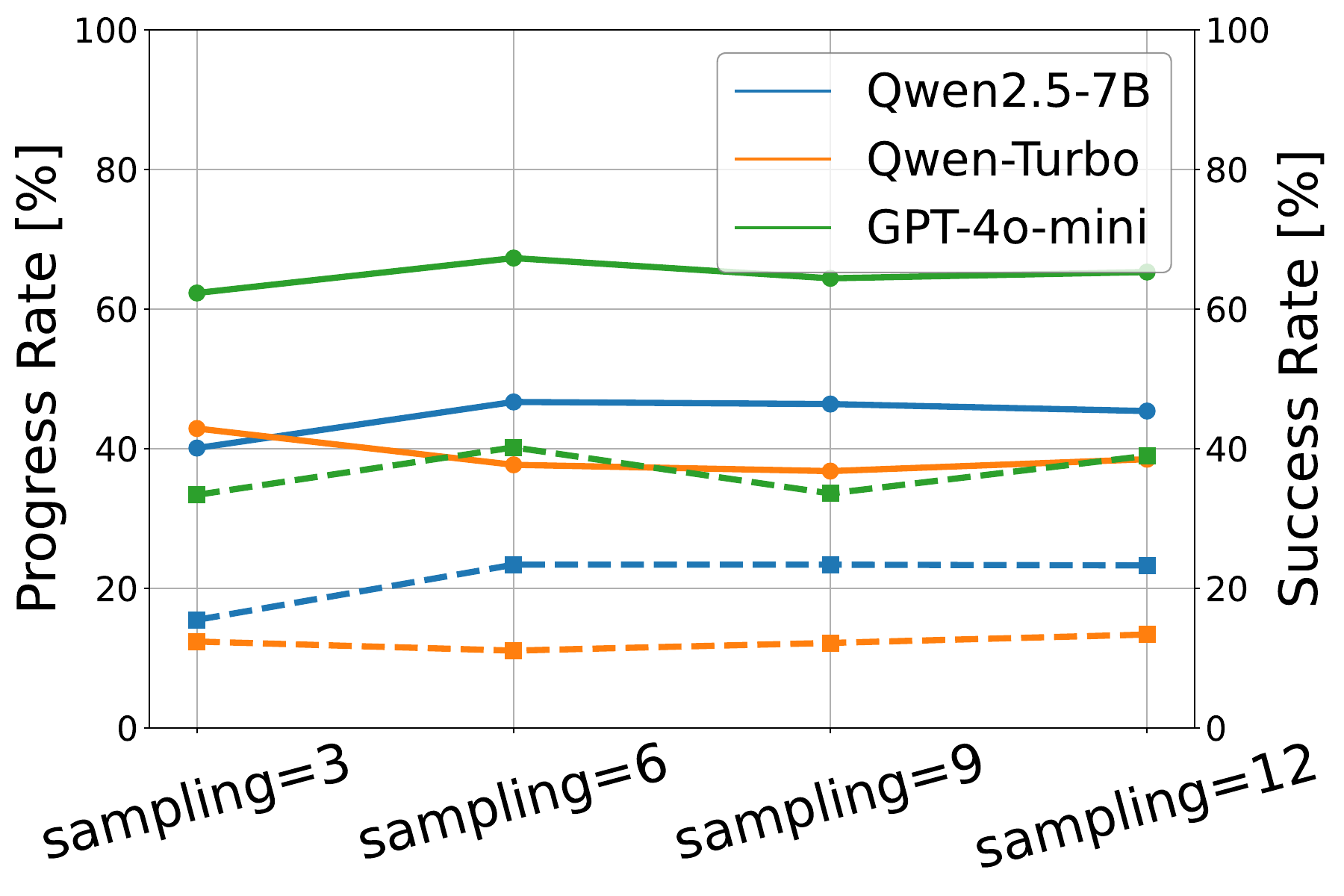}}
\subfigure[ALFWorld]{
\includegraphics[width=0.235\linewidth]{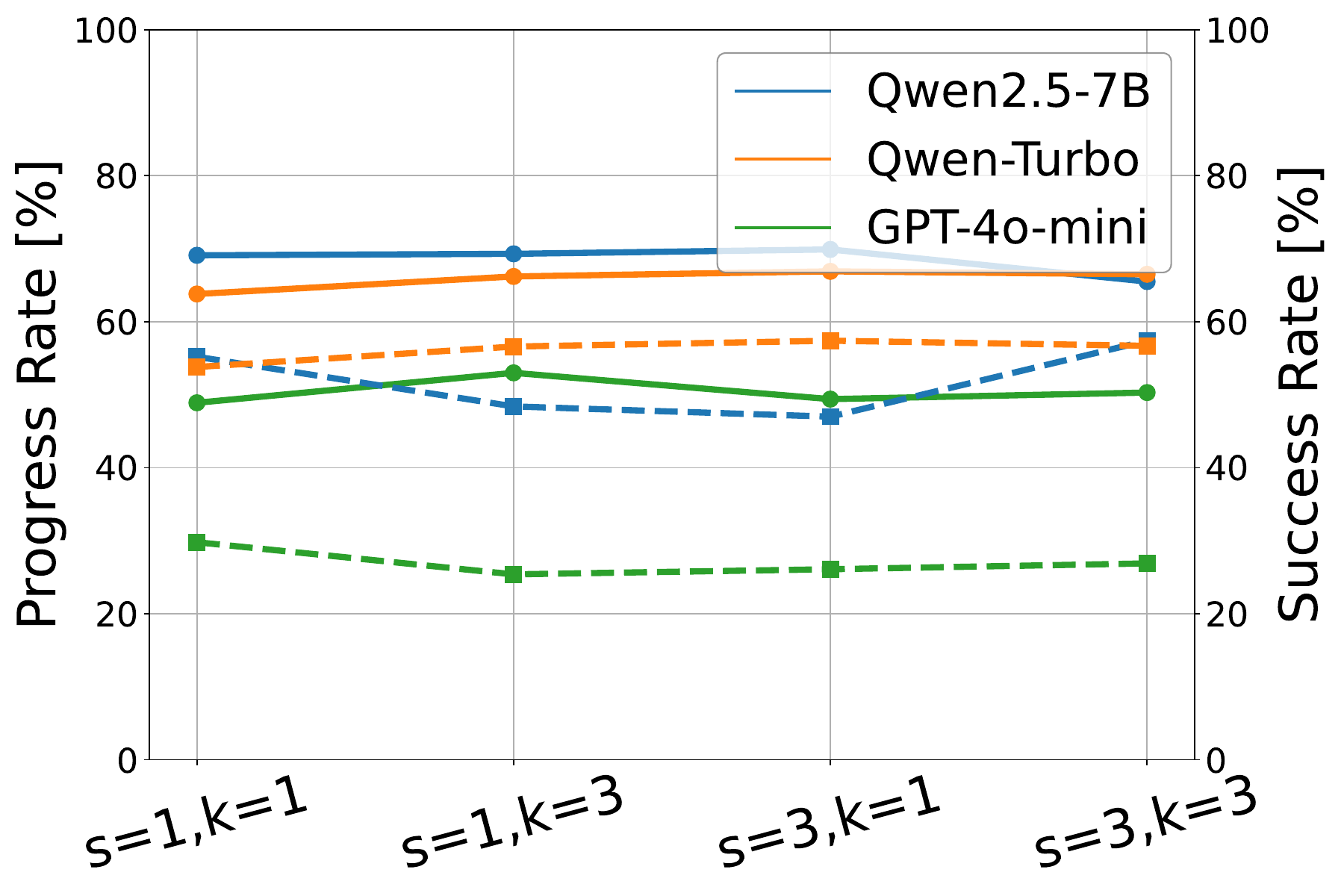}}
\subfigure[ScienceWorld]{
\includegraphics[width=0.235\linewidth]{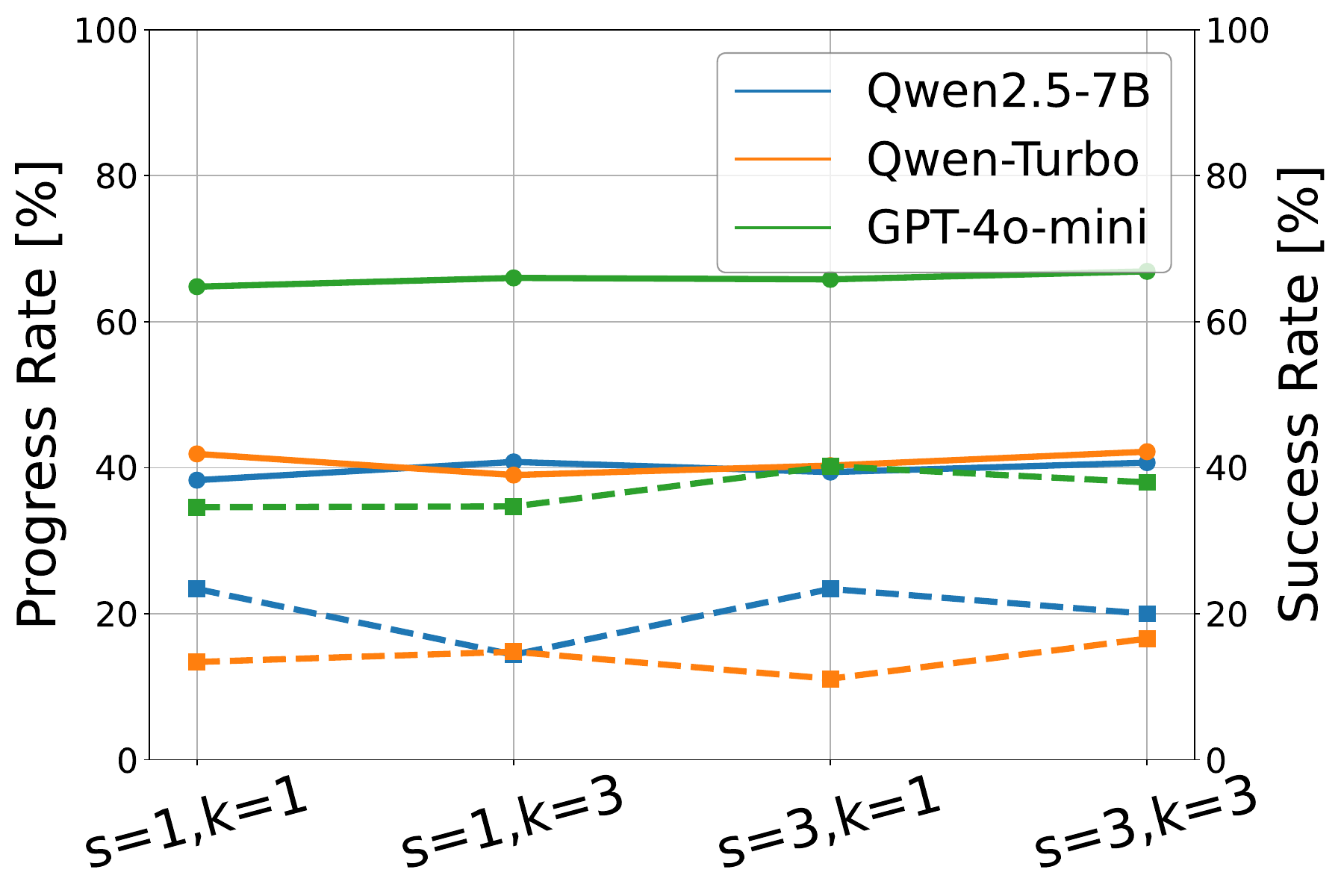}}
\caption{(a) (b): Sensitivity analysis of \name with respect to the number of sampled trajectories per task for training set.\label{fig:sensitivity_sampling} (c) (d): Sensitivity analysis of \name with respect to the number of retrieved skills $(s)$ and the number of antecedents (or consequences) per skill $(k)$.\label{fig:sensitivity_retrieve}}
\end{figure*}

\subsection{Parameter Sensitive Analysis}
\paragraph{Sampling Scale.}
To assess \name's robustness to the sampling scale, we perform sensitivity analysis with varying numbers of sampled trajectories per task. In Figure~\ref{fig:sensitivity_sampling} (a) (b), performance remains stable across settings. The minimal configuration (\textit{sampling=3}) already achieves strong SR—e.g., 53.7\% on ALFWorld (Qwen2.5-7B-Instruct). While increasing to \textit{sampling=6} often yields further gains, larger values (\textit{sampling=9,12}) may introduce noise, occasionally leading to slight drops. This suggests that a small, diverse set of trajectories suffices to extract effective skills, supporting robust performance without exhaustive sampling.

\paragraph{Skill Retrieval.}
As shown in Figure~\ref{fig:sensitivity_retrieve} (c) (d), we evaluate \name under varying retrieval settings. The minimal setting (\(s{=}1, k{=}1\)) already yields strong SR—e.g., 55.2\% on ALFWorld (Qwen2.5-7B-Instruct), 34.6\% on ScienceWorld (GPT-4o-mini). Increasing either skill count (\(k\)) or action scope (\(s\)) improves performance in complex tasks, with GPT-4o-mini reaching 40.2\% SR under (\(s{=}1, k{=}3\)). In contrast, diversity-focused settings like (\(s{=}3, k{=}1\)) help on simpler tasks, but degrade on harder ones (e.g., 14.4\% on ScienceWorld with Qwen2.5-7B-Instruct). These results highlight \name's robustness and the need to align retrieval granularity with task complexity.

\section{Related Work}
\paragraph{Sequential Decision Making with LLMs.}
Recent advances in LLM-based decision-making have led to interactive agents that operate in multi-turn loops, either selecting actions directly or reasoning before acting~\cite{yao2023react,rozanov2025stateactenhancingllmbase, zhao-etal-2025-uncertainty}. To address long-horizon tasks, many methods incorporate feedback-driven refinement~\cite{shinn2023reflexion, wang2023describe, sun2023adaplanner, chenplan} or structured search~\cite{yao2023tree, besta2024graph, hao2023reasoning, zhuang2024toolchain, wu2025avatar, li2025solverllm}. Another line of work improves inference by retrieving expert or history information from offline interactions~\cite{synapse2023, xiao2023o3d, kagaya2024rap, trad2024}. More recently, self-improving agents construct in-context examples from prior episodes, enabling generalization to unseen tasks without relying on expert demonstrations~\cite{sarukkai2025self, liu2025contextual}. These methods highlight a growing emphasis on experience-driven decision-making.

\paragraph{Knowledge-Augmented In-Context Learning.} 
Knowledge augmented methods aim to enrich the prompt with structured information—such as relational or procedural knowledge, to provide stronger inductive bias and support more accurate reasoning. Some approaches guide reasoning with high-level prompts derived from prior interactions~\cite{kim2023language,zhang2024incontext,fu2024autoguide,kong2025meta}. Others inject retrieved graph-based knowledge to support multi-hop inference~\cite{mavromatis2024gnn,hu2024grag,luo2024graph}. LLMs can also self-synthesize reusable strategies from world models for generalization~\cite{ding2024everything, zhou2024agents,qiao-etal-2024-autoact,qiao2024agent}, or incorporate procedural knowledge via rule induction~\cite{zhu2023large,zhang2025ruag} and skill reuse~\cite{zhu2024knowagent, chen2024automanual,zhang2023bootstrap,expel2024}. These approaches enrich in-context learning by integrating external or derived task-specific knowledge as decision support.

\section{Conclusion}
In this study, we explore how to improve in-context learning for sequential decision-making by tackling three key challenges: maintaining decision focus, offering granular guidance, and reducing dependence on expert supervision. To address these issues, we propose \name, a skill-aware prompting framework that leverages structured knowledge and weak supervision to enable fine-grained, context-sensitive reasoning. Our theoretical analysis highlights how decision-critical content supports task identifiability, motivating more principled prompt design. 
Experiments on ALFWorld, BabyAI, and ScienceWorld show that \name delivers consistent gains without expert demonstrations. We believe our findings offer a foundation for future research on skill-aware prompting to improve generalization, efficiency, and coherence in LLM-based decision-making.

\section*{Acknowledgments}
Ruomeng Ding, Wei Cheng, and Chen Zhao did not receive any financial support for this work, and Wei Cheng and Chen Zhao contributed only by developing the research ideas, participating in discussions, and providing feedback on the manuscript.

\bibliographystyle{unsrt}
\bibliography{ref}

\newpage
\appendix

\begin{figure*}[!t]
\begin{center}
\includegraphics[trim=60 184 60 145, clip, width=0.85\textwidth]{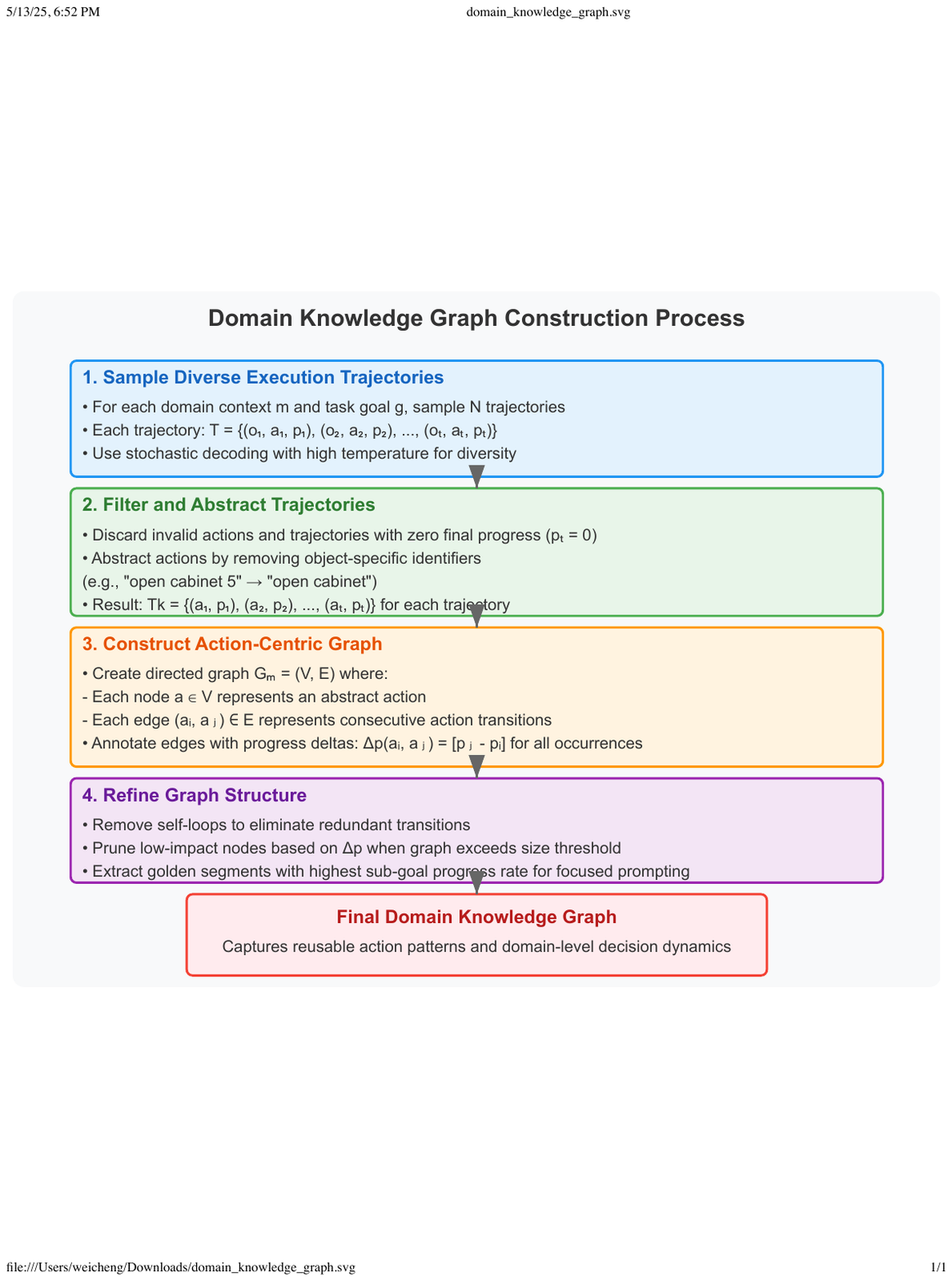}
\end{center}
\caption{Pipeline of Domain Graph Construction}
\label{Fig:graph}
\end{figure*}

\section{Limitations\label{appendix:limitations}}

While \name demonstrates strong performance across diverse tasks and generalizes across both proprietary and open-source LLMs, several limitations remain. \name relies on sampled trajectories with subgoal or progress feedback, which may be unavailable in domains lacking structured intermediate signals. A common workaround is to filter for fully successful trajectories, but this depends on sufficient high-quality samples. Increasing the sampling scale may help, but often yields diminishing returns in quality and incurs high computational cost, while still being constrained by the base model's capabilities. Additionally, constructing domain-level action graphs introduces preprocessing overhead and may not scale well in extensive or dynamic action spaces. Retrieved skills, grounded in local action contexts, may also struggle to generalize to out-of-distribution scenarios or novel task configurations. These limitations do not undermine our core contributions, but point to future directions for enhancing the scalability and robustness of \name.


\section{Additional Methodological Details\label{appendix:method}}

\subsection{Domain Knowledge Graph Construction}\label{appendix:graph_construction}
In Figure~\ref{Fig:graph}, we illustrate the pipeline for constructing the domain knowledge graph, including four stages: (1) sampling diverse execution trajectories, (2) filtering and abstracting actions, (3) constructing an action-centric graph, and (4) refining the graph structure. After obtaining the final domain knowledge graph, we use it to perform TD-based action credit assignment, which enables the model to learn utility-aware skill representations for downstream decision-making tasks.

\subsection{TD-based Credit Assignment\label{appendix:td_estimation}}

In Algorithm~\ref{alg:td_credit_assignment}, we introduce a TD($\lambda$)-based credit assignment procedure over a domain-level action graph. The algorithm repeatedly samples action paths, estimates noisy progress-based rewards, and applies temporal-difference learning with eligibility traces to assign utility-aware scores to abstract actions. Unlike standard TD($\lambda$), which resets traces at the start of each episode, we persist action-level traces across sampled paths to reinforce globally useful action patterns. To prevent unbounded accumulation, we apply exponential decay at each step via \( E(a) \leftarrow \gamma \cdot \lambda \cdot E(a) \). This design draws inspiration from prior work on learning shared action values~\cite{chandak2019learning}, and enables the definition of a state-agnostic action-value function \( Q(a) \) that captures the expected long-term utility of each action.

\begin{algorithm}[!t]
\caption{TD-Based Action Credit Assignment on Refined Domain Knowledge Graph}
\label{alg:td_credit_assignment}
\begin{algorithmic}[1]
\Require Refined domain graph $\mathbb{G}_m = (\mathbb{V}, \mathbb{E})$ with progress annotations $\mathcal{P}_\Delta(a_i, a_j)$; TD iterations $N$; batch size $B$; learning rate $\alpha$; discount factor $\gamma$; trace decay $\lambda$
\Ensure Normalized action credit scores $\bar{Q}(a)$ for all $a \in \mathbb{V}$

\State $Q(a) \gets 0$, $E(a) \gets 0$ \quad for all $a \in \mathbb{V}$ \Comment{Initialize Q-values and eligibility traces for all actions}

\For{$i = 1$ to $N$}
    \State $\{\tau^{(1)}, \dots, \tau^{(B)}\} \sim \mathcal{S}_{\text{path}}^{B}(a_{\text{start}}, a_{\text{end}}, \mathbb{G}_m)$ \Comment{Sample a batch of $B$ paths from $a_{\text{start}}$ to $a_{\text{end}}$}
    
    \For{each path $\tau = [a_0, a_1, \dots, a_T]$ in the batch}
        \For{$t = 0$ to $T - 1$}
            \Comment{Sample noisy reward from empirical progress delta}
            \If{$\mathcal{P}_\Delta(a_t, a_{t+1}) \ne \emptyset$}
                \State $r_t \sim \mathrm{Uniform}(\mathcal{P}_\Delta(a_t, a_{t+1})) + \mathcal{N}(0, \sigma^2)$
            \Else
                \State $r_t \sim \mathcal{N}(0, \sigma^2)$
            \EndIf

            \State $\delta_t \gets r_t + \gamma Q(a_{t+1}) - Q(a_t)$ \Comment{Compute TD error and update eligibility traces}
            \State $E(a_t) \gets E(a_t) + 1$ 
            
            \For{each $a \in \mathbb{V}$ such that $E(a) > 0$}
                \State $Q(a) \gets Q(a) + \alpha \cdot \delta_t \cdot E(a)$
                \State $E(a) \gets \gamma \cdot \lambda \cdot E(a)$
            \EndFor
        \EndFor
    \EndFor
\EndFor
\State $\bar{Q}(a) \gets \dfrac{\max(Q(a), 0)}{\sum_{a' \in \mathbb{V}} \max(Q(a'), 0)} \quad \text{for all } a \in \mathbb{V}$ \Comment{Normalize Q to obtain action credit scores}

\Return $\{\bar{Q}(a)\}_{a \in \mathbb{V}}$
\end{algorithmic}
\end{algorithm}

\subsection{Skill-Based In-Context Learning\label{appendix:skill_ICL}}

In Algorithm~\ref{alg:retrieval_prompting}, we present a skill-based prompting strategy that guides a frozen LLM using both offline domain knowledge and online interaction history. At each time step, the agent retrieves relevant step-wise skills based on the most recent action, combines them with a domain-level golden segment, and assembles a structured prompt that integrates retrieved knowledge with the current trajectory context. The resulting prompt, serialized in natural language, enables credit-guided, context-aware reasoning without any parameter updates to the model.

\begin{algorithm}[!t]
\caption{Skill-Based ICL for Sequential Decision Making}
\label{alg:retrieval_prompting}
\begin{algorithmic}[1]
\Require Domain skill graph $\mathbb{G}_m$, golden segment $\Phi_{\text{gold}}$, frozen LLM, pretrained retriever $\mathrm{Retrieve}(\cdot)$
\Ensure Action sequence $\{a_1, a_2, \dots, a_T\}$ generated via prompt-based inference

\State {Initialize empty trajectory history $h_0 \gets \emptyset$}
\For{$t = 1$ to $T$}
    \State Observe current transition history $h_t = \{(o_0, a_0), \dots, (o_{t-1}, a_{t-1}), o_t\}$

    \State $\hat{\mathcal{A}} \gets \mathrm{Retrieve}(a_{t-1})$ from $\mathbb{G}_m$ \Comment{Top-$k$ relevant actions based on $a_{t-1}$}
    
    \State $\Phi_{\text{skill}} \gets \bigoplus_{a \in \hat{\mathcal{A}}} \mathrm{Skill}(a)$ \Comment{Concatenate retrieved skills}
    
    \State $\Phi \gets \Phi_{\text{gold}} \oplus \Phi_{\text{skill}}$ \Comment{Assemble multi-level contextual knowledge}
    
    \State $\mathrm{Prompt}_t \gets \mathrm{Prompt}(h_t, \Phi)$ \Comment{Format prompt with history and retrieved context}
    
    \State $a_t \sim \mathrm{LLM}(a \mid \mathrm{Prompt}_t)$ \Comment{Sample next action from LLM}
\EndFor
\State \Return Action sequence $\{a_1, a_2, \dots, a_T\}$
\end{algorithmic}
\end{algorithm}

\section{Theoretical Analysis} \label{appendix:theory}

We adopt the latent task mixture framework of Wies et al.~\cite{wies2023learnability}, where prompts are sampled from a mixture distribution \(\mathcal{D} = \sum_{\phi \in \Phi} \pi(\phi) P_\phi\), with \(\phi \in \Phi\) indexing latent tasks, \(\pi(\phi)\) the prior, and \(P_\phi\) the task-specific sequence distribution. The following assumptions are used in our analysis.

\begin{assumption}[Approximate Step-wise Independence] \label{assump:indep}
There exists a constant \( 0 < c_1 \le 1 \) such that for any two strings \(s_1, s_2 \in \Sigma^\star\) and any concept \(\phi\), we have:
\begin{equation}
c_1 \le \frac{P_\phi(s_1 \oplus \texttt{``\textbackslash n''}) \cdot P_\phi(s_2)}{P_\phi(s_1 \oplus \texttt{``\textbackslash n''} \oplus s_2)} \le \frac{1}{c_1}.
\end{equation}
\end{assumption}

This assumption implies that the prompt likelihood approximately factorizes over segments: $P_\phi(p) \approx \prod_{t=1}^T P_\phi(x_t, y_t)$, up to a bounded multiplicative deviation, due to the string-level composition bounds imposed on concatenated segments.

\begin{assumption}[Global Token Probability Lower Bound] \label{assump:global-lower}
There exists a constant \( c_2 > 0 \) such that for all \(\phi \in \Phi\), all inputs \(x_t\), and all output tokens \(y_t\), the conditional probability is bounded below:
\begin{equation}
P_\phi(y_t \mid x_t) > c_2.
\end{equation}
\end{assumption}
This assumption ensures that the KL divergence between task distributions is finite and avoids vanishing likelihoods.

\begin{assumption}[Uniform Residual Token Bias] \label{assump:residual-uniform}
For all \(\phi \ne \phi^\star\) and all \((x_t, y_t) \in p_{\text{irrelevant}}\), we have:
\begin{equation}
P_\phi(y_t \mid x_t) \ge P_{\phi^\star}(y_t \mid x_t).
\end{equation}
\end{assumption}
This assumption states that irrelevant segments are at least as likely to appear under alternative tasks as under the true task. In other words, they are no more informative about the true task than about any other, and therefore provide no useful signal for identifying the correct task. 

\begin{lemma}[Task Likelihood Separation~{\cite[Lemma 1]{wies2023learnability}}]\label{lemma:wies}
Let \(\mathcal{D} = \sum_{\phi \in \Phi} \pi(\phi) P_\phi\) be a task mixture distribution satisfying Assumptions 1 and 2 . Let \(\phi^\star \in \Phi\) be the ground-truth task. Then, there exists a sample complexity threshold \(\tilde{m}_\mathcal{D} : (0, 1)^2 \rightarrow \mathbb{N}\) such that for any \(\epsilon, \delta > 0\), if the number of in-context examples \(k \ge \tilde{m}_\mathcal{D}(\epsilon, \delta)\), we have:
\begin{equation}
\mathbb{P}_{p \sim P_{\phi^\star}^{\otimes k}} \left[
\forall \phi \ne \phi^\star,\;
\frac{P_\phi(p)}{P_{\phi^\star}(p)} < \epsilon
\right] \ge 1 - \delta.
\end{equation}
\end{lemma}

Lemma~\ref{lemma:wies} guarantees task separability using the full prompt, we now show that task identifiability can be preserved using only the focused portion of the prompt.

\paragraph{Proof of Theorem 1}
\begin{proof}
By Lemma~\ref{lemma:wies}, under Assumptions 1 and 2, there exists a sample complexity threshold \(\tilde{m}_\mathcal{D}(\epsilon, \delta)\) such that for any \(k \ge \tilde{m}_\mathcal{D}(\epsilon, \delta)\), with probability at least \(1 - \delta\) (over the sampling of \(p \sim P_{\phi^\star}^{\otimes k}\)), we have:
\begin{equation}
\forall \phi \ne \phi^\star, \quad
\frac{P_\phi(p)}{P_{\phi^\star}(p)} < \epsilon.
\end{equation}

Since the prompt decomposes as \(p = p_{\text{focused}} \cup p_{\text{irrelevant}}\), we can factor the likelihood ratio as:
\begin{equation}
\frac{P_\phi(p)}{P_{\phi^\star}(p)} =
\frac{P_\phi(p_{\text{focused}})}{P_{\phi^\star}(p_{\text{focused}})} \cdot
\frac{P_\phi(p_{\text{irrelevant}})}{P_{\phi^\star}(p_{\text{irrelevant}})}.
\end{equation}

By Assumption 3, we have that for all \(\phi \ne \phi^\star\) and for each \((x_t, y_t) \in p_{\text{irrelevant}}\),
\begin{equation}
P_\phi(y_t \mid x_t) \ge P_{\phi^\star}(y_t \mid x_t).
\end{equation}

Assuming conditional independence across segments within \(p_{\text{irrelevant}}\), we can write:
\begin{equation}
P_\phi(p_{\text{irrelevant}}) = \prod_{(x_t, y_t) \in p_{\text{irrelevant}}} P_\phi(y_t \mid x_t),
\quad
P_{\phi^\star}(p_{\text{irrelevant}}) = \prod_{(x_t, y_t) \in p_{\text{irrelevant}}} P_{\phi^\star}(y_t \mid x_t).
\end{equation}

Combining these, we obtain:
\begin{equation}
P_\phi(p_{\text{irrelevant}}) \ge P_{\phi^\star}(p_{\text{irrelevant}}),
\quad \Rightarrow \quad
\frac{P_\phi(p_{\text{irrelevant}})}{P_{\phi^\star}(p_{\text{irrelevant}})} \ge 1.
\end{equation}

This shows that the irrelevant portion of the prompt weakly favors alternative tasks or, at best, contributes no additional evidence toward the identification of \(\phi^\star\). Next, we recall that the full prompt likelihood factorizes over the focused and irrelevant parts:
\begin{equation}
P_\phi(p) = P_\phi(p_{\text{focused}}) \cdot P_\phi(p_{\text{irrelevant}}),
\quad
P_{\phi^\star}(p) = P_{\phi^\star}(p_{\text{focused}}) \cdot P_{\phi^\star}(p_{\text{irrelevant}}).
\end{equation}

Taking the ratio, we have:
\begin{equation}
\frac{P_\phi(p)}{P_{\phi^\star}(p)} = 
\frac{P_\phi(p_{\text{focused}})}{P_{\phi^\star}(p_{\text{focused}})} \cdot
\frac{P_\phi(p_{\text{irrelevant}})}{P_{\phi^\star}(p_{\text{irrelevant}})}.
\end{equation}

From the inequality above, we know:
\begin{equation}
\frac{P_\phi(p_{\text{irrelevant}})}{P_{\phi^\star}(p_{\text{irrelevant}})} \ge 1,
\end{equation}

which implies:
\begin{equation}
\frac{P_\phi(p)}{P_{\phi^\star}(p)} \ge 
\frac{P_\phi(p_{\text{focused}})}{P_{\phi^\star}(p_{\text{focused}})}.
\end{equation}

Hence, for all \(\phi \ne \phi^\star\), we conclude:
\begin{equation}
\frac{P_\phi(p_{\text{focused}})}{P_{\phi^\star}(p_{\text{focused}})} \le
\frac{P_\phi(p)}{P_{\phi^\star}(p)} < \epsilon.
\end{equation}
\end{proof}

\section{Experimental Setup\label{appendix:setup}}

\subsection{Dataset}

\begin{table*}[!t]
\centering
\caption{Task Datasets~\label{tab:dataset}}
\resizebox{\textwidth}{!}{
\begin{tabular}{lccccccc}
    \toprule
    \textbf{Dataset} & \textbf{Task Type} & \textbf{\# Environment} & \textbf{\# Turns} & \textbf{Action Space}  & \textbf{\# Subgoal Count} \\
    \midrule
    ALFWorld & Household Tasks & 134 & 6 & 13  & 3 \\
    BabyAI & Grid-based Navigation & 112 & 10 & 8 & 4 \\
    ScienceWorld & Scientific Reasoning & 90 & 15 & 21 & 5 \\
    \bottomrule
\end{tabular}}
\end{table*}

As shown in Table~\ref{tab:dataset}, we evaluate our method on three sequential decision-making benchmarks of increasing complexity: ALFWorld, BabyAI, and ScienceWorld. These datasets span diverse domains—ranging from household tasks to scientific experiments—and vary in interaction length, action space size, and subgoal structure, presenting a progressively harder challenge from ALFWorld to ScienceWorld. To enable robust and fair evaluation across diverse sequential decision-making domains, we adopt a 4-fold cross-validation protocol for all three datasets. For each dataset, the full set of task instances is randomly partitioned into four equal subsets, with shuffling enabled and a fixed random seed (42) to ensure reproducibility. In each fold, one subset is held out as the test set while the remaining three subsets are used for training. This procedure is repeated four times, allowing every task instance to serve as a test sample exactly once. This cross-validation scheme allows us to systematically assess the generalization capability of our method under varying task compositions and minimizes the risk of overfitting to a particular subset of tasks.

\paragraph{ALFWorld.} 
It is an embodied task in simulated household environments, testing agents on exploration, object manipulation, and commonsense reasoning through text-based interaction. It contains 134 tasks across various home settings (e.g., kitchen, bedroom, bathroom), involving objectives like \textit{"put two bars of soap into the trash can"} or \textit{"place the clean egg into the microwave"}. The agent interacts with the environment using natural language actions such as “open the fridge” or “take the egg”, and must interpret textual feedback to make decisions. On average, tasks require 6 interaction steps, with an action space of 13 and an average context length of approximately 900 tokens. In addition to the high-level goal, each task is annotated with a sequence of subgoals that define intermediate progress checkpoints. These subgoals correspond to specific, verifiable behaviors—such as locating an object, picking it up, or transforming its state (e.g., cleaning or cooling)—and serve as weak supervision signals during training. For example, the goal \textit{"put a clean soapbar in countertop"} is decomposed into subgoals: (1) observe a soapbar, (2) pick it up, and (3) clean it. These subgoals are typically specified using regular expressions or fixed language templates matched against textual feedback from the environment.

\paragraph{BabyAI.} 
It evaluates agents in a 20×20 partially observable grid world, focusing on navigation and object interaction. The dataset includes 120 tasks, where agents perceive only their local surroundings and must act based on partial information. Observations are rendered as text (e.g., \textit{"a wall is two steps ahead"}), and actions are also issued as textual commands (e.g., \textit{“go to red ball 1”}, \textit{"open and go through green locked door 1"}). Tasks typically span 10 steps, with an action space of 8 and an average context length of around 1800 tokens—longer than ALFWorld but shorter than ScienceWorld—suggesting a moderate level of language and planning complexity. Each task is defined by a high-level goal (e.g., \textit{"go to the red ball"}), along with a set of subgoals that encode verifiable intermediate signals. These subgoals reflect localized perceptual conditions or agent proximity to targets, such as observing an object or aligning with its position. For instance, a task like \textit{"go to the red ball"} might include subgoals like (1) “There is a red ball \*” and (2) "There is a red ball \* right in front of you 1 steps away." These expressions are matched against observation strings during execution, providing weak supervision that allows for fine-grained progress tracking and temporal credit assignment.

\paragraph{ScienceWorld.} 
It provides a rich environment for scientific reasoning, consisting of 90 tasks across 8 rooms (e.g., lab, kitchen, workshop), each equipped with domain-specific tools like thermometers, beakers, and heaters. Tasks involve following experimental procedures and making inferences based on observations, such as \textit{"measure the melting point of orange juice"} or \textit{"observe microbial structures"}. Agents must gather information through interaction and plan a coherent sequence of scientific steps. Tasks average 15 interaction steps, with an action space of 21 and a significantly longer context length of up to 2800 tokens, reflecting the increased complexity of reasoning and language understanding required. Each task is annotated with a sequence of subgoals, which represent semantically intermediate progress. For instance, a task like \textit{"measure the melting point of gallium"} might involve the following subgoals: (1) move the thermometer to the inventory, (2) transfer gallium to a container, (3) apply heat using the stove, (4) wait until a temperature threshold is reached, and (5) focus on the correct output box based on the result. These subgoals serve as weak supervision for learning scientific reasoning progression.

\subsection{Evaluation Metrics}

Across all experiments in this paper, we evaluate each method on the test set using the metrics below: 
\paragraph{Grounding Rate.}  
Grounding Rate measures whether the agent’s generated action is valid given the current environment state. It reflects the agent’s ability to align textual output with the environment’s affordances, indicating its perceptual understanding and language grounding capabilities. It is defined as:
\begin{equation}
\text{Grounding Rate} = \frac{\sum_{t=1}^{T} \mathbbm{1}(\text{IsValid}(a_t, \text{Env}_t))}{T},
\end{equation}
where $ T $ is the total number of interaction steps, $ a_t $ is the agent’s action at step $ t $, and $ \text{Env}_t $ is the environment state at that step. The indicator function $ \mathbbm{1}(\text{IsValid}(a_t, \text{Env}_t)) $ returns 1 if the action is valid in the given environment state, and 0 otherwise. Grounding Rate ranges from 0 to 1, with higher values indicating better alignment between the agent’s actions and the environment.

\paragraph{Progress Rate.}  
Progress Rate measures the proportion of subgoals successfully completed during task execution, providing a fine-grained indicator of overall progress. Given a task defined by a set of subgoals \( G \), we compute:
\begin{equation}
\text{Progress Rate} = \frac{\sum_{g \in G} \mathbbm{1}\left(\exists\, t \le T \text{ s.t. } \text{achieved}(s_t, g)\right)}{|G|},
\end{equation}
where \( \text{achieved}(s_t, g) \) indicates whether subgoal \( g \) is satisfied at step \( t \), and \( |G| \) is the total number of subgoals. The indicator function returns 1 if the subgoal is achieved at any step. The score ranges from 0 to 1, with higher values indicating greater task completion.

\paragraph{Success Rate.}  
Success Rate is a binary indicator of task completion: it evaluates whether the agent successfully achieves all subgoals by the end of the inference. Unlike Progress Rate, which allows partial credit, Success Rate requires full satisfaction of the task objective:
\begin{equation}
\text{Success Rate} = \mathbbm{1}\left( \forall\, g \in G,\ \exists\, t \le T\ \text{s.t. } \text{achieved}(s_t, g) \right),
\end{equation}
where \( G \) is the set of subgoals defining the task, and \( s_t \) is the agent’s state at step \( t \). The indicator returns 1 only if all subgoals are achieved during inference. This score is either 0 or 1 for each task instance.

\paragraph{Area Under the Progress Curve (AUPC).}
Let a task trajectory be represented as a sequence of progress observations:
\begin{equation}
\mathcal{P} = \{(s_0, p_0), (s_1, p_1), \dots, (s_n, p_n)\},
\end{equation}
where \( s_i \in \mathbb{N} \) is the time step and \( p_i \in [0, 1] \) is the corresponding progress value. AUPC quantifies how efficiently an agent accumulates progress over time and is computed using the trapezoidal rule:
\begin{equation}
\text{RawAUPC}(\mathcal{P}) = \sum_{i=1}^{n} \frac{p_{i-1} + p_i}{2} \cdot (s_i - s_{i-1}).
\end{equation}

The raw area is normalized by the total number of steps, ensuring scale-invariant progress comparison:
\begin{equation}
\text{AUPC}(\mathcal{P}) = 
\begin{cases}
\frac{\text{RawAUPC}(\mathcal{P})}{s_n - s_0} & \text{if } s_n > s_0, \\
0 & \text{otherwise}.
\end{cases}
\end{equation}

A higher AUPC indicates more efficient and consistent task progress, rewarding agents that make early and sustained advancements throughout inference. It yields a scalar score in the range [0,1], reflecting both the speed and smoothness of progress.

\subsection{Implementation Details}

\vspace{-0.5em}
\paragraph{Sampling Details.}
We construct the training data by sampling multiple trajectories per task instance using stochastic decoding. The main results are based on 6 sampled trajectories per instance, while 3, 9, and 12 are evaluated in a sensitivity analysis to assess the effect of sampling count. We use temperature-controlled decoding with a default temperature of 1.0 to encourage diversity in action sequences. Lower temperatures (0.3 and 0.7) are included in ablation studies to investigate the impact of more deterministic sampling behavior. All trajectories are truncated to a maximum of 10 steps across datasets (ALFWorld, BabyAI, ScienceWorld) to maintain consistency and reduce variance in long-horizon planning.

\vspace{-0.5em}
\paragraph{Baseline Details.}
For all baselines, inference is run for up to 20 steps using deterministic decoding (temperature = 0). Each prompt includes a general instruction, contextual knowledge (e.g., demonstrations or insights), and the task’s step-wise interaction history. The history window is also set to 20, so no past information is discarded. In the \textit{0-shot} setting, the prompt contains only the instruction and history. The \textit{1-shot} baseline adds a handcrafted demonstration~\cite{ma2024agentboard}, using one shared demo for BabyAI and ScienceWorld, and six domain-specific demos for ALFWorld (one per domain). \textit{Leap} summarizes up to 8 insights from the top-3 diverse success–failure pairs in the training set, which are distilled into general principles and included in the prompt per fold. \textit{Synapse (3-shot)} retrieves the top-3 fully successful trajectories based on task metadata similarity (goal and domain) and includes them in the prompt, while \textit{Synapse (1-shot)} includes only the top-1. \textit{Trad} retrieves the top-3 related nonzero-progress trajectories, decomposes them into (observation, action) pairs, and performs step-wise retrieval. At each step, it selects 3 relevant segments based on the current observation using a sliding window of size 3 (previous, current, and next).

\vspace{-0.5em}
\paragraph{\name Details.}
During domain graph construction, self-loops are removed before and after pruning. The number of nodes is capped at 30 to avoid extreme cases. If exceeded, intermediate nodes are ranked by the average progress of their incoming edges, and the lowest-ranked are iteratively removed. In temporal-difference (TD) learning, we set \(\gamma = 0.95\), \(\lambda = 0.9\), \(\alpha = 0.05\), and reward noise \(\sigma = 0.001\). Q-values are initialized uniformly in \([0.01, 0.05]\). We sample up to 2{,}000 simple paths (maximum length 20) and run TD updates for 500 iterations, with early stopping if the average Q-value change falls below \(10^{-3}\) for five consecutive steps. An analysis of the domain graphs is provided for reference in Appendix~\ref{appendix:domin_graph}. For skill extraction stage, the domain-level golden segment is selected as the trajectory with the highest total progress after filtering out invalid steps. For skill retrieval, we set \(s\) as the number of skills and \(k\) as the number of surrounding steps (antecedents and consequences) per skill. We use \((s{=}1, k{=}1)\) for ALFWorld and BabyAI, and \((s{=}1, k{=}3)\) for ScienceWorld due to its higher complexity.

\vspace{-0.5em}
\paragraph{Computational Resources.}
All experiments were conducted on a multi-GPU server equipped with 4× NVIDIA RTX A6000 GPUs (49GB each; CUDA 12.7; Driver 565.57.01) and an 80-core Intel Xeon Platinum processor (2 NUMA nodes). Open-source models such as Qwen2.5-7B-Instruct were run locally using HuggingFace Transformers v4.51.3 and PyTorch v2.6.0. Inference with proprietary models—Qwen-Turbo and GPT-4o-mini—was performed via step-wise API calls with batch size 1 to simulate interactive decision-making. Embedding computations were conducted using two models: for one-time task-level metadata embedding, we used OpenAI’s \texttt{text-embedding-3-small}; for dynamic, step-wise retrieval, we used the \texttt{all-MiniLM-L6-v2} model~\cite{reimers-gurevych-2019-sentence}. All experiments were executed under Ubuntu 24.04 with Python 3.9.21, and intermediate results were locally cached to reduce redundant compute. API usage totaled approximately \$100 USD, with cost breakdowns based on provider pricing: for GPT-4o-mini, \$1.10/\textit{M} input tokens, \$0.275/\textit{M} cached input tokens, and \$4.40/\textit{M} output tokens; for Qwen-Turbo, minimum rates were \$0.05/\textit{M} input tokens and \$0.20/\textit{M} output tokens, with a maximum context length of 1,008,192 tokens.

\section{Additional Experimental Results\label{appendix:additional_results}}

\begin{table*}[!t]
\centering
\caption{Effect of Skills Across Prompting Strategies. Grounding Rate $[\%]$ ($\uparrow$), Progress Rate $[\%]$ ($\uparrow$), Success Rate $[\%]$ ($\uparrow$), and AUPC $[0, 1]$ ($\uparrow$). The best method for each  LLM is in \textbf{bold}.~\label{tab:appendix_skill2}}
\resizebox{1\textwidth}{!}{\begin{tabular}{c|cccc|cccc|cccc} 
\toprule
\multirow{2}{*}{\textbf{Method} } & \multicolumn{4}{c|}{\textbf{Qwen2.5-7B-Instruct} }                          & \multicolumn{4}{c|}{\textbf{Qwen-Turbo} }                          & \multicolumn{4}{c}{\textbf{GPT-4o-mini} }                           \\
                                  & \textbf{GR}    & \textbf{PR}    & \textbf{SR}    & \textbf{AUPC}   & \textbf{GR}    & \textbf{PR}    & \textbf{SR}    & \textbf{AUPC}   & \textbf{GR}    & \textbf{PR}    & \textbf{SR}    & \textbf{AUPC}    \\ 
\hline\hline
\multicolumn{13}{c}{\textbf{ALFWorld} }                                                                                                                                                                                                           \\ 
\midrule
0-shot                            & 10.5           & 6.0            & 0.8            & 0.027           & 56.3           & 32.2           & 9.7            & 0.212           & 73.7           & 26.8           & 1.5            & 0.184            \\
0-shot + skill                    & 66.0           & 53.0           & 28.4           & 0.350           & 62.1           & 47.5           & 20.9           & 0.301           & 62.2           & 34.3           & 9.7            & 0.170            \\
1-shot                            & 28.1           & 16.0           & 2.2            & 0.095           & 63.9           & 55.3           & 36.5           & 0.380           & 77.3           & 43.3           & 10.5           & 0.292            \\
1-shot + skill                    & 78.0           & \textbf{67.7}  & 49.3           & \textbf{0.459}  & 68             & 55.2           & 36.6           & 0.333           & 77.9           & \textbf{53.7}  & \textbf{23.8}  & 0.343            \\
Synapse                           & 61.5           & 41.6           & 17.1           & 0.278           & 74.9           & 54.7           & 35.8           & 0.379           & 76.3           & 48.8           & 14.8           & 0.340            \\
Synapse + skill                   & \textbf{84.0}  & 67.3           & \textbf{53.0}  & 0.457           & \textbf{83.4}  & \textbf{64.7}  & \textbf{50.1}  & \textbf{0.436}  & \textbf{81.8}  & 53.0           & 23.8           & \textbf{0.363}   \\ 
\hline\hline
\multicolumn{13}{c}{\textbf{BabyAI} }                                                                                                                                                                                                             \\ 
\midrule
0-shot                            & 31.8           & 21.8           & 7.1            & 0.037           & 50.2           & 32.7           & 19.6           & 0.092           & 55.3           & 34.2           & 22.3           & 0.129            \\
0-shot + skill                    & 41.9           & 32.2           & 15.2           & 0.070           & 50.6           & 40.8           & 26.8           & 0.134           & 83.0           & 51.3           & 35.7           & 0.200            \\
1-shot                            & 59.2           & 36.5           & 18.8           & 0.112           & 61.4           & 37.2           & 16.3           & 0.076           & 76.6           & 42.6           & 28.6           & 0.154            \\
1-shot + skill                    & 64.7           & \textbf{49.4}  & 28.6           & 0.152           & 52.4           & 44.4           & 31.2           & 0.132           & 90.9           & \textbf{52.5}  & 40.2           & 0.198            \\
Synapse                           & \textbf{67.2}  & 39.4           & 21.4           & 0.153           & 65.0           & \textbf{55.8}  & \textbf{45.5}  & \textbf{0.242}  & 86.5           & 44.9           & 33.9           & 0.169            \\
Synapse + skill                   & 58.8           & 47.0           & \textbf{31.2}  & \textbf{0.153}  & \textbf{67.4}  & 52.2           & 42.8           & 0.230           & \textbf{91.5}  & 52.2           & \textbf{40.2}  & \textbf{0.211}   \\ 
\hline\hline
\multicolumn{13}{c}{\textbf{ScienceWorld} }                                                                                                                                                                                                       \\ 
\midrule
0-shot                            & 10.8           & 27.1           & 9.0            & 0.136           & \textbf{28.8}  & 19.3           & 4.4            & 0.113           & 34.3           & 44.3           & 7.7            & 0.206            \\
0-shot + skill                    & \textbf{13.6}  & \textbf{38.0}  & \textbf{16.7}  & \textbf{0.237}  & 12.3           & \textbf{37.1}  & 10.0           & \textbf{0.232}  & 25.9           & 54.3           & 18.0           & 0.313            \\
1-shot                            & 9.6            & 19.1           & 5.5            & 0.108           & 11.8           & 19.1           & 7.7            & 0.111           & \textbf{34.3}  & 46.8           & 18.8           & 0.206            \\
1-shot + skill                    & 10.5           & 33.8           & 15.6           & 0.212           & 9.3            & 36.9           & 7.8            & 0.223           & 20.1           & 60.2           & 31.2           & 0.386            \\
Synapse                           & 8.6            & 15.4           & 4.4            & 0.090           & 5.4            & 16.0           & 3.3            & 0.106           & 14.0           & 52.8           & 25.6           & 0.334            \\
Synapse + skill                   & 10.7           & 26.8           & 13.3           & 0.156           & 9.1            & 34.0           & \textbf{10.1}  & 0.224           & 25.1           & \textbf{64.4}  & \textbf{36.8}  & \textbf{0.417}  \\
\bottomrule
\end{tabular}
}
\end{table*}

\begin{table*}[!t]
\centering
\caption{Effect of Step-wise Skills (Supplementary Results for Figure 4). Grounding Rate $[\%]$ ($\uparrow$), Progress Rate $[\%]$ ($\uparrow$), Success Rate $[\%]$ ($\uparrow$), and AUPC $[0, 1]$ ($\uparrow$). The best method for each  LLM is in \textbf{bold}.~\label{tab:appendix_skill1}}
\resizebox{1\textwidth}{!}{
\begin{tabular}{c|cccc|cccc|cccc} 
\toprule
\multirow{2}{*}{\textbf{Method}} & \multicolumn{4}{c|}{\textbf{Qwen2.5-7B-Instruct}}                & \multicolumn{4}{c|}{\textbf{Qwen-Turbo}}                & \multicolumn{4}{c}{\textbf{GPT-4o-mini}}                 \\
                                 & \textbf{GR} & \textbf{PR} & \textbf{SR} & \textbf{AUPC} & \textbf{GR} & \textbf{PR} & \textbf{SR} & \textbf{AUPC} & \textbf{GR} & \textbf{PR} & \textbf{SR} & \textbf{AUPC}  \\ 
\hline\hline
\multicolumn{13}{c}{\textbf{ALFWorld}}                                                                                                                                                                          \\ 
\midrule
\name w/o skills              & 83.0        & 67.0        & 50.7        & 0.471         & 86.2        & 66.8        & 53.3        & 0.462         & 82.0        & 51.6        & 21.7        & 0.354          \\
\name w/o step-wise skills    & 81.1        & 63.5        & 47.1        & 0.432         & 83.6        & 66.6        & 53.1        & 0.450         & 83.9        & 54.7        & 26.8        & 0.380          \\
\name                         & 84.9        & 68.0        & 55.2        & 0.464         & 85.6        & 67.6        & 53.8        & 0.460         & 83.6        & 55.1        & 29.8        & 0.369          \\ 
\hline\hline
\multicolumn{13}{c}{\textbf{BabyAI}}                                                                                                                                                                            \\ 
\midrule
\name w/o skills              & 68.0        & 47.6        & 27.7        & 0.136         & 74.2        & 54.8        & 41.9        & 0.218         & 87.9        & 53.2        & 36.6        & 0.207          \\
\name w/o step-wise skills    & 77.9        & 50.0        & 28.6        & 0.154         & 75.2        & 55.0        & 42.9        & 0.224         & 91.1        & 56.2        & 41.1        & 0.234          \\
\name                         & 66.7        & 50.0        & 31.2        & 0.158         & 73.9        & 59.4        & 45.5        & 0.254         & 89.5        & 57.6        & 41.1        & 0.248          \\ 
\hline\hline
\multicolumn{13}{c}{\textbf{ScienceWorld}}                                                                                                                                                                      \\ 
\midrule
\name w/o skills              & 12.8        & 40.4        & 18.9        & 0.250         & 14.2        & 36          & 10.1        & 0.206         & 23.3        & 63.0        & 35.8        & 0.412          \\
\name w/o step-wise skills    & 13.8        & 43.3        & 17.7        & 0.278         & 10.3        & 36.8        & 11.1        & 0.275         & 26.4        & 66.9        & 39.1        & 0.424          \\
\name                         & 16.1        & 46.7        & 23.4        & 0.298         & 13.3        & 37.7        & 11.1        & 0.242         & 25.3        & 67.3        & 40.2        & 0.442          \\
\bottomrule
\end{tabular}}
\end{table*}

\begin{table*}[!t]
\centering
\caption{Effect of TD-based Credit Assignment (Supplementary Results for Figure 5. Grounding Rate $[\%]$ ($\uparrow$), Progress Rate $[\%]$ ($\uparrow$), Success Rate $[\%]$ ($\uparrow$), and AUPC $[0, 1]$ ($\uparrow$). The best method for each  LLM is in \textbf{bold}.~\label{tab:appendix_td}}
\resizebox{1\textwidth}{!}{
\begin{tabular}{c|cccc|cccc|cccc}
\toprule
\multirow{2}{*}{\textbf{Method}} & \multicolumn{4}{c|}{\textbf{Qwen2.5-7B-Instruct}}                       & \multicolumn{4}{c|}{\textbf{Qwen-Turbo}}                       & \multicolumn{4}{c}{\textbf{GPT-4o-mini}}                        \\
                                 & \textbf{GR}   & \textbf{PR}   & \textbf{SR}   & \textbf{AUPC}  & \textbf{GR}   & \textbf{PR}   & \textbf{SR}   & \textbf{AUPC}  & \textbf{GR}   & \textbf{PR}   & \textbf{SR}   & \textbf{AUPC}   \\ 
\hline\hline
\multicolumn{13}{c}{\textbf{ALFWorld}}                                                                                                                                                                                               \\ 
\midrule
Trajectory Retrieval             & 65.4          & 44.2          & 22.4          & 0.296          & 65.5          & 54.8          & 35.8          & 0.372          & 79.1          & 49.1          & 16.4          & 0.341           \\
Segment Retrieval                & 72.0          & 58.9          & 42.4          & 0.408          & 83.6          & 63.8          & 51.5          & 0.433          & 81.8          & 53.3          & 23.9          & 0.359           \\
\name                         & \textbf{84.9} & \textbf{68.0} & \textbf{55.2} & \textbf{0.464} & \textbf{85.6} & \textbf{67.6} & \textbf{53.8} & \textbf{0.460} & \textbf{83.6} & \textbf{55.1} & \textbf{29.8} & \textbf{0.369}  \\ 
\hline\hline
\multicolumn{13}{c}{\textbf{BabyAI}}                                                                                                                                                                                                 \\ 
\midrule
Trajectory Retrieval             & 68.2          & 36.9          & 19.7          & 0.115          & 64.9          & 46.9          & 34.8          & 0.157          & 87.3          & 40.9          & 30.4          & 0.126           \\
Segment Retrieval                & 70.6          & 44.9          & 25.9          & 0.125          & 71.4          & 55.4          & 41.1          & 0.225          & 89.4          & 52.2          & 37.5          & 0.208           \\
\name                         & \textbf{66.7} & \textbf{50.0} & \textbf{31.2} & \textbf{0.158} & \textbf{73.9} & \textbf{59.4} & \textbf{45.5} & \textbf{0.254} & \textbf{89.5} & \textbf{57.6} & \textbf{41.1} & \textbf{0.248}  \\ 
\hline\hline
\multicolumn{13}{c}{\textbf{ScienceWorld}}                                                                                                                                                                                           \\ 
\midrule
Trajectory Retrieval             & 7.3           & 21.1          & 4.4           & 0.135          & 7.1           & 29.3          & 8.8           & 0.180          & 16.9          & 61.4          & 29.0          & 0.375           \\
Segment Retrieval                & 13.6          & 42.7          & 19            & 0.266          & 11            & 28            & 7.8           & 0.167          & 20.9          & 55.0          & 23.4          & 0.332           \\
\name                         & \textbf{16.1} & \textbf{46.7} & \textbf{23.4} & \textbf{0.298} & \textbf{13.3} & \textbf{37.7} & \textbf{11.1} & \textbf{0.242} & \textbf{25.3} & \textbf{67.3} & \textbf{40.2} & \textbf{0.442}  \\
\bottomrule
\end{tabular}
}
\end{table*}

\begin{table*}[!t]
\centering
\caption{Effect of Focused Prompting with Golden Segments. Grounding Rate $[\%]$ ($\uparrow$), Progress Rate $[\%]$ ($\uparrow$), Success Rate $[\%]$ ($\uparrow$), and AUPC $[0, 1]$ ($\uparrow$). The best method for each  LLM is in \textbf{bold}.~\label{tab:appendix_segment2}}
\resizebox{1\textwidth}{!}{\begin{tabular}{c|cccc|cccc|cccc} 
\toprule
\multirow{2}{*}{\textbf{Method}} & \multicolumn{4}{c|}{\textbf{Qwen2.5-7B-Instruct}}                       & \multicolumn{4}{c|}{\textbf{Qwen-Turbo}}                       & \multicolumn{4}{c}{\textbf{GPT-4o-mini}}                        \\
                                 & \textbf{GR}   & \textbf{PR}   & \textbf{SR}   & \textbf{AUPC}  & \textbf{GR}   & \textbf{PR}   & \textbf{SR}   & \textbf{AUPC}  & \textbf{GR}   & \textbf{PR}   & \textbf{SR}   & \textbf{AUPC}   \\ 
\hline\hline
\multicolumn{13}{c}{\textbf{ALFWorld}}                                                                                                                                                                                               \\ 
\midrule
Static Trajectory                & 28.1          & 16.0          & 2.2           & 0.095          & 63.9          & 55.3          & 36.5          & 0.380          & 77.3          & 43.3          & 10.5          & 0.292           \\
Retrieved Trajectory             & 61.5          & 41.6          & 17.1          & 0.278          & 74.9          & 54.7          & 35.8          & 0.379          & 76.3          & 48.8          & 14.8          & 0.340           \\
Synapse (3-shot)                 & 71.4          & 44.8          & 19.4          & 0.302          & 78.4          & 60.6          & 47.0          & 0.421          & 77.4          & \textbf{52.9} & 17.8          & \textbf{0.360}  \\
Golden Segment                   & \textbf{83.0} & \textbf{67.0} & \textbf{50.7} & \textbf{0.471} & \textbf{86.2} & \textbf{66.8} & \textbf{53.3} & \textbf{0.462} & \textbf{82.0} & 51.6          & \textbf{21.7} & 0.354           \\ 
\hline\hline
\multicolumn{13}{c}{\textbf{BabyAI}}                                                                                                                                                                                                 \\ 
\midrule
Static Trajectory                & 59.2          & 36.5          & 18.8          & 0.112          & 61.4          & 37.2          & 16.3          & 0.076          & 76.6          & 42.6          & 28.6          & 0.154           \\
Retrieved Trajectory             & 67.2          & 39.4          & 21.4          & 0.153          & 65.0          & \textbf{55.8} & 45.5          & \textbf{0.242} & 86.5          & 44.9          & 33.9          & 0.169           \\
Synapse (3-shot)                 & \textbf{78.6} & 44.6          & \textbf{28.6} & \textbf{0.163} & 62.1          & 50.8          & 38.4          & 0.191          & \textbf{92.8} & 49.5          & \textbf{38.4} & 0.188           \\
Golden Segment                   & 68.0          & \textbf{47.6} & 27.7          & 0.136          & \textbf{74.2} & 54.8          & \textbf{b}    & 0.218          & 87.9          & \textbf{53.2} & 36.6          & \textbf{0.207}  \\ 
\hline\hline
\multicolumn{13}{c}{\textbf{ScienceWorld}}                                                                                                                                                                                           \\ 
\midrule
Static Trajectory                & 9.6           & 19.1          & 5.5           & 0.108          & 11.8          & 19.1          & 7.7           & 0.111          & 34.3          & 46.8          & 18.8          & 0.206           \\
Retrieved Trajectory             & 8.6           & 15.4          & 4.4           & 0.090          & 5.4           & 16.0          & 3.3           & 0.106          & 14.0          & 52.8          & 25.6          & 0.334           \\
Synapse (3-shot)                 & 6.0           & 15.3          & 5.5           & 0.086          & 7.2           & 24.0          & 10.1          & 0.155          & 15.5          & 60.0          & 32.4          & 0.390           \\
Golden Segment                   & \textbf{16.1} & \textbf{46.7} & \textbf{23.4} & \textbf{0.298} & \textbf{13.3} & \textbf{37.7} & \textbf{11.1} & \textbf{0.242} & \textbf{25.3} & \textbf{67.3} & \textbf{40.2} & \textbf{0.442}  \\
\bottomrule
\end{tabular}
}
\end{table*}

\subsection{Effect of Step-wise Skills.}

\paragraph{Effect of Skills Across Prompting Strategies.}
Table~\ref{tab:appendix_skill2} presents a detailed comparison of skill prompting across different prompting settings—0-shot, 1-shot, and Synapse—on three domains. Across all models and tasks, the addition of skills consistently improves performance, particularly in progress rate (PR), success rate (SR), and AUPC. For example, in ALFWorld, augmenting both 1-shot and Synapse baselines with skills yields substantial gains in SR and AUPC, with the skill-enhanced Synapse achieving the highest scores across all metrics for Qwen2.5-7B-Instruct and Qwen-Turbo. In BabyAI, while Synapse without skills already performs well, skill augmentation further boosts SR and AUPC, highlighting the complementary effect of structured prompting. The improvements are especially striking in ScienceWorld, where both 0-shot and 1-shot settings benefit significantly from skill injection, leading to large increases in SR (e.g., from 18.8 to 31.2 on GPT-4o-mini) and smoother task execution as measured by AUPC. While using skills alone (e.g., 0-shot + skill) already provides notable gains over the base 0-shot prompt, the combination with demonstration-based or few-shot prompts yields the most consistent improvements. These results confirm that skill-based prompting enhances grounding and accelerates progress, offering a flexible plug-in benefit across prompting strategies and model capacities.

\paragraph{Supplementary Results for Figure 7.}
Table~\ref{tab:appendix_skill1} provides supplementary results analyzing the effect of skills on grounding rate (GR), progress rate (PR), success rate (SR), and AUPC across three domains and LLMs. While PR and SR trends are already visualized in the main text, this table complements those findings by reporting GR and AUPC, offering a more comprehensive view of task grounding and execution efficiency. We observe that incorporating both domain-level and step-wise skills generally improves AUPC and GR, reflecting more consistent and timely progress. Notably, the full \name method achieves the highest performance across most metrics and models, especially in ALFWorld and ScienceWorld. In BabyAI, ablations show more mixed effects, highlighting the sensitivity of fine-grained skill application. These results reinforce the contribution of structured, step-aware skill prompting in enhancing task alignment and decision efficiency.

\subsection{Effect of TD-based Credit Assignment.}
\paragraph{Supplementary Results for Figure 5.}
Table~\ref{tab:appendix_td} provides the full numerical results corresponding to Figure 5, showing the effect of TD-based credit assignment across ALFWorld, BabyAI, and ScienceWorld. Consistent with the trends illustrated in the figure, \textit{\name} outperforms both \textit{Trajectory Retrieval} and \textit{Segment Retrieval} across all LLMs and tasks. In ALFWorld, \textit{\name} improves Progress Rate and Success Rate significantly—for example, achieving 68.0 PR and 55.2 SR on Qwen2.5-7B-Instruct, compared to 58.9 PR and 42.4 SR with Segment Retrieval. On the more compositional ScienceWorld, \name reaches 40.2 SR on GPT-4o-mini—a 16.8\% absolute gain over Segment Retrieval. These results reinforce the claim that TD-based credit assignment is critical for identifying high-utility actions and enabling effective skill-based prompting.

\begin{table*}[!t]
\centering
\caption{Effect of Golden Segment (Supplementary Results for Figure 3.). Grounding Rate $[\%]$ ($\uparrow$), Progress Rate $[\%]$ ($\uparrow$), Success Rate $[\%]$ ($\uparrow$), and AUPC $[0, 1]$ ($\uparrow$). The best method for each  LLM is in \textbf{bold}.~\label{tab:appendix_segment1}}
\resizebox{1\textwidth}{!}{\begin{tabular}{c|cccc|cccc|cccc} 
\toprule
\multirow{2}{*}{\textbf{Method}} & \multicolumn{4}{c|}{\textbf{Qwen2.5-7B-Instruct}}                       & \multicolumn{4}{c|}{\textbf{Qwen-Turbo}}                       & \multicolumn{4}{c}{\textbf{GPT-4o-mini}}                        \\
                                 & \textbf{GR}   & \textbf{PR}   & \textbf{SR}   & \textbf{AUPC}  & \textbf{GR}   & \textbf{PR}   & \textbf{SR}   & \textbf{AUPC}  & \textbf{GR}   & \textbf{PR}   & \textbf{SR}   & \textbf{AUPC}   \\ 
\hline\hline
\multicolumn{13}{c}{\textbf{ALFWorld}}                                                                                                                                                                                               \\ 
\midrule
Fixed Demo+Skills                & 78            & 67.7          & 49.3          & 0.459          & 68            & 55.2          & 36.6          & 0.333          & 77.9          & 53.7          & 23.8          & 0.343           \\
Retrieved   Demo+Skills          & 84.0          & 67.3          & 53.0          & 0.457          & 83.4          & 64.7          & 50.1          & 0.436          & 81.8          & 53.0          & 23.8          & 0.363           \\
\name                         & \textbf{84.9} & \textbf{68.0} & \textbf{55.2} & \textbf{0.464} & \textbf{85.6} & \textbf{67.6} & \textbf{53.8} & \textbf{0.460} & \textbf{83.6} & \textbf{55.1} & \textbf{29.8} & \textbf{0.369}  \\ 
\hline\hline
\multicolumn{13}{c}{\textbf{BabyAI}}                                                                                                                                                                                                 \\ 
\midrule
Fixed Demo+Skills                & 64.7          & 49.4          & 28.6          & 0.152          & 52.4          & 44.4          & 31.2          & 0.132          & 90.9          & 52.5          & 40.2          & 0.198           \\
Retrieved   Demo+Skills          & 58.8          & 47.0          & 31.2          & 0.153          & 67.4          & 52.2          & 42.8          & 0.230          & \textbf{91.5} & 52.2          & 40.2          & 0.211           \\
\name                         & \textbf{66.7} & \textbf{50.0} & \textbf{31.2} & \textbf{0.158} & \textbf{73.9} & \textbf{59.4} & \textbf{45.5} & \textbf{0.254} & 89.5          & \textbf{57.6} & \textbf{41.1} & \textbf{0.248}  \\ 
\hline\hline
\multicolumn{13}{c}{\textbf{ScienceWorld}}                                                                                                                                                                                           \\ 
\midrule
Fixed Demo+Skills                & 10.5          & 33.8          & 15.6          & 0.212          & 9.3           & 36.9          & 7.8           & 0.223          & 20.1          & 60.2          & 31.2          & 0.386           \\
Retrieved   Demo+Skills          & 10.7          & 36.8          & 13.3          & 0.156          & 9.1           & 34.0          & 10.1          & 0.224          & 25.1          & 64.4          & 36.8          & 0.417           \\
\name                         & \textbf{16.1} & \textbf{46.7} & \textbf{23.4} & \textbf{0.298} & \textbf{13.3} & \textbf{37.7} & \textbf{11.1} & \textbf{0.242} & \textbf{25.3} & \textbf{67.3} & \textbf{40.2} & \textbf{0.442}  \\
\bottomrule
\end{tabular}
}
\vspace{-1em}
\end{table*}

\subsection{Effect of Golden Segment.}
\vspace{-0.5em}
\paragraph{Effect of Focused Prompting with Golden Segments.}
Table~\ref{tab:appendix_segment2} provides a deeper analysis of the golden segment’s effectiveness by comparing it with various prompt strategies, including static and retrieved trajectories, as well as the 3-shot Synapse baseline. Across all three benchmarks, the \textit{Golden Segment} consistently delivers the highest performance in most settings—particularly in Success Rate and AUPC—demonstrating its superiority despite being shorter than full demonstrations. For instance, in ScienceWorld with GPT-4o-mini, the golden segment alone achieves a 40.2\% Success Rate and 0.442 AUPC, far surpassing all other methods. These results highlight that focused prompting—not longer or more complex input—is the key to effective decision-making. By isolating high-utility, decision-critical information, our approach substantiates the core principle that focused context matters more than sheer quantity in prompting strategies.

\vspace{-0.5em}
\paragraph{Supplementary Results for Figure 3.}
Table~\ref{tab:appendix_segment1} reports the full results comparing \textit{Fixed Demo+Skills}, \textit{Retrieved Demo+Skills}, and \textit{\name} across three benchmarks. \textit{\name} consistently outperforms the other methods in all settings, validating the benefit of using focused, high-impact segments over full demonstrations. The most pronounced improvements are observed in ALFWorld and ScienceWorld, highlighting that precise, task-relevant context enables more effective reasoning and generalization.

\begin{table*}[!t]
\centering
\caption{Effect of Subgoal Annotations. Grounding Rate $[\%]$ ($\uparrow$), Progress Rate $[\%]$ ($\uparrow$), Success Rate $[\%]$ ($\uparrow$), and AUPC $[0, 1]$ ($\uparrow$). The best method for each  LLM is in \textbf{bold}.~\label{tab:appendix_subgoal}}
\resizebox{1\textwidth}{!}{
\begin{tabular}{c|cccc|cccc|cccc} 
\toprule
\multirow{2}{*}{\textbf{Method} }  & \multicolumn{4}{c|}{\textbf{Qwen2.5-7B-Instruct} }                          & \multicolumn{4}{c|}{\textbf{Qwen-Turbo} }                          & \multicolumn{4}{c}{\textbf{GPT-4o-mini} }                           \\
                                   & \textbf{GR}    & \textbf{PR}    & \textbf{SR}    & \textbf{AUPC}   & \textbf{GR}    & \textbf{PR}    & \textbf{SR}    & \textbf{AUPC}   & \textbf{GR}    & \textbf{PR}    & \textbf{SR}    & \textbf{AUPC}    \\ 
\hline\hline
\multicolumn{13}{c}{\textbf{ALFWorld} }                                                                                                                                                                                                            \\ 
\midrule
Synapse (3-shot)                   & 71.4           & 44.8           & 19.4           & 0.302           & 78.4           & 60.6           & 47.0           & 0.421           & 77.4           & 52.9           & 17.8           & 0.360            \\
\name w/o subgoal   (sampling=6) & 82.0           & 63.2           & 50.1           & 0.440           & \textbf{87.0}  & 64.6           & 52.2           & 0.439           & 47.8           & 32.9           & 14.2           & 0.224            \\
\name(sampling=6)               & \textbf{84.9}  & \textbf{68.0}  & \textbf{55.2}  & \textbf{0.464}  & 85.6           & \textbf{67.6}  & \textbf{53.8}  & \textbf{0.460}  & \textbf{83.6}  & \textbf{55.1}  & \textbf{29.8}  & \textbf{0.369}   \\ 
\hline\hline
\multicolumn{13}{c}{\textbf{BabyAI} }                                                                                                                                                                                                              \\ 
\midrule
Synapse (3-shot)                   & \textbf{78.6}  & 44.6           & 28.6           & \textbf{0.163}  & 62.1           & 50.8           & 38.4           & 0.191           & \textbf{92.8}  & 49.5           & 38.4           & 0.188            \\
\name w/o subgoal   (sampling=6) & 59.6           & 43.2           & \textbf{32.1}  & 0.160           & 66.1           & 50.5           & 36.6           & 0.221           & 88.5           & 58.8           & \textbf{43.8}  & 0.232            \\
\name(sampling=6)               & 66.7           & \textbf{50.0}  & 31.2           & 0.158           & \textbf{73.9}  & \textbf{59.4}  & \textbf{45.5}  & \textbf{0.254}  & 89.5           & \textbf{57.6}  & 41.1           & \textbf{0.248}   \\ 
\hline\hline
\multicolumn{13}{c}{\textbf{ScienceWorld} }                                                                                                                                                                                                        \\ 
\midrule
Synapse (3-shot)                   & 6.0            & 15.3           & 5.5            & 0.086           & 7.2            & 24.0           & 10.1           & 0.155           & 15.5           & 60.0           & 32.4           & 0.390            \\
\name w/o subgoal   (sampling=6) & 11             & 29.1           & 18.9           & 0.175           & 9              & 27.7           & 10             & 0.147           & \textbf{27.8}  & 56.2           & 32.4           & 0.357            \\
\name(sampling=6)               & \textbf{16.1}  & \textbf{46.7}  & \textbf{23.4}  & \textbf{0.298}  & \textbf{13.3}  & \textbf{37.7}  & \textbf{11.1}  & \textbf{0.242}  & 25.3           & \textbf{67.3}  & \textbf{40.2}  & \textbf{0.442}   \\
\bottomrule
\end{tabular}
}
\vspace{-1em}
\end{table*}

\subsection{Effect of Subgoal Annotations}
\vspace{-0.5em}
\paragraph{Subgoal annotation significantly enhances \name’s generalization in challenging settings.} As shown in Table~\ref{tab:appendix_subgoal}, when equipped with subgoal supervision, \name consistently outperforms its subgoal-free counterpart across tasks and model backbones, with the largest gains observed in complex domains like ScienceWorld. For instance, on ScienceWorld with Qwen2.5-7B-Instruct, the success rate increases from 18.9\% to 23.4\%, and the progress rate improves from 29.1\% to 46.7\%. On GPT-4o-mini, AUPC rises substantially from 0.357 to 0.442—a 23.8\% relative improvement. Similarly, on ALFWorld with GPT-4o-mini, subgoal-aware \name nearly doubles the success rate from 14.2\% to 29.8\%. These results underscore the value of intermediate structure in improving reasoning efficiency, particularly in tasks that involve long-horizon dependencies or ambiguous subgoal transitions.

\vspace{-0.5em}
\paragraph{\name achieves strong generalization even without subgoal supervision.} Despite using only sampled successful trajectories without subgoal labels, \name outperforms strong multi-shot prompting baselines like Synapse (3-shot) across all benchmarks. On ALFWorld with Qwen2.5-7B-Instruct, it boosts the success rate from 19.4\% to 50.1\%, a 2.6× improvement. On ScienceWorld, it nearly doubles the progress rate (15.3\% → 29.1\%) and triples the success rate (5.5\% → 18.9\%). Even in BabyAI with Qwen-Turbo, where improvements are more modest, AUPC increases from 0.191 to 0.221 (+16\%). These gains highlight the robustness of \name’s interaction-derived skill extraction, which enables effective generalization without relying on explicit subgoal supervision.

\subsection{Sensitive Analysis on Sampling Scaling}

\begin{table*}[!t]
\centering
\caption{Sensitive Analysis on Sampling Scaling (Supplementary Results for Figure 6). Grounding Rate $[\%]$ ($\uparrow$), Progress Rate $[\%]$ ($\uparrow$), Success Rate $[\%]$ ($\uparrow$), and AUPC $[0, 1]$ ($\uparrow$) on ScienceWorld. The best method for each  LLM is in \textbf{bold}.~\label{tab:appendix_sensitive_sampling}}
\resizebox{1\textwidth}{!}{\begin{tabular}{c|cccc|cccc|cccc} 
\toprule
\multirow{2}{*}{\textbf{Method}} & \multicolumn{4}{c|}{\textbf{Qwen2.5-7B-Instruct}}                       & \multicolumn{4}{c|}{\textbf{Qwen-Turbo}}                       & \multicolumn{4}{c}{\textbf{GPT-4o-mini}}                        \\
                                 & \textbf{GR}   & \textbf{PR}   & \textbf{SR}   & \textbf{AUPC}  & \textbf{GR}   & \textbf{PR}   & \textbf{SR}   & \textbf{AUPC}  & \textbf{GR}   & \textbf{PR}   & \textbf{SR}   & \textbf{AUPC}   \\ 
\hline
\multicolumn{13}{c}{\textbf{ALFWorld}}                                                                                                                                                                                               \\ 
\midrule
\name(sampling=3)             & 84.4          & 68.8          & 53.7          & 0.475          & 84.0          & 64.5          & 50.1          & 0.423          & 82.7          & 51.6          & 25.4          & 0.353           \\
\name(sampling=6)             & \textbf{84.9} & \textbf{68.0} & \textbf{55.2} & \textbf{0.464} & 85.6          & \textbf{67.6} & 53.8          & \textbf{0.460} & \textbf{83.6} & \textbf{55.1} & \textbf{29.8} & \textbf{0.369}  \\
\name(sampling=9)             & 83.6          & 68.3          & 53.7          & 0.463          & \textbf{86.5} & 66.4          & \textbf{54.4} & 0.442          & 80.1          & 49.5          & 29.4          & 0.334           \\
\name(sampling=12)            & 84.4          & 66.4          & 51.4          & 0.455          & 85.2          & 65.2          & 50.7          & 0.442          & 79.6          & 49.1          & 23.1          & 0.322           \\ 
\hline\hline
\multicolumn{13}{c}{\textbf{BabyAI}}                                                                                                                                                                                                 \\ 
\midrule
\name(sampling=3)             & 66.5          & 45.2          & 26.8          & 0.128          & 69.8          & 50.9          & 35.7          & 0.212          & 88.1          & 55.2          & 41.0          & 0.219           \\
\name(sampling=6)             & 66.7          & \textbf{50.0} & 31.2          & 0.158          & \textbf{73.9} & \textbf{59.4} & \textbf{45.5} & \textbf{0.254} & 89.5          & \textbf{57.6} & \textbf{41.1} & \textbf{0.248}  \\
\name(sampling=9)             & 67.8          & 46.2          & 30.4          & 0.162          & 72.8          & 52.8          & 39.3          & 0.218          & \textbf{89.6} & 56.9          & 41.1          & 0.228           \\
\name(sampling=12)            & \textbf{68.4} & 49.7          & \textbf{33.0} & \textbf{0.186} & 72.9          & 54.1          & 42.0          & 0.234          & 89.4          & 54.6          & 39.3          & 0.216           \\ 
\hline\hline
\multicolumn{13}{c}{\textbf{ScienceWorld}}                                                                                                                                                                                           \\ 
\midrule
\name(sampling=3)             & 14.7          & 40.1          & 15.5          & 0.243          & \textbf{14.6} & \textbf{42.9} & 12.4          & 0.262          & 24.6          & 62.3          & 33.4          & 0.396           \\
\name(sampling=6)             & 16.1          & \textbf{46.7} & \textbf{23.4} & \textbf{0.298} & 13.3          & 37.7          & 11.1          & \textbf{0.242} & \textbf{25.3} & \textbf{67.3} & \textbf{40.2} & \textbf{0.442}  \\
\name(sampling=9)             & 15.5          & 46.4          & 23.4          & 0.294          & 13            & 36.8          & 12.2          & 0.235          & 24.7          & 64.4          & 33.6          & 0.424           \\
\name(sampling=12)            & \textbf{16.4} & 45.4          & 23.3          & 0.285          & 12.4          & 38.5          & \textbf{13.4} & 0.247          & 24.5          & 65.3          & 39            & 0.435           \\
\bottomrule
\end{tabular}}
\vspace{-1em}
\end{table*}

\begin{table*}[!t]
\centering
\caption{Sensitivity Analysis on Sampling Temperature. Grounding Rate $[\%]$ ($\uparrow$), Progress Rate $[\%]$ ($\uparrow$), Success Rate $[\%]$ ($\uparrow$), and AUPC $[0, 1]$ ($\uparrow$). The best method for each setting is in \textbf{bold}.~\label{tab:appendix_sensitive_temperature}}
\resizebox{\textwidth}{!}{
\begin{tabular}{c|cccc|cccc|cccc}
\toprule
\multirow{2}{*}{\textbf{Temperature}} & \multicolumn{4}{c|}{\textbf{ALFWorld}} & \multicolumn{4}{c|}{\textbf{BabyAI}} & \multicolumn{4}{c}{\textbf{ScienceWorld}} \\
 & \textbf{GR} & \textbf{PR} & \textbf{SR} & \textbf{AUPC} 
 & \textbf{GR} & \textbf{PR} & \textbf{SR} & \textbf{AUPC} 
 & \textbf{GR} & \textbf{PR} & \textbf{SR} & \textbf{AUPC} \\
\hline\hline
0.3 & 80.7 & 63.8 & 47.8 & 0.447 
    & 65.8 & 46.1 & 26.8 & 0.134 
    & \textbf{17.5} & 41.1 & 14.6 & 0.232 \\
0.7 & \textbf{85.5} & \textbf{70.1} & \textbf{56.7} & \textbf{0.485} 
    & 64.3 & 44.4 & 28.6 & 0.136 
    & 14.9 & 39.1 & 14.3 & 0.230 \\
1.0 & 84.9 & 68.0 & 55.2 & 0.464 
    & \textbf{66.7} & \textbf{50.0} & \textbf{31.2} & \textbf{0.158} 
    & 16.1 & \textbf{46.7} & \textbf{23.4} & \textbf{0.298} \\
\bottomrule
\end{tabular}
}
\end{table*}

\begin{table*}[!t]
\centering
\caption{Sensitive Analysis on Skill Retrieval (Supplementary Results for Figure 6). Grounding Rate $[\%]$ ($\uparrow$), Progress Rate $[\%]$ ($\uparrow$), Success Rate $[\%]$ ($\uparrow$), and AUPC $[0, 1]$ ($\uparrow$). The best method for each  LLM is in \textbf{bold}.~\label{tab:appendix_sensitive_retrieval}}
\resizebox{1\textwidth}{!}{\begin{tabular}{c|cccc|cccc|cccc} 
\toprule
\multirow{2}{*}{\textbf{Method}} & \multicolumn{4}{c|}{\textbf{Qwen2.5-7B-Instruct}}                       & \multicolumn{4}{c|}{\textbf{Qwen-Turbo}}                       & \multicolumn{4}{c}{\textbf{GPT-4o-mini}}                        \\
                                 & \textbf{GR}   & \textbf{PR}   & \textbf{SR}   & \textbf{AUPC}  & \textbf{GR}   & \textbf{PR}   & \textbf{SR}   & \textbf{AUPC}  & \textbf{GR}   & \textbf{PR}   & \textbf{SR}   & \textbf{AUPC}   \\ 
\hline\hline
\multicolumn{13}{c}{\textbf{ALFWorld}}                                                                                                                                                                                               \\ 
\midrule
\name(s=1, k=1)               & 84.9          & 68.0          & 55.2          & 0.464          & 85.6          & 67.6          & 53.8          & 0.460          & 83.6          & \textbf{55.1} & \textbf{29.8} & \textbf{0.369}  \\
\name(s=3, k=1)               & 84.0          & 64.0          & 48.4          & 0.444          & \textbf{86.4} & 66.9          & 56.6          & 0.448          & 83.8          & 52.5          & 25.4          & 0.354           \\
\name(s=1, k=3)               & 83.1          & 62.1          & 47.0          & 0.423          & 85.8          & 68.0          & 57.4          & 0.459          & 83.4          & 52.2          & 26.1          & 0.349           \\
\name(s=3, k=3)               & \textbf{88.7} & \textbf{70.2} & \textbf{57.4} & \textbf{0.481} & 83.8          & \textbf{69.2} & \textbf{56.7} & \textbf{0.468} & \textbf{84.4} & 52            & 26.9          & 0.361           \\ 
\hline\hline
\multicolumn{13}{c}{\textbf{BabyAI}}                                                                                                                                                                                                 \\ 
\midrule
\name(s=1, k=1)               & 66.7          & \textbf{50.0} & \textbf{31.2} & 0.158          & \textbf{73.9} & \textbf{59.4} & \textbf{45.5} & 0.254          & 89.5          & 57.6          & 41.1          & \textbf{0.248}  \\
\name(s=3, k=1)               & 69.5          & 46.2          & 27.7          & 0.166          & 73.4          & 52.0          & 39.3          & 0.229          & 90.1          & 52.8          & 38.4          & 0.198           \\
\name(s=1, k=3)               & 66.6          & 46.1          & 28.6          & 0.156          & 72.8          & 57.7          & 44.7          & \textbf{0.255} & 89.4          & \textbf{56.7} & \textbf{42.9} & 0.222           \\
\name(s=3, k=3)               & \textbf{73.9} & 49.3          & 30.4          & \textbf{0.173} & 72.5          & 52.2          & 38.4          & 0.206          & \textbf{90.6} & 52.0          & 37.5          & 0.204           \\ 
\hline\hline
\multicolumn{13}{c}{\textbf{ScienceWorld}}                                                                                                                                                                                           \\ 
\midrule
\name(s=1, k=1)               & 16.1          & 44.8          & 23.4          & 0.277          & 13.6          & 39.3          & 13.4          & 0.241          & 24.4          & 63.2          & 34.6          & 0.416           \\
\name(s=3, k=1)               & 14            & 39.6          & 14.4          & 0.245          & \textbf{14.6} & \textbf{45.2} & 14.8          & 0.299          & 24.4          & 66.5          & 34.7          & 0.437           \\
\name(s=1, k=3)               & \textbf{16.1} & \textbf{46.7} & \textbf{23.4} & \textbf{0.298} & 13.3          & 37.7          & 11.1          & 0.242          & 25.3          & 67.3          & \textbf{40.2} & 0.442           \\
\name(s=3, k=3)               & 15            & 42.4          & 20            & 0.282          & 13.9          & 40.8          & \textbf{16.6} & \textbf{0.267} & \textbf{26.2} & \textbf{68.3} & 38            & \textbf{0.443}  \\
\bottomrule
\end{tabular}}
\end{table*}

\paragraph{Supplementary Results for Figure 6.}
Table~\ref{tab:appendix_sensitive_sampling} reports a sensitivity analysis of the number of sampled paths used in \name. Across all three environments, performance remains stable over a range of sampling sizes, with 6 samples often yielding the best trade-off between efficiency and effectiveness. For example, in ScienceWorld with GPT-4o-mini, sampling 6 paths achieves the highest Success Rate (40.2) and AUPC (0.442). These results suggest that moderate sampling is sufficient to capture high-utility skills, and that our method is robust to the sampling budget.

\subsection{Sensitive Analysis on Sampling Temperature.}
\paragraph{Sampling Temperature Sensitivity.}
Table~\ref{tab:appendix_sensitive_temperature} reports the effect of varying the sampling temperature when generating training trajectories for \name. A higher temperature encourages more diverse trajectory sampling, which can uncover additional high-utility skills. On ALFWorld, increasing the temperature to $0.7$ significantly boosts performance, achieving the highest Success Rate (56.7) and AUPC (0.485). However, on BabyAI and ScienceWorld, the best results are obtained at $T{=}1.0$, indicating that increased diversity is especially beneficial in tasks requiring fine-grained reasoning or multi-step composition. These results suggest that controlled sampling diversity improves the coverage of decision-relevant behaviors and enhances downstream prompt quality, supporting more effective skill extraction and reuse.

\subsection{Sensitive Analysis on Skill Retrieval}
\paragraph{Supplementary Results for Figure 6.}  
Table~\ref{tab:appendix_sensitive_retrieval} presents a sensitivity analysis of \textit{\name} under different retrieval configurations, varying the number of sampled paths ($s$) and the number of top-$k$ skills retrieved per step. Overall, the setting $s{=}3$, $k{=}3$ yields the strongest performance on ALFWorld and ScienceWorld, while $s{=}1$, $k{=}1$ performs best on BabyAI. These results indicate that focused retrieval is effective even with minimal samples, but leveraging multiple paths and top-$k$ retrieval further enhances performance in more complex environmentsg.

\subsection{Token Cost Comparison\label{appendix:tokens}}

\begin{table*}[!t]
\centering
\caption{Token Efficiency and Task Progress Comparison. Progress Rate [\%] $(\uparrow$), Success Rate [\%] ($\uparrow$),  average number of tokens used per task in thousands (Avg. kTok) ($\downarrow$), and progress increase per 1,000 tokens (PR / kTok) [\%] ($\uparrow$). The best method for each  LLM is in \textbf{bold}; the second-best method is \uline{underlined}.~\label{tab:appendix_token}}
\resizebox{1\textwidth}{!}{\begin{tabular}{c|cccc|cccc|cccc} 
\toprule
\multirow{2}{*}{\textbf{Method} } & \multicolumn{4}{c|}{\textbf{Qwen2.5-7B-Instruct} }                          & \multicolumn{4}{c|}{\textbf{Qwen-Turbo} }                          & \multicolumn{4}{c}{\textbf{GPT-4o-mini} }                           \\
                                 & \textbf{PR}   & \textbf{SR}   & \textbf{Avg. kTok } & \textbf{PR / kTok} & \textbf{PR}   & \textbf{SR}   & \textbf{Avg. kTok } & \textbf{PR / kTok} & \textbf{PR}   & \textbf{SR}   & \textbf{Avg. kTok } & \textbf{PR / kTok} \\
\hline\hline
\multicolumn{13}{c}{\textbf{ALFWorld} }                                                \\ 
\midrule
Leap                             & 21.2          & 5.2           & \uline{20.2}           & 1.05                & 55.6          & 37.3          & 16.8                 & 3.3                 & 50.8          & 11.2          & 21.6                 & 2.3                 \\
Synapse (1-shot)                 & 41.6          & 17.1          & 22.6                 & 1.84                & 54.7          & 35.8          & \uline{16.0}           & \uline{3.4}           & 48.8          & 14.8          & \uline{17.2}           & \uline{2.8}           \\
Synapse (3-shot)                 & \uline{44.8}    & 19.4          & 41.2                 & 1.09                & \uline{60.6}    & \uline{47.0}    & 25.2                 & 2.4                 & \uline{52.9}    & \uline{17.8}    & 28.0                 & 1.9                 \\
Trad                             & 44.2          & \uline{22.4}    & 23.5                 & \uline{1.88}          & 54.8          & 35.8          & 17.1                 & 3.2                 & 49.1          & 16.4          & 23.5                 & 2.1                 \\
\name                         & \textbf{68.0} & \textbf{55.2} & \textbf{12.5}        & \textbf{5.45}       & \textbf{67.6} & \textbf{53.8} & \textbf{11.4}        & \textbf{5.9}        & \textbf{55.1} & \textbf{29.8} & \textbf{15.7}        & \textbf{3.5}        \\

\hline\hline
\multicolumn{13}{c}{\textbf{BabyAI} }                                          \\ 
\midrule
Leap                             & \uline{46.3}    & 27.7          & 31.2                 & 1.5                 & 52.9          & 38.4          & 26.0                 & 2.0                 & 43.8          & 29.4          & 30.1                 & 1.5                 \\
Synapse (1-shot)                 & 39.4          & 21.4          & 24.5                 & \uline{1.6}           & \uline{55.8}    & \uline{45.5}    & \uline{16.5}           & \uline{3.4}           & 44.9          & 33.9          & \uline{19.7}           & \uline{2.3}           \\
Synapse (3-shot)                 & 44.6          & \uline{28.6}    & 42.8                 & 1.0                 & 50.8          & 38.4          & 28.2                 & 1.8                 & \uline{49.5}    & \uline{38.4}    & 32.0                 & 1.5                 \\
Trad                             & 36.9          & 19.7          & \uline{28.8}           & 1.3                 & 46.9          & 34.8          & 22.1                 & 2.1                 & 40.9          & 30.4          & 23.4                 & 1.7                 \\
\name                         & \textbf{50.0} & \textbf{31.2} & \textbf{15.4}        & \textbf{3.2}        & \textbf{59.4} & \textbf{45.5} & \textbf{11.9}        & \textbf{5.0}        & \textbf{57.6} & \textbf{41.1} & \textbf{14.0}        & \textbf{4.1}        \\

\hline\hline
\multicolumn{13}{c}{\textbf{ScienceWorld} }                                                                                        \\ 
\midrule
Leap                             & 25.8          & \uline{11.1}    & 43.4                 & 0.6                 & 21.8          & 6.6           & 40.7                 & 0.5                 & 51.7          & 21.1          & 44.0                 & 1.2                 \\
Synapse (1-shot)                 & 15.4          & 4.4           & \uline{22.6}           & \uline{0.7}           & 16.0          & 3.3           & \uline{20.4}           & 0.8                 & 52.8          & 25.6          & \uline{21.0}           & \uline{2.5}           \\
Synapse (3-shot)                 & 15.3          & 5.5           & 39.6                 & 0.4                 & 24.0          & \uline{10.1}    & 32.6                 & 0.7                 & 60.0          & \uline{32.4}    & 31.9                 & 1.9                 \\
Trad                             & 21.1          & 4.4           & 32.6                 & 0.6                 & \uline{29.3}    & 8.8           & 27.6                 & \uline{1.1}           & \uline{61.4}    & 29.0          & 26.8                 & 2.3                 \\
\name                         & \textbf{46.7} & \textbf{23.4} & \textbf{15.9}        & \textbf{2.9}        & \textbf{37.7} & \textbf{11.1} & \textbf{16.1}        & \textbf{2.3}        & \textbf{67.3} & \textbf{40.2} & \textbf{16.1}        & \textbf{4.2}       \\
\bottomrule
\end{tabular}
}
\end{table*}

In Table~\ref{tab:appendix_token}, we examine token efficiency, measured as the progress rate achieved per 1,000 tokens used (PR/kTok). This metric reflects how effectively each method translates language model computation into task progress. \name consistently achieves the highest token efficiency across all benchmarks and models. On ALFWorld, \name reaches 5.9 PR/kTok with Qwen-Turbo, outperforming Synapse (1-shot: 3.4, 3-shot: 2.4) and Trad (3.2), while using only ~11.4k tokens per task, compared to Synapse’s 25k–41k. A similar pattern appears on BabyAI, where \name achieves 5.0 PR/kTok (Qwen-Turbo), compared to 3.4 for Synapse (1-shot) and 1.8 for Synapse (3-shot). The advantage is most pronounced on the challenging ScienceWorld domain. Here, \name achieves 4.2 PR/kTok on GPT-4o-mini, which is 1.68× higher than the best-performing baseline, Synapse (1-shot, 2.5 PR/kTok), and far ahead of Synapse (3-shot, 1.9), despite their higher token consumption. Notably, increasing the number of in-context demonstrations leads to diminishing efficiency returns: although Synapse (3-shot) uses over 31k tokens per task, its PR/kTok consistently drops compared to the 1-shot variant.

These findings demonstrate that \name not only excels in performance but delivers the best return on token usage, making it the most cost-effective solution among all evaluated methods.

\subsection{Sampling Strategies for TD-Based Credit Assignment\label{appendix:td_sampling}}

To explore different sampling strategies for TD-based credit assignment, we consider not only uniform sampling but also a weighted alternative, where trajectory selection probabilities are derived from empirical progress deltas. For a given path \( \tau = (a_0, a_1, \dots, a_T) \), we compute its score as the cumulative sum of edge-level progress:
$\text{score}(\tau) = \sum_{t=0}^{T-1} \bar{\Delta}(a_t, a_{t+1})$,
where  $\bar{\Delta}(a_t, a_{t+1})$  denotes the empirical mean of recorded progress deltas for transition $(a_t, a_{t+1})$. To convert scores into a probabilistic sampling distribution, we apply a numerically stable softmax:
$p_i = \frac{\exp(\text{score}(\tau_i) - \max_j \text{score}(\tau_j))}{\sum_j \exp(\text{score}(\tau_j) - \max_k \text{score}(\tau_k))}$,
allowing high-score paths to be prioritized while preserving nonzero probability for exploration. Sampled paths are then drawn without replacement from this weighted distribution.

We compare the performance of \name under two sampling strategies—weighted sampling and uniform sampling. All results are conducted using a single retrieved skill (s=1) and top-1 antecedent and consequence (k=1). In Table~\ref{tab:sampling_strategy}, uniform sampling consistently outperforms weighted sampling across all three tasks in PR, SR, and AUPC. This finding suggests that while prioritizing high-scoring trajectories can potentially focus learning on informative regions of the search space, it may also amplify noise or overfit to progress signals—especially when empirical deltas are sparse or unevenly distributed. By treating all observed transitions equally, uniform sampling promotes broader trajectory diversity and more stable value propagation, which appears crucial for generalization and long-horizon task completion.

\begin{table*}[!t]
\centering
\caption{Comparison of \name with weighted vs. uniform sampling strategies under the Qwen-7B-Instruct model with $s=1$, $k=1$.\label{tab:sampling_strategy}}
\resizebox{1\textwidth}{!}{
\begin{tabular}{c|cccc|cccc|cccc} 
\toprule
                            & \multicolumn{4}{c|}{\textbf{AlFWorld}}                         & \multicolumn{4}{c|}{\textbf{BabyAI}}                           & \multicolumn{4}{c}{\textbf{ScienceWorld}}                       \\
\textbf{Method}             & \textbf{GR}   & \textbf{PR}   & \textbf{SR}   & \textbf{AUPC}  & \textbf{GR}   & \textbf{PR}   & \textbf{SR}   & \textbf{AUPC}  & \textbf{GR}   & \textbf{PR}   & \textbf{SR}   & \textbf{AUPC}   \\ 
\midrule
\name (Weighted Sampling) & 82.9          & 67            & 53            & 0.455          & \textbf{66.9} & 47.4          & 27.6          & 0.142          & 15.2          & 41.7          & 21.1          & 0.259           \\
\name  (Uniform Sampling)  & \textbf{84.9} & \textbf{68.0} & \textbf{55.2} & \textbf{0.464} & 66.7          & \textbf{50.0} & \textbf{31.2} & \textbf{0.158} & \textbf{16.1} & \textbf{44.8} & \textbf{23.4} & \textbf{0.277}  \\
\bottomrule
\end{tabular}}
\end{table*}

\begin{table*}[!t]
\centering
\small
\caption{Comparison between Act-style (\textit{observation $\rightarrow$ action}) and React-style (\textit{observation $\rightarrow$ thought $\rightarrow$ action}) prompting strategies for \name in ScienceWorld using GPT-4o-mini.\label{tab:react_vs_act}}
\resizebox{0.4\textwidth}{!}{
\begin{tabular}{ccccc}
\toprule
\multirow{2}{*}{\textbf{Method}} & \multicolumn{4}{c}{\textbf{GPT-4o-mini}}                       \\
                                 & \textbf{GR}   & \textbf{PR}   & \textbf{SR}   & \textbf{AUPC}  \\
\midrule
\name(Act)                    & \textbf{25.3} & 67.3          & 40.2          & 0.442          \\
\name(React)                  & 23            & \textbf{70.1} & \textbf{43.4} & \textbf{0.451}\\
\bottomrule
\end{tabular}}
\end{table*}

\subsection{Act / ReAct for \name Inference}

Our main experiments use a minimalist \textit{Act-style} prompting framework (\textit{observation $\rightarrow$ action}), which allows us to directly evaluate the decision-making abilities of LLMs without relying on additional components such as memory modules or external tools. This lightweight design makes the framework broadly applicable to most instruction-following models and helps us focus on the model’s intrinsic inference capabilities, without interference from extra reasoning mechanisms.

To further examine the impact of explicit reasoning structure, we conduct a controlled comparison on the ScienceWorld dataset using \texttt{GPT-4o-mini}. In this setting, we evaluate \name under the \textit{React-style} prompting format, which extends the Act-style formulation (\textit{observation $\rightarrow$ action}) with an intermediate reasoning step (\textit{observation $\rightarrow$ thought $\rightarrow$ action}). To ensure a fair comparison, we limit the total number of action steps to 20 for both settings. For Act-style, each decision is made in a single inference step. For React-style, each decision comprises two inference steps—one for generating the thought and one for producing the final action—resulting in a total of 20 thoughts and 20 actions per task. As shown in Table~\ref{tab:react_vs_act}, React-style prompting improves progress rate (70.1 vs. 67.3), success rate (43.4 vs. 40.2), and AUPC (0.451 vs. 0.442), while incurring a slight drop in goal completion (23.0 vs. 25.3). These results suggest that incorporating intermediate thoughts can enhance local decision quality and stabilize planning, although at the cost of increased prompt complexity.

\section{Analysis on Domain Knowledge Graphs\label{appendix:domin_graph}}
Tables~\ref{tab:appendix_ac_graph_qwen7b}–\ref{tab:appendix_ac_graph_gpt4omini} summarize the structural statistics of the action-centric domain graphs constructed from sampled trajectories using different base models—Qwen2.5-7B-Instruct, Qwen-Turbo, and GPT-4o-mini—across the three task domains: ALFWorld, BabyAI, and ScienceWorld. For each setting, we vary the number of sampled trajectories per task (\#Sampling) from 3 to 12, and report the average number of nodes, edges, and the average node degree in the resulting graph.

We observe that increasing the number of sampled trajectories consistently leads to denser graphs, reflected by higher node counts and edge connectivity. This trend is especially pronounced in the ALFWorld domain, which exhibits the largest graph size and connectivity across all models. In contrast, BabyAI and ScienceWorld produce smaller graphs, but still show steady growth in structural complexity with more samples. Across models, GPT-4o-mini yields the most extensive graphs, suggesting that its behavior leads to more diverse action coverage during trajectory sampling. These statistics provide insight into the richness of the induced action space and the granularity of skills extracted during graph-based knowledge construction.

\begin{table*}[!t]
\centering
\small
\caption{Action-Centric Graphs Statistics of Qwen2.5-7B-Instruct.\label{tab:appendix_ac_graph_qwen7b}}
\resizebox{0.7\textwidth}{!}{
\begin{tabular}{c|c|ccc}
\toprule
\textbf{Dataset} & \textbf{\#Sampling} & \textbf{Avg \#Nodes} & \textbf{Avg \#Edges} & \textbf{Avg Degree} \\
\midrule
\multirow{4}{*}{ALFWorld} & 3 & 17.46 & 34.0 & 3.8 \\
& 6 & 21.71 & 46.79 & 4.27 \\
& 9 & 24.62 & 59.17 & 4.89 \\
& 12 & 27.12 & 69.33 & 5.17 \\
\hline
\multirow{4}{*}{BabyAI} & 3 & 7.81 & 13.71 & 3.27 \\
& 6 & 9.05 & 18.54 & 3.93 \\
& 9 & 9.78 & 22.03 & 4.36 \\
& 12 & 10.12 & 24.1 & 4.63 \\
\hline
\multirow{4}{*}{ScienceWorld} & 3 & 6.92 & 10.68 & 2.59 \\
& 6 & 8.49 & 14.7 & 3.17 \\
& 9 & 9.13 & 17.73 & 3.6 \\
& 12 & 9.71 & 19.46 & 3.74 \\
\bottomrule
\end{tabular}}
\end{table*}

\begin{table*}[!t]
\centering
\small
\caption{Action-Centric Graphs  Statistics of Qwen-Turbo.\label{tab:appendix_ac_graph_qwenturbo}}
\resizebox{0.7\textwidth}{!}{
\begin{tabular}{c|c|ccc}
\toprule
\textbf{Dataset} & \textbf{\#Sampling} & \textbf{Avg \#Nodes} & \textbf{Avg \#Edges} & \textbf{Avg Degree} \\
\midrule
\multirow{4}{*}{ALFWorld} & 3 & 28.67 & 60.71 & 4.27 \\
& 6 & 33.21 & 77.21 & 4.67 \\
& 9 & 35.04 & 89.08 & 5.04 \\
& 12 & 36.42 & 98.5 & 5.34 \\
\hline
\multirow{4}{*}{BabyAI} & 3 & 9.88 & 15.44 & 2.9 \\
& 6 & 11.69 & 22.36 & 3.61 \\
& 9 & 12.09 & 24.17 & 3.75 \\
& 12 & 12.22 & 25.29 & 3.86 \\
\hline
\multirow{4}{*}{ScienceWorld} & 3 & 8.52 & 12.35 & 2.41 \\
& 6 & 10.54 & 17.43 & 2.82 \\
& 9 & 12.11 & 21.38 & 3.03 \\
& 12 & 12.9 & 23.98 & 3.2 \\
\bottomrule
\end{tabular}}
\end{table*}

\begin{table*}[!t]
\centering
\small
\caption{Action-Centric Graphs Statistics of GPT-4o-mini.\label{tab:appendix_ac_graph_gpt4omini}}
\resizebox{0.7\textwidth}{!}{
\begin{tabular}{c|c|ccc}
\toprule
\textbf{Dataset} & \textbf{\#Sampling} & \textbf{Avg \#Nodes} & \textbf{Avg \#Edges} & \textbf{Avg Degree} \\
\midrule
\multirow{4}{*}{ALFWorld} & 3 & 30.71 & 75.38 & 4.84 \\
& 6 & 39.29 & 105.08 & 5.31 \\
& 9 & 43.79 & 126.67 & 5.79 \\
& 12 & 46.04 & 138.25 & 5.99 \\
\hline
\multirow{4}{*}{BabyAI} & 3 & 9.08 & 15.95 & 3.03 \\
& 6 & 11.04 & 22.48 & 3.75 \\
& 9 & 11.46 & 23.88 & 3.9 \\
& 12 & 11.76 & 24.84 & 3.98 \\
\hline
\multirow{4}{*}{ScienceWorld} & 3 & 13.13 & 23.94 & 3.46 \\
& 6 & 16.22 & 33.6 & 3.94 \\
& 9 & 18.29 & 41.19 & 4.26 \\
& 12 & 20.27 & 47.52 & 4.43 \\
\bottomrule
\end{tabular}}
\end{table*}




\section{Prompt Examples\label{appendix:examples}}

We present several examples to illustrate the prompt structure.The step-wise skill section displays only the skill retrieved for the final action.

\begin{tcolorbox}[
  enhanced,
  breakable,
  colback=blue!3!white,
  colframe=blue!60!black,
  coltitle=black,
  title=\faBook\hspace{0.5em} \name Prompt Template,
  fonttitle=\bfseries,
  colbacktitle=blue!10!white,
  sharp corners,
  boxrule=0.5pt,
  left=4pt,
  right=4pt,
  top=4pt,
  bottom=4pt,
]

\{task\_description\}

\vspace{0.75em}
\textbf{\#\# Golden Segment (What to Imitate)}

Here is a related action sequence, which may not be fully accurate, but help identify promising directions:

\{golden\_segment\}

\vspace{0.75em}
\textbf{\#\# Step-wise Reusable Skills (Context-Aware Guidance)}

These skills are relevant to the current context based on your most recent action.

They suggest promising steps to explore the environment:  
\{step\_wise\_skills\}

\vspace{0.75em}
\textbf{\#\# Instruction}

You should use the following commands for help when your action cannot be understood: \texttt{check valid actions}.

You should use the following commands for help when your action cannot be understood: \texttt{inventory}.  

Generate the \textbf{next best action} to reach the goal.

\textbf{Goal}: \{goal\}

\{history\_transitions\}

\textbf{Action}:
\end{tcolorbox}

\begin{tcolorbox}[
  enhanced,
  breakable,
  colback=blue!3!white,
  colframe=blue!60!black,
  coltitle=black,
  title=\faRobot\hspace{0.5em} \name Prompt Example (ALFWorld),
  fonttitle=\bfseries,
  colbacktitle=blue!10!white,
  sharp corners,
  boxrule=0.5pt,
  left=4pt,
  right=4pt,
  top=4pt,
  bottom=4pt,
]

Your task is to interact with a virtual household simulator to accomplish a specific task. With each interaction, you will receive an observation.Your role is to decide on an action based on the observation

\vspace{0.5em}
\textbf{Golden Segment (What to Imitate):}

Here is a related demonstration, which may not be fully accurate, but help identify promising directions:

\textit{Goal: examine the alarmclock with the desklamp}

You are in the middle of a room. Looking quickly around you, you see a bed 1, a desk 2, a desk 1, a drawer 6, a drawer 5, a drawer 4, a drawer 3, a drawer 2, a drawer 1, a garbagecan 1, a laundryhamper 1, a safe 1, a shelf 6, a shelf 5, a shelf 4, a shelf 3, a shelf 2, and a shelf 1.

Your task is to: examine the alarmclock with the desklamp.

\texttt{ACTION: go to desk 1} \\
\texttt{ACTION: check valid actions} \\
\texttt{ACTION: use desklamp 1} \\
\texttt{ACTION: go to desk 2} \\
\texttt{ACTION: check valid actions} \\
\texttt{ACTION: take alarmclock 1 from desk 2}

\vspace{0.5em}
\textbf{Step-wise Reusable Skills (Context-Aware Guidance):}

These skills are relevant to the current context based on your most recent action.
They suggest promising steps to explore the environment:

\textbf{Skill 1:} Centered on action \texttt{'go to desk'}

\begin{itemize}
  \item \textbf{Common precursors:} \texttt{check valid actions}
  \item \textbf{Typical next steps:} \texttt{use desklamp}
\end{itemize}

\vspace{0.5em}
\textbf{Instruction:}
\begin{itemize}
  \item Use \texttt{check valid actions} if your action is not understood.
  \item Use \texttt{inventory} to see what you are holding.
  \item Generate the \textbf{next best action} to reach the goal.
\end{itemize}

\vspace{0.5em}
\textbf{Interaction History:}

\textbf{Goal:} Examine the bowl with the desklamp

\begin{itemize}
\item \texttt{ACTION: go to desk 1} \\
\texttt{OBSERVATION:} On the desk 1, you see a alarmclock 2, a alarmclock 1, a desklamp 1, a mug 1, and a pencil 1.

\item \texttt{ACTION: check valid actions} \\
\texttt{OBSERVATION:} Choose from: examine desk 1, go to bed 1, go to desk 2, go to drawer 1--6, go to garbagecan 1, laundryhamper 1, safe 1, shelf 1--6, inventory, look, take alarmclock 1/2, take mug 1, take pencil 1, use desklamp 1.

\item \texttt{ACTION: go to desk 2} \\
\texttt{OBSERVATION:} On the desk 2, you see a bowl 3, a bowl 2, and a mug 2.

\item \texttt{ACTION: check valid actions} \\
\texttt{OBSERVATION:} Choose from: examine desk 2, go to other objects, inventory, look, take bowl 2/3, take mug 2.

\item \texttt{ACTION: take bowl 2 from desk 2} \\
\texttt{OBSERVATION:} You pick up the bowl 2 from the desk 2.

\item \texttt{ACTION: go to desk 1} \\
\texttt{OBSERVATION:} On the desk 1, you see a alarmclock 2, a alarmclock 1, a desklamp 1, a mug 1, and a pencil 1.

\item \textbf{ACTION: \texttt{use desklamp 1}} \hfill \textit{Progress Rate: 1.0}
\end{itemize}

\textit{Note: This example is provided to illustrate the overall prompt structure; the step-wise skill section only displays the skill retrieved for the final action.}
\end{tcolorbox}

\begin{tcolorbox}[
  enhanced,
  breakable,
  colback=blue!3!white,
  colframe=blue!60!black,
  coltitle=black,
  title=\faRobot\hspace{0.5em} \name Prompt Example (BabyAI),
  fonttitle=\bfseries,
  colbacktitle=blue!10!white,
  sharp corners,
  boxrule=0.5pt,
  left=4pt,
  right=4pt,
  top=4pt,
  bottom=4pt,
]

You are placed in a room and you need to accomplish the given goal with actions.

\vspace{0.5em}
\textbf{Golden Segment (What to Imitate):}

Here is a related action sequence, which may not be fully accurate, but help identify promising directions:

\textit{Goal: open the blue door}

In front of you in this room, you can see several objects: There is a blue closed door 1 1 steps in front of you and 1 steps to your left. The room has walls around you. You are facing a wall 2 steps away. You are not carrying anything.

\texttt{ACTION: go to blue closed door 1} \\
\texttt{ACTION: toggle and go through blue closed door 1} \\
\texttt{ACTION: move forward} \\

\vspace{0.5em}
\textbf{Step-wise Reusable Skills (Context-Aware Guidance):}

These skills are relevant to the current context based on your most recent action.
They suggest promising steps to explore the environment:

\textbf{Skill 1:} Centered on action \texttt{'move forward'}

\begin{itemize}
  \item \textbf{Common precursors:} \texttt{toggle and go through blue closed door}
  \item \textbf{Typical next steps:} \texttt{turn left}
\end{itemize}

\vspace{0.5em}
\textbf{Instruction:}
\begin{itemize}
  \item Use \texttt{check valid actions} if your action is not understood.
  \item Use \texttt{inventory} to see what you are holding.
  \item Generate the \textbf{next best action} to reach the goal.
\end{itemize}

\vspace{0.5em}
\textbf{Interaction History:}

\textbf{Goal:} Open the red door, and open the blue door

\begin{itemize}
\item \texttt{ACTION: check valid actions} \\
\texttt{OBSERVATION:} You can take the following actions: turn left, turn right, check available actions.

\item \texttt{ACTION: go to blue closed door 1} \\
\texttt{OBSERVATION:} The action is not recognized. Please check valid actions.

\item \texttt{ACTION: turn left} \\
\texttt{OBSERVATION:} There is a red closed door 1 4 steps in front of you and 3 steps to your left.

\item \texttt{ACTION: go to red closed door 1} \\
\texttt{OBSERVATION:} Red closed door 1 is right in front of you 1 step away.

\item \texttt{ACTION: toggle and go through red closed door 1} \\
\texttt{OBSERVATION:} You are facing a wall 5 steps away.

\item \texttt{ACTION: move forward} \\
\texttt{OBSERVATION:} You are facing a wall 4 steps away.

\item \texttt{ACTION: turn left} \\
\texttt{OBSERVATION:} There is a red open door 1 0 steps in front of you and 1 step to your left.

\item \texttt{ACTION: go to blue closed door 1} \\
\texttt{OBSERVATION:} The action is not recognized.

\item \texttt{ACTION: go to blue closed door 1} \\
\texttt{OBSERVATION:} The action is not recognized.

\item \texttt{ACTION: check valid actions} \\
\texttt{OBSERVATION:} You can take the following actions: turn left, turn right, go through red open door 1, go to red open door 1.

\item \texttt{ACTION: go through red open door 1} \\
\texttt{OBSERVATION:} There is a blue closed door 1 5 steps in front of you and 2 steps to your left.

\item \texttt{ACTION: go to blue closed door 1} \\
\texttt{OBSERVATION:} Blue closed door 1 is right in front of you 1 step away.

\item \texttt{ACTION: toggle and go through blue closed door 1} \\
\texttt{OBSERVATION:} You are facing a wall 5 steps away.

\item \texttt{ACTION: move forward} \\
\texttt{OBSERVATION:} You are facing a wall 4 steps away.

\item \textbf{ACTION: \texttt{turn left}} \hfill \textit{Progress Rate: 1.0}

\textit{Note: This example is provided to illustrate the overall prompt structure; the step-wise skill section only displays the skill retrieved for the final action.}
\end{itemize}

\end{tcolorbox}

\begin{tcolorbox}[
  enhanced,
  breakable,
  colback=blue!3!white,
  colframe=blue!60!black,
  coltitle=black,
  title=\faRobot\hspace{0.5em} \name Prompt Example (ScienceWorld),
  fonttitle=\bfseries,
  colbacktitle=blue!10!white,
  sharp corners,
  boxrule=0.5pt,
  left=4pt,
  right=4pt,
  top=4pt,
  bottom=4pt,
]

You are an agent in a virtual science school environment, tasked to interact with various elements.

\vspace{0.5em}
\textbf{Golden Segment (What to Imitate):}

Here is a related action sequence, which may not be fully accurate, but help identify promising directions:

\textit{Goal: Find butterfly egg in the outside. First, focus on the thing. Then, move it to the blue box in the bedroom.}

\texttt{This room is called the hallway. In it, you see:} air, a painting, doors to the art studio, bedroom, greenhouse, kitchen, living room, and workshop (all open).\\
\texttt{Inventory:} an orange.

\texttt{ACTION: go to greenhouse} \\
\texttt{ACTION: look around} \\
\texttt{ACTION: look around} \\
\texttt{ACTION: go to outside} \\
\texttt{ACTION: look around} \\
\texttt{ACTION: focus on butterfly egg} \\
\texttt{ACTION: pick up butterfly egg} \\
\texttt{ACTION: move butterfly egg to blue box}

\vspace{0.5em}
\textbf{Step-wise Reusable Skills (Context-Aware Guidance):}

These skills are relevant to the current context based on your most recent action.
They suggest promising steps to explore the environment:

\textbf{Skill 1:} Centered on action \texttt{'look around'}

\begin{itemize}
  \item \textbf{Common precursors:} \texttt{go to outside}
  \item \textbf{Typical next steps:} \texttt{move butterfly egg to blue box}
\end{itemize}

\vspace{0.5em}
\textbf{Instruction:}
\begin{itemize}
  \item Use \texttt{check valid actions} if your action is not understood.
  \item Use \texttt{inventory} to inspect your current items.
  \item Generate the \textbf{next best action} to complete the goal.
\end{itemize}

\vspace{0.5em}
\textbf{Interaction History:}

\textbf{Goal:} Find a dove egg in the outside. First, focus on the thing. Then, move it to the red box in the bathroom.

\begin{itemize}
\item \texttt{OBSERVATION (greenhouse):} air, multiple flower pots, jug, shovel, sink, open doors to hallway and outside \\
\texttt{Inventory:} an orange

\item \texttt{ACTION: go to outside} \\
\texttt{OBSERVATION:} You move to the outside.

\item \texttt{ACTION: look around} \\
\texttt{OBSERVATION:} You see air, an axe, blue jay egg, butterfly egg, dove egg, fire pit, fountain, ground, wood, and open doors to foundry, greenhouse, kitchen.

\item \texttt{ACTION: focus on dove egg} \\
\texttt{OBSERVATION:} You focus on the dove egg.

\item \texttt{ACTION: pick up dove egg} \\
\texttt{OBSERVATION:} You move the dove egg to the inventory.

\item \texttt{ACTION: go to bathroom} \\
\texttt{OBSERVATION:} No known action matches that input.

\item \texttt{ACTION: go to hallway} \\
\texttt{OBSERVATION:} No known action matches that input.

\item \texttt{ACTION: go to kitchen} \\
\texttt{OBSERVATION:} You move to the kitchen.

\item \texttt{ACTION: look around} \\
\texttt{OBSERVATION:} air, chair, counter, cupboard, freezer, fridge, jar of sodium chloride, lighter, oven, painting, sink, soap, stopwatch, stove, table, thermometer; open doors to bathroom, hallway, and outside.

\item \texttt{ACTION: go to bathroom} \\
\texttt{OBSERVATION:} You move to the bathroom.

\item \texttt{ACTION: look around} \\
\texttt{OBSERVATION:} air, bathtub, glass cup, painting, red box (empty), sink, toilet (with water and a closed drain)

\item \textbf{ACTION: \texttt{move dove egg to red box}} \hfill \textit{Progress Rate: 1.0}
\end{itemize}

\vspace{0.5em}
\textit{Note: This example is provided to illustrate the overall prompt structure; the step-wise skill section only displays the skill retrieved for the final action.}
\end{tcolorbox}

\vspace{-1.5em}

\section{Case Study\label{appendix:case_study}}

Here, we present two cases to illustrate how skill-enhanced prompting guides LLMs through structured, multi-step decision-making. 

Skill knowledge, distilled from past interactions, captures reusable patterns of behavior that are both interpretable and broadly transferable across tasks. Case A presents a manipulation-centered task in ALFWorld,  where the agent must retrieve and place two peppershakers into a drawer. The trajectory demonstrates how skill-enhanced prompting enables modular and recoverable decision-making in object-centric settings. Early in the task, the agent execute “go to countertop” to explore items. Upon discovering the first peppershaker, the agent reuses the structured routine of “take object,” followed by “go to drawer,” “open drawer,” and “put object”—a sequence grounded in manipulation skills that apply broadly across object types and locations. When the first peppershaker is placed in the drawer, the agent invokes the same search pattern on another countertop, demonstrating flexible reuse of skills under uncertainty. Throughout the trajectory, the agent follows interpretable subgoal transitions aligned with learned affordances, such as only placing items after opening drawers. The agent’s ability to recover from failure, recompose subroutines, and generalize across instances of “peppershaker” and “drawer” reflects the power of skills as mid-level priors. This example further affirms that skill-driven prompting enables compositional manipulation planning, supporting both action precision and long-horizon success in LLM-based sequential decision making.

\begin{tcolorbox}[
  enhanced,
  breakable,
  colback=white,
  colframe=blue!60!black,
  coltitle=black,
  title=\faTasks\hspace{0.5em} Case A: Put Two Peppershakers in Drawer,
  label={box:skill-traj-pp},
  fonttitle=\bfseries,
  colbacktitle=blue!10!white,
  sharp corners,
  boxrule=0.8pt,
  left=6pt,
  right=6pt,
  top=6pt,
  bottom=6pt,
]\label{box:skill-traj-alf}

\textbf{Goal:} put two peppershaker in drawer\\
\textbf{Progress Rate:} 1.0

\vspace{0.5em}
\textbf{OBS:} You are in the middle of a room. Looking quickly around you, you see a cabinet 6, a cabinet 5, a cabinet 4, a cabinet 3, a cabinet 2, a cabinet 1, a coffeemachine 1, a countertop 3, a countertop 2, a countertop 1, a drawer 3, a drawer 2, a drawer 1, a fridge 1, a garbagecan 1, a microwave 1, a shelf 3, a shelf 2, a shelf 1, a sinkbasin 1, a stoveburner 4, a stoveburner 3, a stoveburner 2, a stoveburner 1, and a toaster 1.\\
Your task is to: put two peppershaker in drawer.

\textbf{SKILL Centered on action 'the beginning of the task'}
\begin{itemize}
  \item Typical next steps after this action:
  \begin{itemize}
    \item After 'the beginning of the task', agents usually continue with 'go to shelf'.
  \end{itemize}
\end{itemize}
\textbf{ACTION:} go to shelf 1 \hfill \textbf{PROGRESS RATE:} 0.0

\textbf{OBS:} On the shelf 1, you see a statue 1.

\textbf{SKILL Centered on action 'go to shelf'}
\begin{itemize}
  \item Common precursors to this action:
  \begin{itemize}
    \item Agents often perform 'the beginning of the task' before 'go to shelf'.
  \end{itemize}
  \item Typical next steps after this action:
  \begin{itemize}
    \item After 'go to shelf', agents usually continue with 'go to countertop'.
  \end{itemize}
\end{itemize}

\textbf{ACTION:} go to countertop 1 \hfill \textbf{PROGRESS RATE:} 0.0

\textbf{OBS:} On the countertop 1, you see a apple 1, a creditcard 2, a egg 1, and a tomato 1.

\textbf{SKILL Centered on action 'go to countertop'}
\begin{itemize}
  \item Common precursors to this action:
  \begin{itemize}
    \item Agents often perform 'the beginning of the task' before 'go to countertop'.
  \end{itemize}
  \item Typical next steps after this action:
  \begin{itemize}
    \item After 'go to countertop', agents usually continue with 'take soapbottle from countertop'.
  \end{itemize}
\end{itemize}

\textbf{ACTION:} go to countertop 2 \hfill \textbf{PROGRESS RATE:} 0.25

\textbf{OBS:} On the countertop 2, you see a bread 1, a butterknife 1, a cellphone 1, a cup 3, a knife 1, a pan 1, a papertowelroll 1, a peppershaker 1, a saltshaker 1, and a spatula 1.

\textbf{ACTION:} take peppershaker 1 from countertop 2 \hfill \textbf{PROGRESS RATE:} 0.5

\textbf{OBS:} You pick up the peppershaker 1 from the countertop 2.

\textbf{SKILL Centered on action 'take peppershaker from countertop'}
\begin{itemize}
  \item Common precursors to this action:
  \begin{itemize}
    \item Agents often perform 'go to countertop' before 'take peppershaker from countertop'.
  \end{itemize}
  \item Typical next steps after this action:
  \begin{itemize}
    \item After 'take peppershaker from countertop', agents usually continue with 'go to countertop'.
  \end{itemize}
\end{itemize}

\textbf{ACTION:} go to drawer 1 \hfill \textbf{PROGRESS RATE:} 0.5

\textbf{OBS:} The drawer 1 is closed.

\textbf{SKILL Centered on action 'go to drawer'}
\begin{itemize}
  \item Common precursors to this action:
  \begin{itemize}
    \item Agents often perform 'go to countertop' before 'go to drawer'.
  \end{itemize}
  \item Typical next steps after this action:
  \begin{itemize}
    \item After 'go to drawer', agents usually continue with 'open drawer'.
  \end{itemize}
\end{itemize}

\textbf{ACTION:} open drawer 1 \hfill \textbf{PROGRESS RATE:} 0.5

\textbf{OBS:} You open the drawer 1. The drawer 1 is open. In it, you see a spoon 1.

\textbf{SKILL Centered on action 'open drawer'}
\begin{itemize}
  \item Common precursors to this action:
  \begin{itemize}
    \item Agents often perform 'go to drawer' before 'open drawer'.
  \end{itemize}
  \item Typical next steps after this action:
  \begin{itemize}
    \item After 'open drawer', agents usually continue with 'put peppershaker in/on drawer'.
  \end{itemize}
\end{itemize}

\textbf{ACTION:} put peppershaker 1 in/on drawer 1 \hfill \textbf{PROGRESS RATE:} 0.75

\textbf{OBS:} You put the peppershaker 1 in/on the drawer 1.

\textbf{SKILL Centered on action 'put peppershaker in/on drawer'}
\begin{itemize}
  \item Common precursors to this action:
  \begin{itemize}
    \item Agents often perform 'open drawer' before 'put peppershaker in/on drawer'.
  \end{itemize}
  \item Typical next steps after this action:
  \begin{itemize}
    \item After 'put peppershaker in/on drawer', agents usually continue with 'go to countertop'.
  \end{itemize}
\end{itemize}

\textbf{ACTION:} go to countertop 2 \hfill \textbf{PROGRESS RATE:} 0.75

\textbf{OBS:} On the countertop 2, you see a bread 1, a butterknife 1, a cellphone 1, a cup 3, a knife 1, a pan 1, a papertowelroll 1, a saltshaker 1, and a spatula 1.

\textbf{ACTION:} take peppershaker 2 from countertop 2 \hfill \textbf{PROGRESS RATE:} 0.75

\textbf{OBS:} Nothing happens.

\textbf{ACTION:} go to countertop 3 \hfill \textbf{PROGRESS RATE:} 0.75

\textbf{OBS:} On the countertop 3, you see a butterknife 2, a creditcard 3, a dishsponge 3, a houseplant 1, a knife 2, a lettuce 1, a peppershaker 2, a spatula 3, a vase 3, a vase 2, and a vase 1.

\textbf{ACTION:} take peppershaker 2 from countertop 3 \hfill \textbf{PROGRESS RATE:} 0.75

\textbf{OBS:} You pick up the peppershaker 2 from the countertop 3.

\textbf{ACTION:} go to drawer 1 \hfill \textbf{PROGRESS RATE:} 0.75

\textbf{OBS:} The drawer 1 is open. In it, you see a peppershaker 1, and a spoon 1.

\textbf{SKILL Centered on action 'go to drawer'}
\begin{itemize}
  \item Common precursors to this action:
  \begin{itemize}
    \item Agents often perform 'go to countertop' before 'go to drawer'.
  \end{itemize}
  \item Typical next steps after this action:
  \begin{itemize}
    \item After 'go to drawer', agents usually continue with 'open drawer'.
  \end{itemize}
\end{itemize}

\textbf{ACTION:} put peppershaker 2 in/on drawer 1 \hfill \textbf{PROGRESS RATE:} 1.0

\end{tcolorbox}

Case B showcases how skill-enhanced prompting enables structured and efficient multi-step decision-making in ScienceWorld, a more complex environment. At the task's outset, the agent leverages the skill suggesting “go to hallway” as a common first step, thereby avoiding unproductive exploration in the workshop. As it navigates toward the target, skills centered on “look around” and “go to greenhouse” guide the agent through meaningful spatial transitions. These skills are domain-general: they help not only in tasks involving egg retrieval, but also in any setting requiring room exploration. Upon encountering the blue jay egg, the agent applies a skill pattern originally associated with the “butterfly egg” task—namely, a two-step focus-then-pick-up strategy—demonstrating the generalization of interaction structures across object types. Later, skills associated with “go to kitchen” and “look around” support the correct identification of the delivery location. Finally, although no explicit skill specifies “move X to Y,” the agent successfully executes the goal action by generalizing from previously learned manipulation patterns. This example illustrates how skill-driven scaffolding provides trajectory templates, narrows the action space, and promotes compositional generalization, ultimately allowing the agent to complete the task with a final progress rate of 1.0. This compositional reuse of mid-level routines highlights how skills operate as modular and interpretable units of behavior, enabling robust generalization and precise control for LLMs.

\begin{tcolorbox}[
  enhanced,
  breakable,
  colback=white,
  colframe=blue!60!black,
  coltitle=black,
  title=\faTasks\hspace{0.5em} Case B: Blue Jay Egg → Purple Box,
  fonttitle=\bfseries,
  colbacktitle=blue!10!white,
  sharp corners,
  boxrule=0.8pt,
  left=6pt,
  right=6pt,
  top=6pt,
  bottom=6pt,
]\label{box:skill-traj-sc}

\textbf{Goal:} Your task is to find a blue jay egg in the outside. First, focus on the thing. Then, move it to the purple box in the kitchen.\\
\textbf{Progress Rate:} 1.0

\vspace{0.8em}
\textbf{OBS:} This room is called the workshop. In it, you see: the agent, a substance called air, a table. On the table is: a battery, a black wire, a blue wire, a orange wire, a red light bulb, which is off, a switch, which is off, a violet light bulb, which is off, a yellow light bulb, which is off. An ultra low temperature freezer. The ultra low temperature freezer door is closed.\\
You also see: A door to the hallway (that is open).\\
Inventory: an orange

\textbf{SKILL: Centered on action 'the beginning of the task'}
\begin{itemize}
  \item Typical next steps after this action:
  \begin{itemize}
    \item After 'the beginning of the task', agents usually continue with 'go to hallway'.
    \item After 'the beginning of the task', agents usually continue with 'check valid actions'.
    \item After 'the beginning of the task', agents usually continue with 'go to greenhouse'.
  \end{itemize}
\end{itemize}

\textbf{ACTION:} go to hallway \hfill \textbf{PROGRESS RATE:} 0.0

\vspace{0.5em}
\textbf{OBS:} You move to the hallway.

\textbf{SKILL: Centered on action 'go to hallway'}
\begin{itemize}
  \item Common precursors to this action:
  \begin{itemize}
    \item Agents often perform 'the beginning of the task' before 'go to hallway'.
    \item Agents often perform 'look around' before 'go to hallway'.
    \item Agents often perform 'go to living room' before 'go to hallway'.
  \end{itemize}
  \item Typical next steps after this action:
  \begin{itemize}
    \item After 'go to hallway', agents usually continue with 'go to bedroom'.
    \item After 'go to hallway', agents usually continue with 'look around'.
  \end{itemize}
\end{itemize}

\textbf{ACTION:} look around \hfill \textbf{PROGRESS RATE:} 0.0

\vspace{0.5em}
\textbf{OBS:} This room is called the hallway. In it, you see: the agent, a substance called air, a painting.\\
You also see: A door to the art studio (that is open), A door to the bedroom (that is open), A door to the greenhouse (that is open), A door to the kitchen (that is open), A door to the living room (that is open), A door to the workshop (that is open)

\textbf{SKILL: Centered on action 'look around'}
\begin{itemize}
  \item Common precursors to this action:
  \begin{itemize}
    \item Agents often perform 'the beginning of the task' before 'look around'.
    \item Agents often perform 'pick up flower pot' before 'look around'.
    \item Agents often perform 'check valid actions' before 'look around'.
  \end{itemize}
  \item Typical next steps after this action:
  \begin{itemize}
    \item After 'look around', agents usually continue with 'go to hallway'.
    \item After 'look around', agents usually continue with 'focus on butterfly egg'.
    \item After 'look around', agents usually continue with 'pick up shovel'.
  \end{itemize}
\end{itemize}

\textbf{ACTION:} go to greenhouse \hfill \textbf{PROGRESS RATE:} 0.0

\vspace{0.5em}
\textbf{OBS:} You move to the greenhouse.

\textbf{SKILL: Centered on action 'go to greenhouse'}
\begin{itemize}
  \item Common precursors to this action:
  \begin{itemize}
    \item Agents often perform 'the beginning of the task' before 'go to greenhouse'.
    \item Agents often perform 'look around' before 'go to greenhouse'.
  \end{itemize}
  \item Typical next steps after this action:
  \begin{itemize}
    \item After 'go to greenhouse', agents usually continue with 'look around'.
  \end{itemize}
\end{itemize}

\textbf{ACTION:} look around \hfill \textbf{PROGRESS RATE:} 0.0

\vspace{0.5em}
\textbf{OBS:} This room is called the greenhouse. In it, you see: the agent, a substance called air, a bee hive (closed), flower pots, a jug, a shovel, a sink (off).\\
You also see: A door to the hallway (that is open), A door to the outside (that is open)

\textbf{SKILL: Centered on action 'look around'}
\begin{itemize}
  \item Common precursors to this action:
  \begin{itemize}
    \item Agents often perform 'the beginning of the task' before 'look around'.
    \item Agents often perform 'pick up flower pot' before 'look around'.
    \item Agents often perform 'check valid actions' before 'look around'.
  \end{itemize}
  \item Typical next steps after this action:
  \begin{itemize}
    \item After 'look around', agents usually continue with 'go to hallway'.
    \item After 'look around', agents usually continue with 'focus on butterfly egg'.
    \item After 'look around', agents usually continue with 'pick up shovel'.
  \end{itemize}
\end{itemize}

\textbf{ACTION:} go to outside \hfill \textbf{PROGRESS RATE:} 0.2

\vspace{0.5em}
\textbf{OBS:} You move to the outside.

\textbf{SKILL: Centered on action 'go to outside'}
\begin{itemize}
  \item Common precursors to this action:
  \begin{itemize}
    \item Agents often perform 'the beginning of the task' before 'go to outside'.
    \item Agents often perform 'check valid actions' before 'go to outside'.
    \item Agents often perform 'look around' before 'go to outside'.
  \end{itemize}
  \item Typical next steps after this action:
  \begin{itemize}
    \item After 'go to outside', agents usually continue with 'look around'.
  \end{itemize}
\end{itemize}

\textbf{ACTION:} look around \hfill \textbf{PROGRESS RATE:} 0.2

\vspace{0.5em}
\textbf{OBS:} This outside location is called the outside. Here you see: the agent, a substance called air, an axe, a blue jay egg, a butterfly egg, a dove egg, a fire pit (empty), a fountain (with water), the ground, a substance called wood.\\
You also see: A door to the foundry (open), A door to the greenhouse (open), A door to the kitchen (open)

\textbf{SKILL: Centered on action 'look around'}
\begin{itemize}
  \item Common precursors to this action:
  \begin{itemize}
    \item Agents often perform 'the beginning of the task' before 'look around'.
    \item Agents often perform 'pick up flower pot' before 'look around'.
    \item Agents often perform 'check valid actions' before 'look around'.
  \end{itemize}
  \item Typical next steps after this action:
  \begin{itemize}
    \item After 'look around', agents usually continue with 'go to hallway'.
    \item After 'look around', agents usually continue with 'focus on butterfly egg'.
    \item After 'look around', agents usually continue with 'pick up shovel'.
  \end{itemize}
\end{itemize}

\textbf{ACTION:} focus on blue jay egg \hfill \textbf{PROGRESS RATE:} 0.4

\vspace{0.5em}
\textbf{OBS:} You focus on the blue jay egg.

\textbf{SKILL: Centered on action 'focus on butterfly egg'}
\begin{itemize}
  \item Common precursors to this action:
  \begin{itemize}
    \item Agents often perform 'look around' before 'focus on butterfly egg'.
  \end{itemize}
  \item Typical next steps after this action:
  \begin{itemize}
    \item After 'focus on butterfly egg', agents usually continue with 'pick up butterfly egg'.
  \end{itemize}
\end{itemize}

\textbf{ACTION:} pick up blue jay egg \hfill \textbf{PROGRESS RATE:} 0.6

\vspace{0.5em}
\textbf{OBS:} You move the blue jay to the inventory.

\textbf{ACTION:} go to kitchen \hfill \textbf{PROGRESS RATE:} 0.8

\vspace{0.5em}
\textbf{OBS:} You move to the kitchen.

\textbf{SKILL: Centered on action 'go to kitchen'}
\begin{itemize}
  \item Typical next steps after this action:
  \begin{itemize}
    \item After 'go to kitchen', agents usually continue with 'look around'.
  \end{itemize}
\end{itemize}

\textbf{ACTION:} look around \hfill \textbf{PROGRESS RATE:} 0.8

\vspace{0.5em}
\textbf{OBS:} The kitchen contains: various furniture, a bowl with fruit, a purple box (empty), stove, oven, thermometer (10°C), etc.\\
You also see: A door to the bathroom (open), hallway (open), outside (open)

\textbf{SKILL: Centered on action 'look around'}
\begin{itemize}
  \item Common precursors to this action:
  \begin{itemize}
    \item Agents often perform 'the beginning of the task' before 'look around'.
    \item Agents often perform 'pick up flower pot' before 'look around'.
    \item Agents often perform 'check valid actions' before 'look around'.
  \end{itemize}
  \item Typical next steps after this action:
  \begin{itemize}
    \item After 'look around', agents usually continue with 'go to hallway'.
    \item After 'look around', agents usually continue with 'focus on butterfly egg'.
    \item After 'look around', agents usually continue with 'pick up shovel'.
  \end{itemize}
\end{itemize}

\textbf{ACTION:} move blue jay egg to purple box \hfill \textbf{PROGRESS RATE:} 1.0

\end{tcolorbox}


\end{document}